\documentclass[10pt,twocolumn,letterpaper]{article}

\usepackage[final,algorithms]{wacv} 
\usepackage{times}
\usepackage{epsfig}
\usepackage{graphicx}
\usepackage{amsmath}
\usepackage{amssymb}

\usepackage{float}
\usepackage{lipsum}
\usepackage{stfloats}
\usepackage{multicol}
\usepackage{multirow}
\usepackage{bm}
\usepackage{etoolbox}

\usepackage{array}
\usepackage{tabulary}
\usepackage[table,dvipsnames]{xcolor}
\usepackage{paralist}
\usepackage{booktabs}
\usepackage{adjustbox}
\usepackage[shortlabels]{enumitem}
\usepackage{tabu}
\usepackage{caption}
\usepackage{subcaption}
\usepackage[pagebackref=true,breaklinks=true,letterpaper=true,colorlinks,bookmarks=false,citecolor=blue,linkcolor=red]{hyperref}
\usepackage{icomma}
\usepackage{arydshln}

\definecolor{gray}{rgb}{0.3,0.3,0.3}
\definecolor{blue}{rgb}{0,0.5,1}
\definecolor{mask_red}{rgb}{1,0,0.8}
\definecolor{green}{rgb}{0.2,1,0.2}
\definecolor{rblue}{rgb}{0,0,1}
\definecolor{lightblue}{HTML}{6495ed}
\definecolor{lightred}{HTML}{F19C99}

\newcommand{\green}[1]{\textcolor[RGB]{96,177,87}{#1}}

\newcommand{\fn}[1]{\footnotesize{#1}}
\newcommand{\gbf}[1]{\green{\bf{\fn{(#1)}}}}

\newcommand{\obf}[1]{\textcolor{orange}{\bf{\fn{(#1)}}}}
\definecolor{graytablerow}{gray}{0.6}

\usepackage[accsupp]{axessibility}  

\usepackage{tikz}
\newcommand*\circled[1]{\tikz[baseline=(char.base)]{
\node[shape=circle,fill=gray,inner sep=0.5pt] (char) {\textcolor{white}{\footnotesize \textbf{#1}}};}}

\begin{document}

\title{360BEV: Panoramic Semantic Mapping for Indoor Bird's-Eye View}

\author{
Zhifeng Teng$^{1,}$\thanks{Equal contribution.},
~~Jiaming Zhang$^{1,*,}$\thanks{Corresponding author (e-mail: {\tt jiaming.zhang@kit.edu}).},
~~Kailun Yang$^2$,
~~Kunyu Peng$^1$,\\
~~Hao Shi$^3$,
~~Simon Reiß$^{1}$,
~~Ke Cao$^{1}$,
~~Rainer Stiefelhagen$^1$\\
\normalsize
$^1$Karlsruhe Institute of Technology,
\normalsize
~$^2$Hunan University,
\normalsize
~$^3$Zhejiang University
}

\maketitle

\begin{abstract}
Seeing only a tiny part of the whole is not knowing the full circumstance. Bird's-eye-view (BEV) perception, a process of obtaining allocentric maps from egocentric views, is restricted when using a narrow Field of View (FoV) alone. In this work, mapping from 360{\textdegree} panoramas to BEV semantics, the \textbf{360BEV} task, is established for the first time to achieve holistic representations of indoor scenes in a top-down view. Instead of relying on narrow-FoV image sequences, a panoramic image with depth information is sufficient to generate a holistic BEV semantic map. To benchmark 360BEV, we present two indoor datasets, 360BEV-Matterport and 360BEV-Stanford, both of which include egocentric panoramic images and semantic segmentation labels, as well as allocentric semantic maps. Besides delving deep into different mapping paradigms, we propose a dedicated solution for panoramic semantic mapping, namely \textbf{360Mapper}. Through extensive experiments, our methods achieve $44.32\%$ and $45.78\%$ mIoU on both datasets respectively, surpassing previous counterparts with gains of ${+}7.60\%$ and ${+}9.70\%$ in mIoU.\footnote{The datasets and code are available at the project page \href{https://jamycheung.github.io/360BEV.html}{\texttt{360BEV}}.}
\end{abstract}

\begin{figure}[t]
    \centering
    \includegraphics[width=0.99\linewidth]{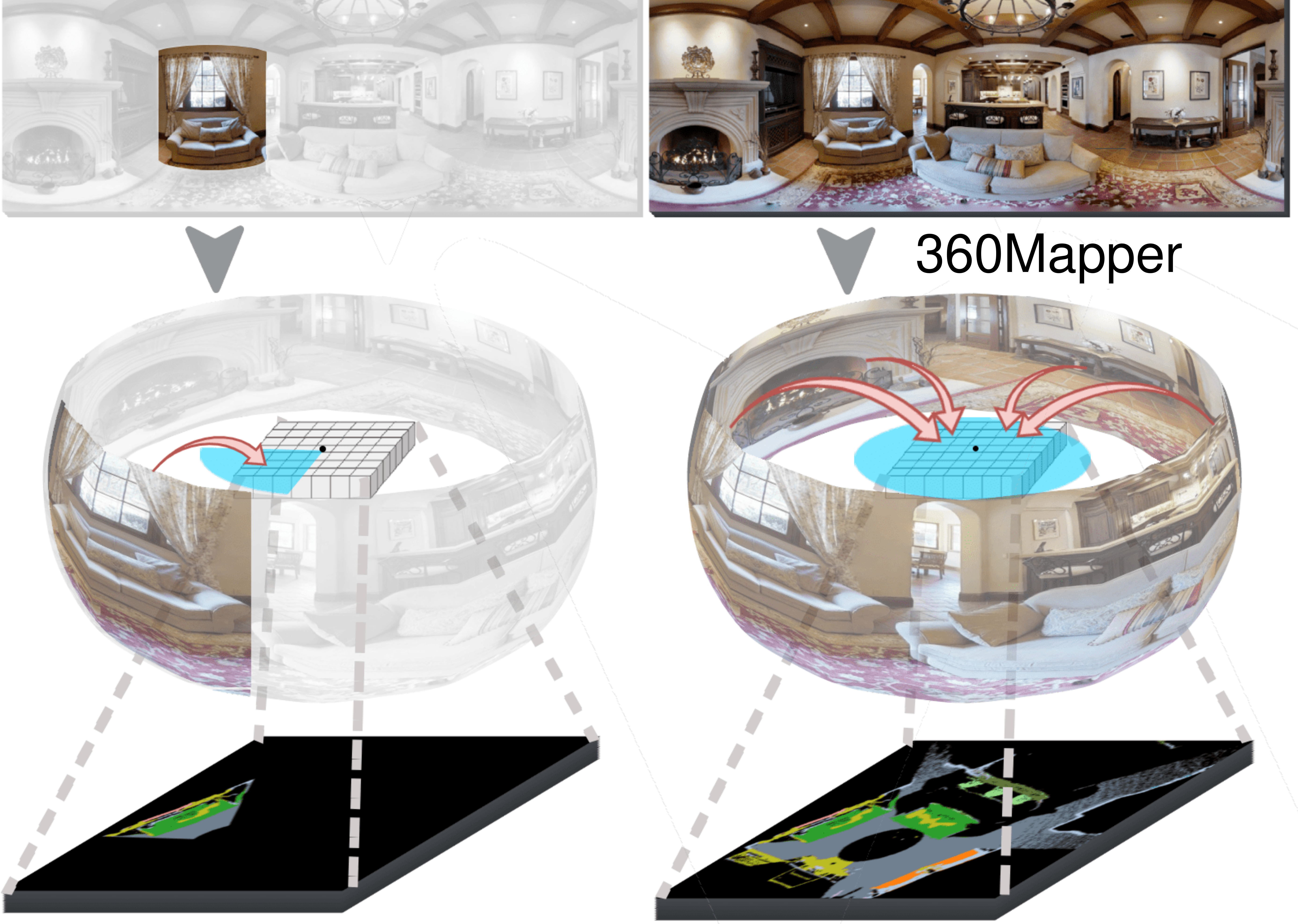}
    \begin{minipage}[t]{.5\columnwidth}
        \vskip-3ex
        \subcaption{Narrow BEV}\label{fig1_narrowbev}
    \end{minipage}%
    \begin{minipage}[t]{.5\columnwidth}
        \vskip-3ex
        \subcaption{360BEV}\label{fig1_360bev}
    \end{minipage}%
    \vskip-1ex
    \caption{Semantic mapping from egocentric front-view images to allocentric BEV semantics. While (a) the narrow-BEV method has limited perception and map range, (b) 360BEV has an omnidirectional \textcolor[HTML]{3cc7fa}{Field of View}, yielding a more complete BEV map by using our 360Mapper model.}
    \label{fig:360bev}
    \vskip-3ex
\end{figure}

\section{Introduction}
\label{sec:intro}
Semantic scene understanding has achieved remarkable performance on indoor- and outdoor scenes via pixel-wise semantic segmentation~\cite{mivcuvslik2009semantic}. It can be utilized directly on a wide range of downstream applications, such as autonomous driving~\cite{feng2020deep,janai2020cv4auto}, navigation in robotics~\cite{SMNet,chen2022trans4map} or in assistive technologies~\cite{zhang2021trans4trans} to name a few.
Recently, Bird's-Eye-View (BEV) semantic perception~\cite{bevformer} can be a solution for enabling a straightforward understanding of the environment and objects therein.
While BEV semantic segmentation has gained traction in outdoor scenes for autonomous driving~\cite{bevformer}, BEV perception has not yet been extensively explored for indoor scenes, which are often characterized by complex and varied structures, objects, and challenging lighting conditions.
For semantically mapping these indoor scenes, sequence-based methods~\cite{SMNet,chen2022trans4map} were proposed, which have to process whole videos and entail a moving camera. As shown in Fig.~\ref{fig1_narrowbev}, (1) these methods rely on computationally expensive processing of entire sequences of video-frames due to the narrow \textcolor[HTML]{3cc7fa}{Field of View} of the pinhole camera, and (2) they are constrained to explore indoor mapping on synthetic simulators~\cite{savva2019habitat,xia2018gibson}, due to the lack of real indoor datasets.
These drawbacks limit their applicability to real-world indoor semantic mapping.

To solve these limitations, in this work, we introduce \textbf{360BEV} to achieve panoramic semantic mapping for indoor BEV, which is illustrated in Fig.~\ref{fig1_360bev}.
Our considerations are twofold:
(1) To unleash the potential of indoor semantic mapping in real-world scenarios, real indoor databases with BEV semantic labels are crucial; (2) To reduce the computational complexity of narrow-FoV sequence methods~\cite{SMNet} (${\geq}20$ video-frames to process) 
or the complexity of multi-camera setups~\cite{bevformer} (${\geq}6$ camera views needed), 
we leverage a single-frame 360{\textdegree} image with depth information and thus bypass multi-sensor calibration, synchronization, and data fusion procedures.
With this in mind and to enable 360BEV segmentation we present two real indoor BEV datasets, which are extended from the Matterport3D~\cite{Matterport3D} and Stanford2D3D~\cite{stanford2d3d} datasets. First, the Front-View images captured by pinhole cameras from Matterport3D are extended to 360{\textdegree} panoramas for benchmarking on \textbf{360FV-Matterport}.
Furthermore, for the first time, two BEV datasets, \textbf{360BEV-Matterport} and \textbf{360BEV-Stanford} are established to enable bird's-eye view panoramic semantic mapping, \ie, predicting a complete BEV semantic map from a single-frame 360{\textdegree} image with depth.
Moreover, by decoupling the computationally expensive processing of sequences or multiple views, our direct 360BEV semantic mapping is more streamlined for generating indoor semantic maps.  

However, spatial distortions and object deformations in panoramic images~\cite{trans4pass} severely harm the performance of methods proposed for narrow-range image~\cite{guo2022segnext,xie2021segformer} or multi-view perception~\cite{bevformer}.
Thus, to comprehensively investigate the established 360BEV task, we first revisit three possible projection paradigms, including: (1) \textit{Early projection}, (2) \textit{Late projection}, and (3) \textit{Intermediate projection}. Based on our observation that intermediate features maintain dense information, we explore the intermediate projection paradigm and propose a dedicated solution for 360BEV mapping, which we call \textbf{360Mapper}. The challenge in this scheme resides in the feature conversion. While the prior BEVFormer~\cite{bevformer} relied on multi-view perception and SMNet~\cite{SMNet} projects the extracted feature directly via the depth-based transformation index, which is not appropriate for panoramic imagery due to its distortions and deformations, we propose a new transformation method, the \textbf{Inverse Radial Projection (IRP)}, to project features from 2D to 3D representations using only depth information. 
An additional benefit is that the depth information helps maintain object shape and space layout after being transferred to top-down views, rendering the 2D reference index for the feature map as well as the BEV representation more accurate and consistent. Besides, unlike the deformable attention~\cite{bevformer,zhu2020deformabledetr} using multi-scale layers and fusion from multi-view cameras, we adopt \textbf{360Attention} with adaptive sampling offsets to extract information from omnidirectional feature maps, yielding the bird's-eye-view feature with less distortion in an adaptive manner. 
These are combined with the 2D index obtained by IRP to include a deformation-aware mechanism in 360 scenes, which in turn serves to compensate for the adverse effects of distortion. 
With these designs, our 360Mapper model represents a step towards a more complete and accurate indoor semantic mapping, which has important implications for downstream applications such as indoor navigation and scene understanding.

Through extensive experiments, the new 360BEV task is thoroughly benchmarked with two real indoor BEV datasets, three projection paradigms, and more than ten methods, respectively. Compared to the semantic mapping counterparts, our 360Mapper models achieve state-of-the-art performance, with mean intersection-over-union (mIoU) gains of ${>}7\%$ on the 360BEV-Matterport dataset and ${>}9\%$ on the 360BEV-Stanford dataset. 

To summarize, we present the following contributions:
\begin{compactitem}
\item A new \textit{360BEV} task is introduced for the first time to address indoor semantic mapping via a single-frame panoramic image, decoupling complex processing of multi-view or sequence inputs. 

\item Two indoor BEV datasets, \ie, \textit{360BEV-Matterport} and \textit{360BEV-Stanford}, are extended with front-view panoramic images and BEV semantic labels, thoroughly benchmarking panoramic semantic mapping. 

\item \textit{360Mapper} model -- addressing spatial distortions and object deformations in panoramas -- is proposed as a dedicated solution for interior panoramic semantic mapping and achieves state-of-the-art performance.

\end{compactitem}

\section{Related Work}
\label{sec:related_work}
\subsection{Panoramic Semantic Segmentation}
Image semantic segmentation~\cite{wang2021hrnet,xie2021segformer,yuan2020ocr,zhao2017pspnet} has achieved great progress.
In contrast to narrow-FoV perception, panoramic semantic segmentation~\cite{gauge_equivariant,tangent,esteves2020spin_weighted_spherical,spherical_unstructured_grids,hohonet,distortion_aware,orientation,trans4pass}, yielding holistic scene understanding by using a single 360{\textdegree} front-view image, has received increasing attention in recent years. Besides, 3D60~\cite{zioulis20193D60} and Pano3D~\cite{albanis2021pano3d} datasets are generated for depth estimation from 360{\textdegree} images, but lack semantic labels.
In indoor panorama segmentation, there are some benchmarks that provide synthetic~\cite{InteriorNet18,structured3d} and real~\cite{stanford2d3d} panoramic images and labels for training. Matterport3D~\cite{Matterport3D} has large-scale panoramic images collected from $90$ indoor buildings, yet, it has not been benchmarked due to the lack of corresponding panoramic semantic labels.
To enable this, we generate the panoramic semantic segmentation labels by combining the original $18$ pinhole camera labels regarding their camera transformation matrices. Therefore, a 360{\textdegree} Front-View (FV) dataset, \textit{360FV-Matterport}, with large-scale real indoor scenes, is provided to facilitate panoramic semantic segmentation. Besides, the 360FV-Matterport dataset is required to perform the late-projection paradigm of BEV semantic mapping.

\subsection{BEV Semantic Mapping}
Apart from front-view image semantic segmentation, some previous work explored top-view semantic segmentation, known as semantic mapping~\cite{chen2022trans4map,mivcuvslik2009semantic} in indoor scenes and bird's-eye-view semantic segmentation~\cite{bevformer,peng2023bevsegformer} in outdoor driving scenes. 
The indoor semantic mapping methods can be divided into three categories according to the level of projection from the front view to the top-down view: \textit{Early-projection} approaches~\cite{mattyus2015enhancing,singh2018self} are performed via general semantic segmentation methods, which first construct the BEV views from perspective images and then apply segmentation. Unfortunately, these pipelines lose fine-grained visual cues during the projection and thus result in unsatisfactory performance for small object segmentation.
\textit{Intermediate-projection} methods~\cite{SMNet,chen2022trans4map} directly take front views as input for holistic indoor scene understanding,
however, they work on synthetic data generated from Gibson~\cite{xia2018gibson} or Habitat~\cite{savva2019habitat} simulators and rely on time-consuming image sequences. For example, SMNet~\cite{SMNet} gradually captures an average of $2,500$ view-points for each floor to generate a semantic map for indoor scenes.  
Instead, we explore achieving efficient allocentric scene understanding via a single panorama image.
The \textit{Late-projection} pipeline~\cite{anderson2018vision,grinvald2019volumetric,maturana2018real,ran2021rs,sengupta2012automatic} performs egocentric semantic segmentation and project labels to top-down views, which are sensitive to depth map and agent pose information, inevitably facing the projection error and under-fitting of model training, thus remaining a suboptimal solution.
There are some BEV-related methods that leverage multiple perspective view sensors or LiDAR sensors and focus on outdoor object detection~\cite{bevformer,liu2022petr,yang2022bevformer}, optical flow estimation~\cite{lee2020pillarflow,luo2021self}, and semantic segmentation~\cite{pan2020cross,zhou2022cross}. Different from previous methods, our 360Mapper is carefully designed for learning indoor holistic representations by forwarding a single panorama without using multi-view images, image sequences, or point clouds.

\begin{figure*}[t]
    \centering
    \includegraphics[width=0.99\linewidth]{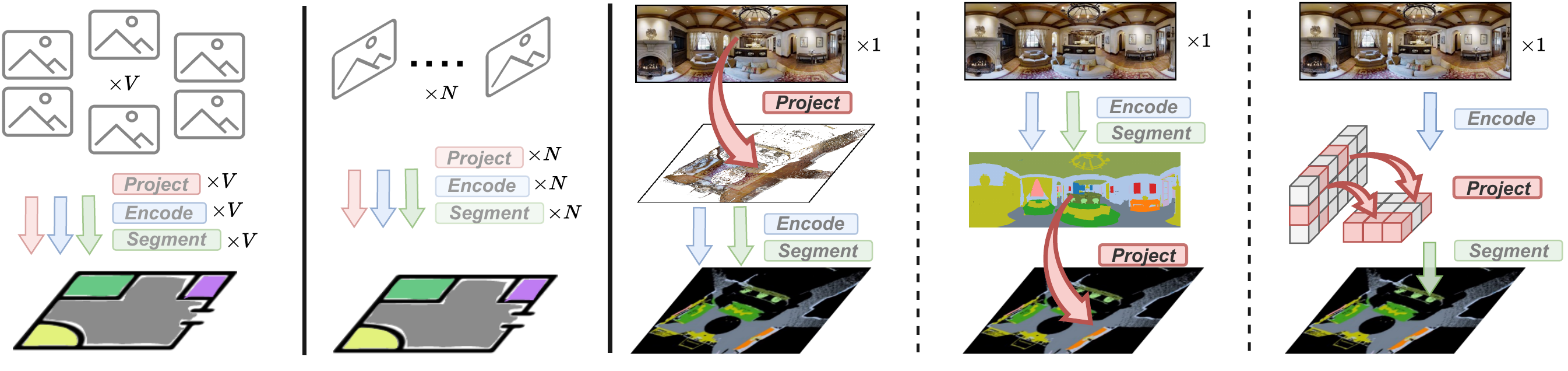}
    \begin{minipage}[t]{.2\linewidth}
        \vskip-3ex
        \subcaption{Multi-view}\label{paradimgs_1}
    \end{minipage}%
    \begin{minipage}[t]{.2\linewidth}
        \vskip-3ex
        \subcaption{Sequence-based}\label{paradimgs_2}
    \end{minipage}%
    \begin{minipage}[t]{.2\linewidth}
        \vskip-3ex
        \subcaption{Early-projection}\label{paradimgs_3}
    \end{minipage}%
    \begin{minipage}[t]{.2\linewidth}
        \vskip-3ex
        \subcaption{Late-projection}\label{paradimgs_4}
    \end{minipage}%
    \begin{minipage}[t]{.2\linewidth}
        \vskip-3ex
        \subcaption{Intermediate-projection}\label{paradimgs_5}
    \end{minipage}%
    \vskip-1ex
    \caption{\textbf{Paradigms of semantic mapping.} While the narrow-FoV (a) multi-view and (b) sequence-based methods rely on $V{\geq}6$ and $N{\geq}20$ views, the 360{\textdegree}-BEV (c) Early-, (d) Late-, and (e) Intermediate-projection methods use a single panorama.}
    \label{fig:paradigms}
    \vskip-2ex
\end{figure*}

\section{Panorama Semantic Mapping (360BEV)}
\label{sec:methodology}
To investigate the 360BEV task, we analyze potential panoramic projection paradigms in Sec.~\ref{sec:3_1_paradigm}. The generation and data statistics of the dataset are detailed in Sec.~\ref{sec:3_3_dataset}. To tackle the challenging panoramic semantic mapping, in Sec.~\ref{sec:3_4_model} we present our solution \textbf{360Mapper} with the \textbf{Inverse Radial Projection} method and \textbf{360Attention} module, which enable distortion-aware feature processing.

\subsection{360 Projection Paradigms}~\label{sec:3_1_paradigm}
As shown in Fig.~\ref{fig:paradigms}, unlike multi-view methods relying on more than six views ($V$ in Fig.~\ref{paradimgs_1}) and sequence-based methods using more than $20$ narrow views ($N$ in Fig.~\ref{paradimgs_2}), panoramic semantic mapping uses a single image with depth.
We investigate three projection paradigms, \ie, \textit{how to process data from front-view panoramas to bird's-eye-view semantics}, which are:
\begin{compactitem}
    \item[(1)] \textit{Early projection: \textbf{Proj.}${\rightarrow}$Enc.${\rightarrow}$Seg.} in Fig.~\ref{paradimgs_3}. 
    \item[(2)] \textit{Late projection: Enc.${\rightarrow}$Seg.${\rightarrow}$\textbf{Proj.}} in Fig.~\ref{paradimgs_4}.
    \item[(3)] \textit{Intermediate projection: Enc.${\rightarrow}$\textbf{Proj.}${\rightarrow}$Seg.} in Fig.~\ref{paradimgs_5}. 
\end{compactitem}
Based on these properties, we mainly explore 360BEV with intermediate projections, in which we identify the following challenges: In the feature extraction stage, spatial distortions and object deformations severely hinder the encoder from extracting representative features from the front-view panoramic image. For the intermediate feature projection, only depth information is utilized for consistent view transformation of high-dimensional features. In addition, many large objects in the front view (\eg, \textit{walls}) are projected to thin objects in the top-down view, which greatly impedes capturing wide-range features during projection.

\subsection{360FV and 360BEV Data Generation}~\label{sec:3_3_dataset}

\begin{figure}[t]
    \centering
    \includegraphics[width=0.99\linewidth]{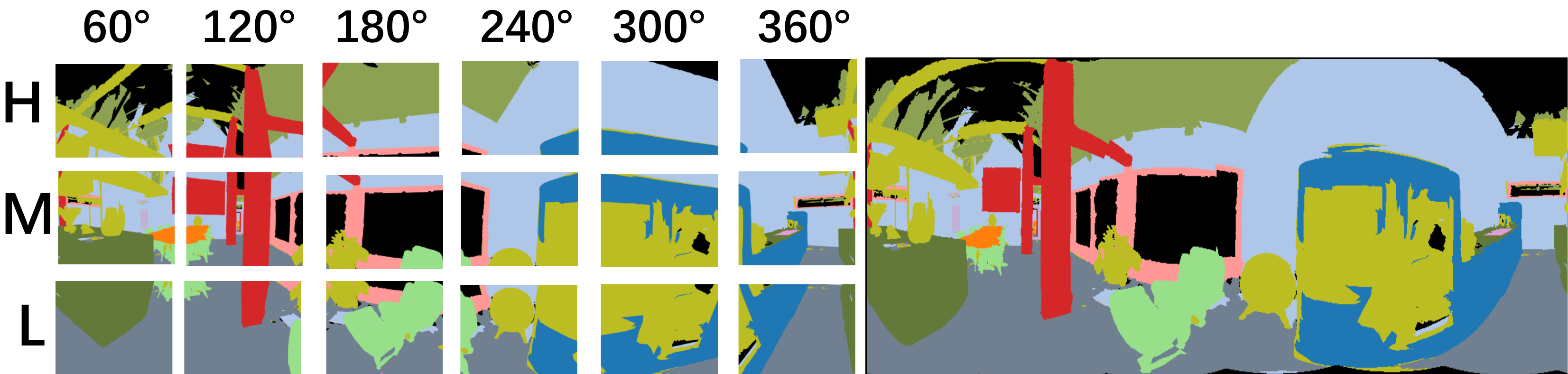}
    \vskip-1ex
    \caption{\textbf{360FV semantics generation} from $18$ narrow views to a panoramic view on the 360FV-Matterport dataset. $H$, $M$, and $L$ represent high, medium, and low positions, respectively. } 
    \label{fig:data_gen}
    \vskip-1ex
\end{figure}

\noindent\textbf{360FV-Matterport.} 
The original Matterport3D~\cite{Matterport3D} was collected via narrow-FoV cameras. As shown in Fig.~\ref{fig:data_gen}, we convert the $18$ narrow-view images and annotations into the 360{\textdegree} format by using rotation-translation matrices. 

\begin{figure}[t]
    \centering
    \includegraphics[width=0.99\linewidth]{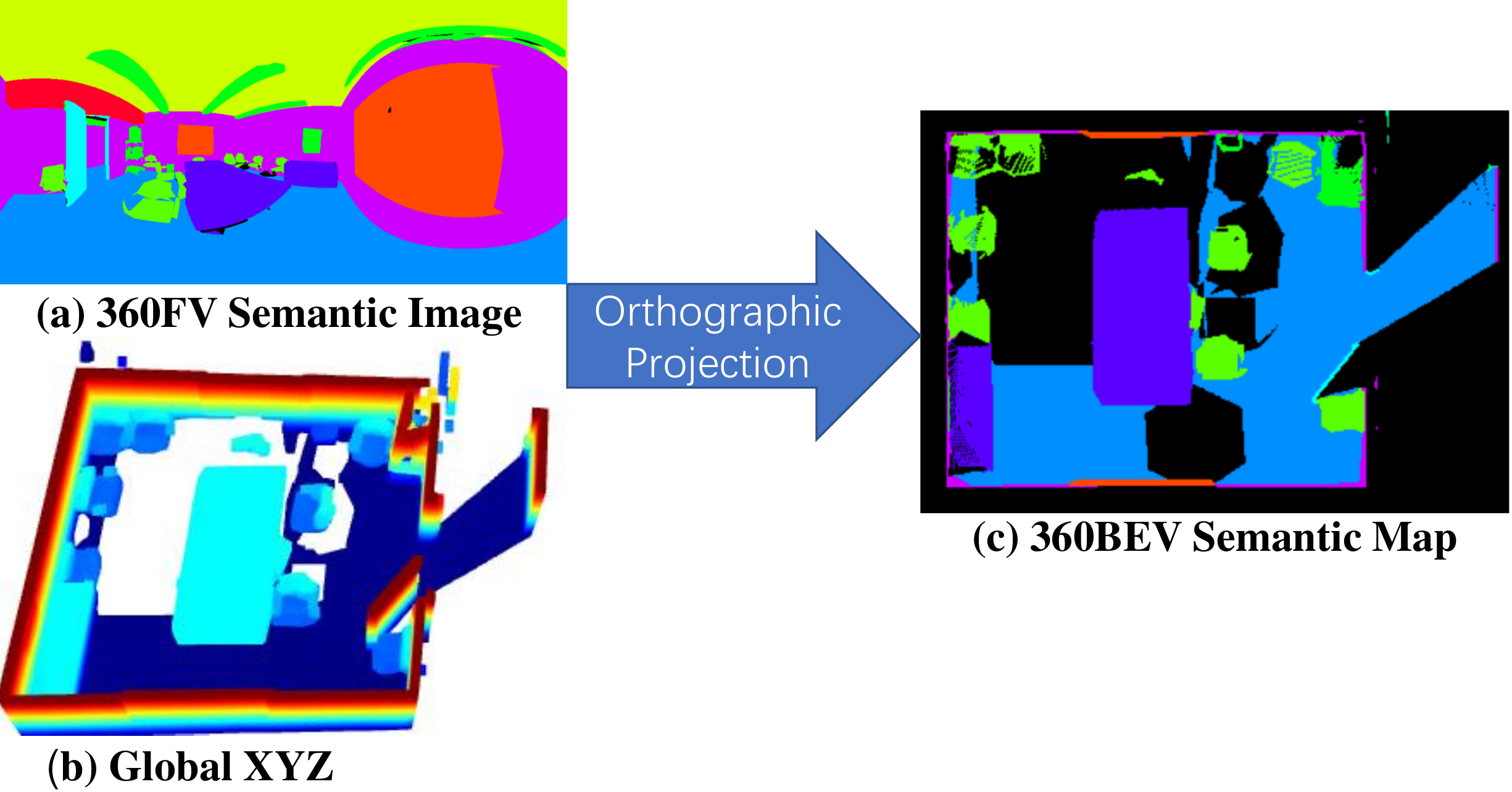}
    \vskip-1ex
    \caption{\textbf{360BEV semantics generation} by orthographic projection, from (a) the front-view semantic image and (b) the global XYZ image, to (c) the 360BEV semantic map. }
    \label{fig:data_gen_bev}
    \vskip-2ex
\end{figure}

\noindent \textbf{360BEV-Stanford.} 
The Stanford2D3D dataset~\cite{stanford2d3d} has front-view panoramic images and semantic labels. However, it lacks BEV semantic labels. As presented in Fig.~\ref{fig:data_gen_bev}, we utilize the spatial semantic information from the global XYZ image to generate the corresponding BEV semantic map. 
By applying orthographic projection, we generate the BEV semantic maps within a visible range as BEV ground truth, enabling end-to-end training from font-view images to top-down semantics. 

\noindent \textbf{360BEV-Matterport.} 
Inspired by the global XYZ modality~\cite{stanford2d3d}, we generate a global XYZ for each panoramic image by using the provided depth ground truth. In order to generate BEV semantic ground truth corresponding to the panoramic view, several key steps must be considered. Firstly, a panoramic image can be processed as a sphere with rays shooting from the center of the sphere, where the camera is located. 
\begin{equation}\label{eq:1}
\begin{aligned}
\Theta_{i,j} &= \frac{i\pi}{H} + \frac{\pi}{2H}, \\  &i =\{0,\ldots,H{-}1\}, \ j=\{0,\ldots,W{-}1\}, \\
\Phi_{i,j} &= - \frac{2\pi j}{W} + \pi - \frac{\pi}{W},\\ &i=\{0,\ldots,H{-}1\}, \ j=\{0,\ldots,W{-}1\}.
\end{aligned}
\end{equation}
Here, $\Theta$ and $\Phi$ are angle matrices of panoramic images with size $H {\times} W$, which consist of two dimensional Euler angular equivariant series.
Given the representation in spherical coordinate systems, each 3D point $(X_{i,j}, Y_{i,j}, Z_{i,j})$ in the camera coordinate system will be obtained through the calculation in Eq.~\eqref{eq:2},
\begin{equation}\label{eq:2}
\begin{aligned}
X_{i,j} &= D_{i,j} \cdot \sin(\Theta_{i,j}) \cdot \sin(\Phi_{i,j}), \\
Y_{i,j} &= D_{i,j} \cdot \cos(\Theta_{i,j}), \\
Z_{i,j} &= D_{i,j} \cdot \sin(\Theta_{i,j}) \cdot \cos(\Phi_{i,j}), 
\end{aligned}
\end{equation}
where $D$ is the panoramic depth information.
After obtaining 3D points, the orthographic projection matrix $P_{v}$ is applied to transform 3D coordinates to 2D panoramic BEV indices $(u, v)$, which is presented in Eq.~\eqref{eq:3}, where $[\mathbf{R}|\mathbf{t}]$ is the transformation matrix.
\begin{equation}\label{eq:3}
\begin{aligned}
\left[\begin{array}{c}
x \\ y \\ z
\end{array}\right] = \mathbf{R}^{-1} \left[\begin{array}{l}
X_{i,j} \\  Y_{i,j} \\ Z_{i,j}
\end{array}\right]-\mathbf{t}, \\ 
\underbrace{\left[\begin{array}{l}
u \\ v \\ 0 \\ 1
\end{array}\right] = P_{v}\left[\begin{array}{l}
x \\ y \\ z \\ 1
\end{array}\right]}_{\text {Orthographic projection}}.
\end{aligned}
\end{equation}

\begin{table}[t]
\begin{center}
\caption{\textbf{The data statistics} of the generated 360BEV-Matterport and 360BEV-Stanford datasets. 
}
\vskip -2ex
\label{tab:datasets}
\setlength{\tabcolsep}{2mm}
\renewcommand{\arraystretch}{0.9}
\resizebox{\columnwidth}{!}{
\begin{tabular}{l|cccc}
\toprule
\textbf{Dataset} & \textbf{\#Scene} & \textbf{\#Room} & \textbf{\#Frame} & \textbf{\#Category} \\
\midrule\midrule
\texttt{train} &5 & 215 & 1,040  & 13\\
\texttt{val}   &1 & 55  & 373   & 13 \\
360BEV-Stanford &6 & 270 & 1,413 & 13 \\ \midrule
\texttt{train}& 61 & -- &7,829 & 20\\
\texttt{val}  & 7  & -- &772  & 20\\
\texttt{test} & 18 & -- &2,014 & 20\\
360BEV-Matterport & 86 & 2,030 & 10,615 & 20 \\
\bottomrule
\end{tabular}
}    
\end{center}
\vskip -3ex
\end{table}

\begin{figure}[t]
    \centering
    \begin{minipage}[t]{.99\linewidth}
        \frame{\includegraphics[width=0.99\linewidth]{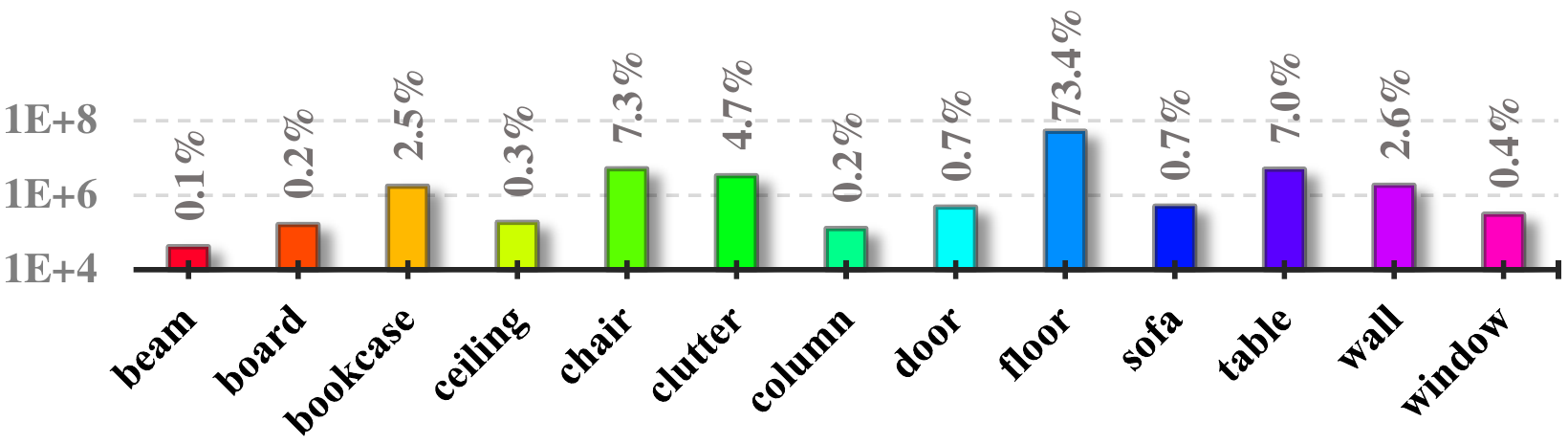}}
        \subcaption{Class distribution of 360BEV-Stanford dataset }\label{stat_s2d3d}
    \end{minipage}%
    \hfill
    \begin{minipage}[t]{.99\linewidth}
        \frame{\includegraphics[width=0.99\linewidth]{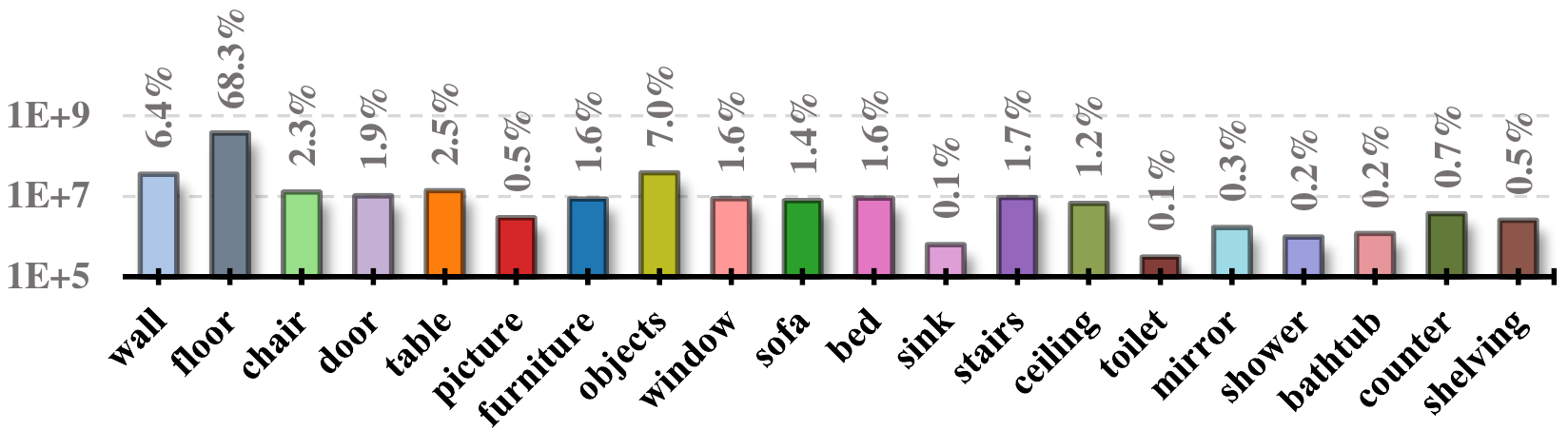}}
        \subcaption{Class distribution of 360BEV-Matterport dataset}\label{stat_mp3d}
    \end{minipage}%
    \vskip-1ex
    \caption{\textbf{Per-class pixel number (logarithmic) and frequency ($\%$) distribution} of two 360BEV datasets.}
    \label{fig:dataset_stat}
    \vskip-3ex
\end{figure}

\noindent\textbf{Dataset statistics.}
As a result, two BEV datasets for panoramic semantic mapping are obtained. The detailed data statistics of 360BEV-Stanford and 360BEV-Matterport datasets are shown in Table~\ref{tab:datasets}. While the 360BEV-Stanford dataset has $13$ classes and $1,413$ images, the 360BEV-Matterport dataset includes $20$ classes and $10,615$ samples. 
In the Matterport3D dataset~\cite{Matterport3D}, there are $40$ object categories in the dense annotation. However, many of them are relatively rare in the original dataset, \eg, \textit{TV} and \textit{beam} (${\ll}0.1\%$), which are excluded. 
Thus, 360BEV-Matterport maintains the $20$ most common object categories and merges some uncommon classes.
Besides, we further present the per-class pixel number and per-class frequency in Fig.~\ref{fig:dataset_stat}. It is worth noting that the $floor$ class has a much higher frequency on both datasets. This category is important for tasks that rely on complete maps, such as indoor navigation and is therefore also retained.

\subsection{Proposed Model: 360Mapper}~\label{sec:3_4_model}

\begin{figure*}[t]
    \centering
    \includegraphics[width=0.99\linewidth]{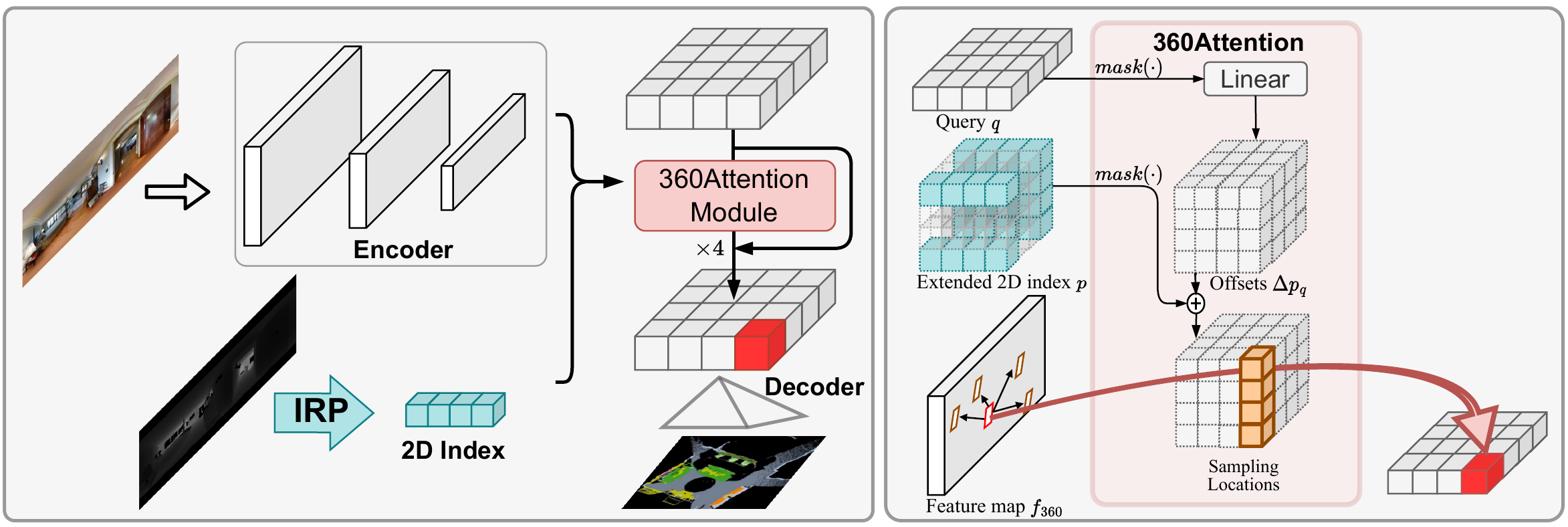}
    \begin{minipage}[t]{.5\linewidth}
        \vskip-3ex
        \subcaption{360Mapper model}\label{fig:modela}
    \end{minipage}%
    \begin{minipage}[t]{.4\linewidth}
        \vskip-3ex
        \subcaption{360Attention module}\label{fig:modelb}
    \end{minipage}%
    \vskip-1ex
    \caption{\textbf{Architecture of 360Mapper and the 360Attention module.} The 360Mapper model includes the encoder for extracting features from the front-view panoramic image, the 360Attention module for feature projection, and the decoder for parsing the projected feature to the BEV semantic map. The offsets are obtained by a linear layer and added with the 2D index that is obtained by Inverse Radial Projection (IRP), yielding the sampling locations for 360BEV feature projection.}
    \label{fig:model}
    \vskip-2ex
\end{figure*}

\noindent\textbf{Overall Architecture.} 
As shown in Fig.~\ref{fig:model}, our end-to-end {360Mapper} framework includes four steps:
(1) The transformer-based backbone extracts features from the panoramic image.
(2) The \textbf{Inverse Radial Projection (IRP)} module obtains a 2D index by projecting reference points generated by depth. (3) The \textbf{360Attention} module enhances the front-view feature by 2D index and generates offsets from BEV queries to eliminate the effects of distortion. (4) The lightweight decoder parses the projected feature map and predicts the semantic BEV map.

\noindent \textbf{Inverse Radial Projection.} 
Next, we propose a flexible projection method, the Inverse Radial Projection (IRP), for which the input of panoramic depth is included. We can easily obtain a top-view mask map by projecting from depth information. This mask map is then used to generate 3D reference points with the corresponding map height. 3D reference points are projected onto the sphere to generate 2D reference indexes, as shown in Eq.~\eqref{eq:4}, where $ID_h$ and $ID_w$ represent the index values of the 2D reference for the height and width of the feature map, respectively. The 2D reference indexes are then used to locate the corresponding feature points on the encoded front-view feature map. 
\begin{equation}\label{eq:4}
\begin{aligned}
\Phi  &= \tan^{-1} \frac{y}{x},\\
\Theta &= \tan^{-1} \left(\frac{x}{z} \cdot \frac{1}{\cos(\Phi)} \right) ,\\
ID_h &= \left\lceil\frac{H \Theta}{\pi}\right\rceil ,\\
ID_w &= \left\lceil\left(\frac{\Phi}{\pi} - \frac{1}{W}\right) \cdot \frac{W}{2}\right\rceil.
\end{aligned}
\end{equation}
Due to the distortions in the stitching process of the panorama, it is hard to project the 3D reference points exactly onto the 2D front-view plane by rotation and translation. Thus, we use the depth map to generate a map mask that better describes the shape of the map, so that the accurate projection with the mask not only makes the amount of data entering the 360Attention much smaller, which is conducive to the fast convergence of the model but also facilitates the use of sampling offsets for 360Attention.

\noindent \textbf{360Attention.}
In Fig.~\ref{fig:modelb}, the proposed 360Attention generates sampling offsets through the linear layer in an adaptive manner.
Given the BEV query $\boldsymbol{q} \in \mathbb{R}^{N \times C_{Emb}}$ as input, where $N{=}{h{\times}w}$ is the length of query, a $mask{(\cdot)}$ operation is applied on $\boldsymbol{q}$ and $\boldsymbol{p}$ to mask out irrelevant points and 2D indexes according to the mask map $M_{map}$ from IRP, which is crucial to keep $\boldsymbol{q}$ and $\boldsymbol{p}$ efficient and reducing computation of 360Attention ($\sum M_{map} {<} N$). 
The sampling offset $\Delta \boldsymbol{p}_{q,ij}$ and attention weight $\mathcal{A}_{ij}{\in}[0, 1]$ are predicted through BEV query by linear layers respectively. The adaptive sampling offsets are then added to the extended 2D index $\boldsymbol{p}$ to obtain distortion-aware sampling locations.
The 360Attention module can be denoted as:
\begin{equation}\label{eq:5}
\begin{aligned}
&\operatorname{360Attn}(\boldsymbol{q}, \boldsymbol{p}, \boldsymbol{f}_{360}) =\\ &\sum_{i = 1}^{N_{\text {head}}}\mathcal{W}_{i} \sum_{j = 1}^{N_{\text{point}}}\mathcal{A}_{ij}{\cdot} \boldsymbol{f}_{360}\left(mask\left(\boldsymbol{p}\right)+\Delta \boldsymbol{p}_{q,ij}\right),
\end{aligned}
\end{equation}
where $\boldsymbol{q}$, $\boldsymbol{p}$, and $\boldsymbol{f}_{360}$ indicate the query, the extended 2D index, and panoramic feature map, respectively. The linear layer $\mathcal{W}_{i} {\in} \mathbb{R}^{C\times (C/N_{head})}$ is specific to each attention head $i$, where $C$ is the feature dimension and $N_{head}$ is the number of heads.
The attention weight $\mathcal{A}_{ij}$ represents the importance of the sampled points $j$, where $\sum \mathcal{A}_{ij} {=}1$. 
The panoramic features $\boldsymbol{f}_{360}$ and the adaptive sampling locations $(mask(\boldsymbol{p}){+}\Delta \boldsymbol{p}_{q,ij})$ are aggregated using attention weights $\mathcal{A}_{ij}$ to produce a BEV output. Afterwards, the mask map $M_{map}$ is applied to assemble the BEV output as $\boldsymbol{q}' {\in} \mathbb{R}^{N \times C_{Emb}}$. After being added with a residual term of $\boldsymbol{q}$, the BEV result from $\boldsymbol{q}{+}\boldsymbol{q}'$ is forwarded to the next 360Attention module.

Compared to the Spatial Cross-Attention module in BEVFormer~\cite{bevformer}, the difference lies in (1) Instead of relying on multi-view features across multiple cameras, our 360Attention module is designed to directly adopt adaptive sampling offsets to extract features from a single panoramic feature map. (2) Our module gets rid of the projection of 3D reference points to different image views using the projection matrix, thus compensating for the lack of front-view perception. (3) The mask operation is applied to maintain the BEV query efficient and adaptive to front-view panoramic features by using depth information as a bridge. Through these non-trivial designs, the BEV feature map generated by 360Attention is able to effectively neutralize the effects of front-view distortion.  

\section{Experiments}
\label{sec:experiments}

\subsection{Implementation Details}
We train 360Mapper models with 4 A100 GPUs with an initial learning rate of $6\textrm{e}^{-5}$, scheduled by the step strategy over $50$ epochs. AdamW is the optimizer with epsilon $1\textrm{e}^{-8}$, weight decay is $0.01$ and batch size is $4$ on each GPU. The panoramic image size of 360FV-Matterport and Stanford2D3D~\cite{stanford2d3d} are both $512{\times}1024$. The resolution of panoramic images on both 360BEV-Stanford and 360BEV-Matterport datasets are $512{\times}1024$ as input for 360Mapper training, while the output BEV maps are set to $500{\times}500$, which correspond to a perception range of $10m{\times}10m$. Following~\cite{SMNet,chen2022trans4map}, evaluation metrics are pixel-wise accuracy~(Acc), pixel recall~(mRecall), precision~(mPrecision), and mean Intersection-over-Union~(mIoU).

\subsection{Panorama Semantic Segmentation (360FV)}

\begin{table}[!t]
\begin{center}
\caption{\textbf{Panoramic semantic segmentation (360FV)} on the Stanford2D3D dataset.
}
\vskip -1ex
\label{tab:s2d3d_front}
\setlength{\tabcolsep}{8mm}
\resizebox{\columnwidth}{!}{
    \begin{tabular}{ l | c | c}
    \toprule[1pt]
    \textbf{Method} & \textbf{Backbone} & \textbf{mIoU(\%)} \\ \midrule\midrule
    Tangent~\cite{tangent} & ResNet-101 &45.6  \\
    SegFormer~\cite{xie2021segformer} & MiT-B2  & 51.9  \\
    HoHoNet~\cite{hohonet} & ResNet-101 & 52.0  \\
    Trans4PASS~\cite{trans4pass} & MiT-B2 & 52.1 \\
    CBFC~\cite{zheng2023complementary} & ResNet-101 & 52.2 \\
    \rowcolor{gray!15} Ours & MiT-B2 & \textbf{54.3}\\
    \bottomrule[1pt]
    \end{tabular}
}
\end{center}
\vskip -3ex
\end{table}

\noindent\textbf{Results on Stanford2D3D.}
To verify the capacity to handle object deformations and image distortions, we first evaluate our method on front-view panoramic semantic segmentation. The results on the Stanford2D3D dataset are presented in Table~\ref{tab:s2d3d_front}.
All results are averaged over $3$ cross-validation folds. Thanks to the proposed 360Attention module, our 360Mapper model is better capable of handling deformations in panoramas, yielding $54.3\%$ in mIoU, with ${>}2\%$ performance gains as compared to the previous state-of-the-art Trans4PASS~\cite{trans4pass} and 	
CBFC~\cite{zheng2023complementary}. 
The promising result in front-view panoramas has initially revealed the potential of our model in extracting 360{\textdegree} front-view features, which is crucial for the BEV semantic mapping task as well.

\begin{table}[!t]\begin{center}
\caption{\textbf{Panoramic semantic segmentation (360FV)} on the \texttt{val} set of 360FV-Matterport dataset.}
\vskip -1ex
\label{tab:mp3d_front_val}
\setlength{\tabcolsep}{6mm}
\resizebox{\columnwidth}{!}{
    \begin{tabular}{ l | c | c}
    \toprule[1pt]
    \textbf{Method} & \textbf{Backbone} & \textbf{mIoU(\%)} \\ \midrule\midrule
    HoHoNet~\cite{hohonet} & ResNet-101 &   44.10\\ 
    Trans4PASS~\cite{trans4pass} & MiT-B2 & 41.91\\ 
    Trans4PASS+~\cite{trans4passplus} & MiT-B2  & 42.60\\ 
    SegFormer~\cite{xie2021segformer} & MiT-B2  &  45.53\\ 
    \rowcolor{gray!15} Ours & MiT-B2  &  \textbf{46.35}\\
    \bottomrule[1pt]
    \end{tabular}
}
\end{center}
\vskip -4ex
\end{table}

\noindent\textbf{Results on 360FV-Matterport.}
For the first time, a large-scale 360FV-Matterport is brought to the community of front-view panoramic semantic segmentation.
In Table~\ref{tab:mp3d_front_val}, four state-of-the-art methods are selected and reproduced.
Compared to the Trans4PASS~\cite{trans4pass} and Trans4PASS+~\cite{trans4passplus} models, our model has respective ${+}4.44\%$ and ${+}3.75\%$ improvements. Furthermore, our model surpasses RGB-D HoHoNet~\cite{hohonet} and SegFormer~\cite{xie2021segformer} with ${+}1.50\%$ and ${+}0.82\%$ mIoU gains. The results indicate that our model can consistently achieve state-of-the-art performance on large-scale datasets for panoramic semantic segmentation. 

\subsection{Panorama Semantic Mapping (360BEV)}
To thoroughly investigate the 360BEV task, we consistently analyze the early-, late-, and intermediate projections, as well as compare their state-of-the-art methods in both 360BEV benchmarks.

\noindent\textbf{Results on 360BEV-Stanford.} 
In Table~\ref{tab:s2d3d_topdown}, to study the Early projection mode, SegFormer~\cite{xie2021segformer} and SegNeXt~\cite{guo2022segnext} with different backbones, are selected, which merely reach unsatisfactory results. The results indicate that the pre-projected RGB maintains less rich spatial and visual information of front-view images. Using Late projection, SegFormer with the same MiT-B2 backbone achieves $18.65\%$ mIoU and surpasses the one using Early projection, still yielding sub-optimal semantic mapping results. Interestingly, all methods using Intermediate projection obtain more than $30\%$ mIoU. While using the same MiT-B2 backbone, our proposed 360Mapper achieves $45.78\%$ with ${+}9.70\%$ gains compared to the baseline Trans4Map~\cite{chen2022trans4map}. Further, our efficient model~(MiT-B0) outperforms Trans4Map~(MiT-B4) with ${+}05.73\%$ mIoU gains. With a stronger CNN backbone MSCA-B from SegNeXt~\cite{guo2022segnext}, our method reaches the best score with $46.44\%$ in mIoU, which indicates 360Mapper is flexible to both CNN- and Transformer-based backbones.

\begin{table}[!t]
\begin{center}
\caption{\textbf{Panoramic semantic mapping (360BEV)} on the 360BEV-Stanford dataset.}
\vskip -1ex
\label{tab:s2d3d_topdown}
\setlength{\tabcolsep}{1mm}
\resizebox{\columnwidth}{!}{
\begin{tabular}{ l c c c c l }
\toprule

\textbf{Method} & \textbf{Backbone} & \textbf{Acc} & \textbf{mRecall} & \textbf{mPrecision} & \textbf{mIoU}\\
\midrule\midrule
\multicolumn{6}{c}{\textit{(1) Early projection: Proj.${\rightarrow}$Enc.${\rightarrow}$Seg.}} \\\midrule

SegFormer~\cite{xie2021segformer} & MiT-B2  & 71.69  & 20.82 & 26.34 & 14.15 \\
SegNeXt~\cite{guo2022segnext}  & MSCA-B &79.77 & 34.13 & 47.39 & 25.85\\
\midrule
\multicolumn{6}{c}{\textit{(2) Late projection: Enc.${\rightarrow}$Seg.${\rightarrow}$Proj.}} \\\midrule
HoHoNet~\cite{hohonet}  & ResNet101 & 70.01 & 31.62 & 30.46 & 18.49\\
Trans4PASS~\cite{trans4pass}  & MiT-B2  &65.73  & 31.08 & 33.15 & 17.86 \\
Trans4PASS+~\cite{trans4passplus}  & MiT-B2 & 66.11 & 38.06 & 34.14  &20.44 \\
SegFormer~\cite{xie2021segformer}  & MiT-B2 &70.50  & 30.97 & 30.65 & 18.65\\

\midrule
\multicolumn{6}{c}{\textit{(3) Intermediate projection: Enc.${\rightarrow}$Proj.${\rightarrow}$Seg.}} \\\midrule

BEVFormer~\cite{bevformer} &  MiT-B2 & 85.50 & 40.22 & 51.71 & 31.69\\
Trans4Map~\cite{chen2022trans4map}& MiT-B0 & 86.41 & 40.45 & 57.47 & 32.26 \\
Trans4Map~\cite{chen2022trans4map} &  MiT-B2 & 86.53 & 45.28 & 62.61  & 36.08\\
Trans4Map~\cite{chen2022trans4map} & MiT-B4  & 86.99 & 46.18 & 58.19  & 36.69   \\
\rowcolor{gray!20} Ours & MiT-B0 & 92.07 & 50.14 & 65.37 & 42.42 ~\gbf{+10.16}\\ 
\rowcolor{gray!20} Ours & MiT-B2 & \textbf{92.80}  & 53.56  &67.72  &45.78  ~\gbf{+09.70} \\ 
\rowcolor{gray!10} Ours & MSCA-B & 92.67  & \textbf{55.02}  &\textbf{68.02}  &\textbf{46.44}  \\ 
\bottomrule
\end{tabular}
}
\end{center}
\vskip -3ex
\end{table}
\begin{table}[!t]
\begin{center}
\caption{ \textbf{Panoramic semantic mapping (360BEV)} on the \textit{val} set of 360BEV-Matterport dataset.}
\vskip -1ex
\label{tab:mp3d_topdown_val}
\setlength{\tabcolsep}{1mm}
\resizebox{\columnwidth}{!}{
\begin{tabular}{ l c c c c l }
\toprule
\textbf{Method} & \textbf{Backbone} & \textbf{Acc} & \textbf{mRecall} & \textbf{mPrecision} & \textbf{mIoU}\\
\midrule\midrule
\multicolumn{6}{c}{\textit{(1) Early projection: Proj.${\rightarrow}$Enc.${\rightarrow}$Seg.}} \\\midrule
SegFormer~\cite{xie2021segformer} & MiT-B2  & 68.12  & 41.33 & 45.25 & 29.22 \\
SegNeXt~\cite{guo2022segnext}  & MSCA-B & 68.53 & 42.13 & 46.12 & 30.01\\

\midrule
\multicolumn{6}{c}{\textit{(2) Late projection: Enc.${\rightarrow}$Seg.${\rightarrow}$Proj.}} \\\midrule
HoHoNet~\cite{hohonet}  & ResNet101 & 62.84 & 38.99 & 44.22 & 26.21\\
Trans4PASS~\cite{trans4pass}  & MiT-B2  & 55.99 & 29.59 & 40.91 & 20.07\\
Trans4PASS+~\cite{trans4passplus}  & MiT-B2 & 57.89 & 32.75 & 40.93 & 21.58\\
SegFormer~\cite{xie2021segformer}  & MiT-B2 & 62.98 & 41.84 & 45.30 & 27.78\\
\midrule
\multicolumn{6}{c}{\textit{(3) Intermediate projection: Enc.${\rightarrow}$Proj.${\rightarrow}$Seg.}} \\\midrule

BEVFormer~\cite{bevformer} &  MiT-B2 &72.99  &43.61  & 51.70 & 32.51\\
Trans4Map~\cite{chen2022trans4map}& MiT-B0 & 70.19 & 44.31& 50.39 & 31.92   \\
Trans4Map~\cite{chen2022trans4map} &  MiT-B2 & 73.28 & 51.60 & 53.02 & 36.72  \\
Trans4Map~\cite{chen2022trans4map} & MiT-B4  &73.51  & 50.78 & 56.67 & 38.04   \\
\rowcolor{gray!20} Ours & MiT-B0 & 75.44 & 48.80 & 56.01 & 36.98~\gbf{+5.06} \\ 
\rowcolor{gray!20} Ours & MiT-B2 &78.80 &59.54 &59.97  & 44.32~\gbf{+7.60} \\ 
\rowcolor{gray!10} Ours & MSCA-B &\textbf{78.93} & \textbf{60.51} & \textbf{62.83} & \textbf{46.31} \\ 
\bottomrule
\end{tabular}
}
\end{center}
\vskip -3ex
\end{table} 
\noindent\textbf{Results on 360BEV-Matterport.}
In Table~\ref{tab:mp3d_topdown_val}, we further present the results on the 360BEV-Matterport dataset. SegFormer~\cite{xie2021segformer} and SegNeXt~\cite{guo2022segnext} adopt Early projection and show better performance than the Late projection ones.
The reason for this is Late projection methods are constrained by their lower performance in front-view semantic segmentation, which affects the projected BEV semantic maps.
In contrast, using Intermediate projection, our 360Mapper models based on two different model scales, \ie, MiT-B0 and MiT-B2, show overall promising performance with $36.98\%$ and $44.32\%$ in mIoU, respectively. Compared to the previous state-of-the-art Trans4Map~\cite{chen2022trans4map} (MiT-B2), our approach with MiT-B2 has improvements by ${+}5.52\%$ in accuracy, ${+}7.94\%$ in mRecall, ${+}6.95\%$ in mPrecision, and ${+}7.60\%$ in mIoU. Surprisingly, our 360Mapper with MiT-B2 outperforms Trans4Map with MiT-B4 with ${+}6.28\%$ in mIoU. Besides, to compare multi-view methods, we reproduce BEVFormer~\cite{bevformer} by using a single panorama instead of six views of pinhole cameras. Our 360Mapper outperforms BEVFormer (MiT-B2) with ${+}11.81\%$ mIoU. Furthermore, we verify the flexibility of 360Mapper by using a CNN-based MSCA-B backbone~\cite{guo2022segnext}, which obtains the highest mIoU score with $46.31\%$. All results are in line with our observation that Intermediate projection can preserve dense visual cues and long-range information from front-view panoramas, and deliver more valuable context for BEV semantic mapping, leading to this superiority of 360Mapper, as compared to the other paradigms.

\begin{figure*}[t]
    \centering
    \subfloat[360BEV-Stanford\label{subfig-1:class_iou_s2d3d}]{%
    \includegraphics[trim={2 2 2 5},clip,width=0.40\textwidth]{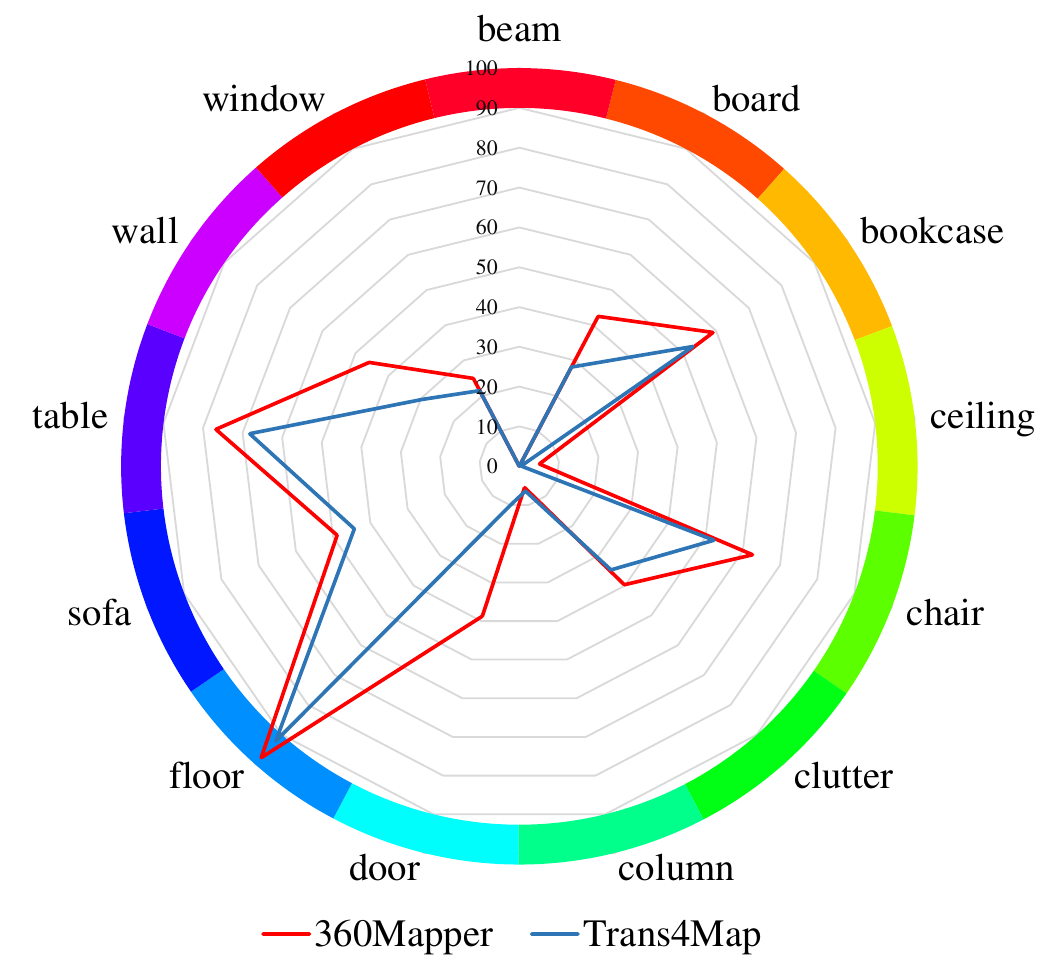}
    }
    \subfloat[360BEV-Matterport\label{subfig-2:class_iou_mp3d}]{%
      \includegraphics[trim={2 2 2 5},clip, width=0.55\textwidth]{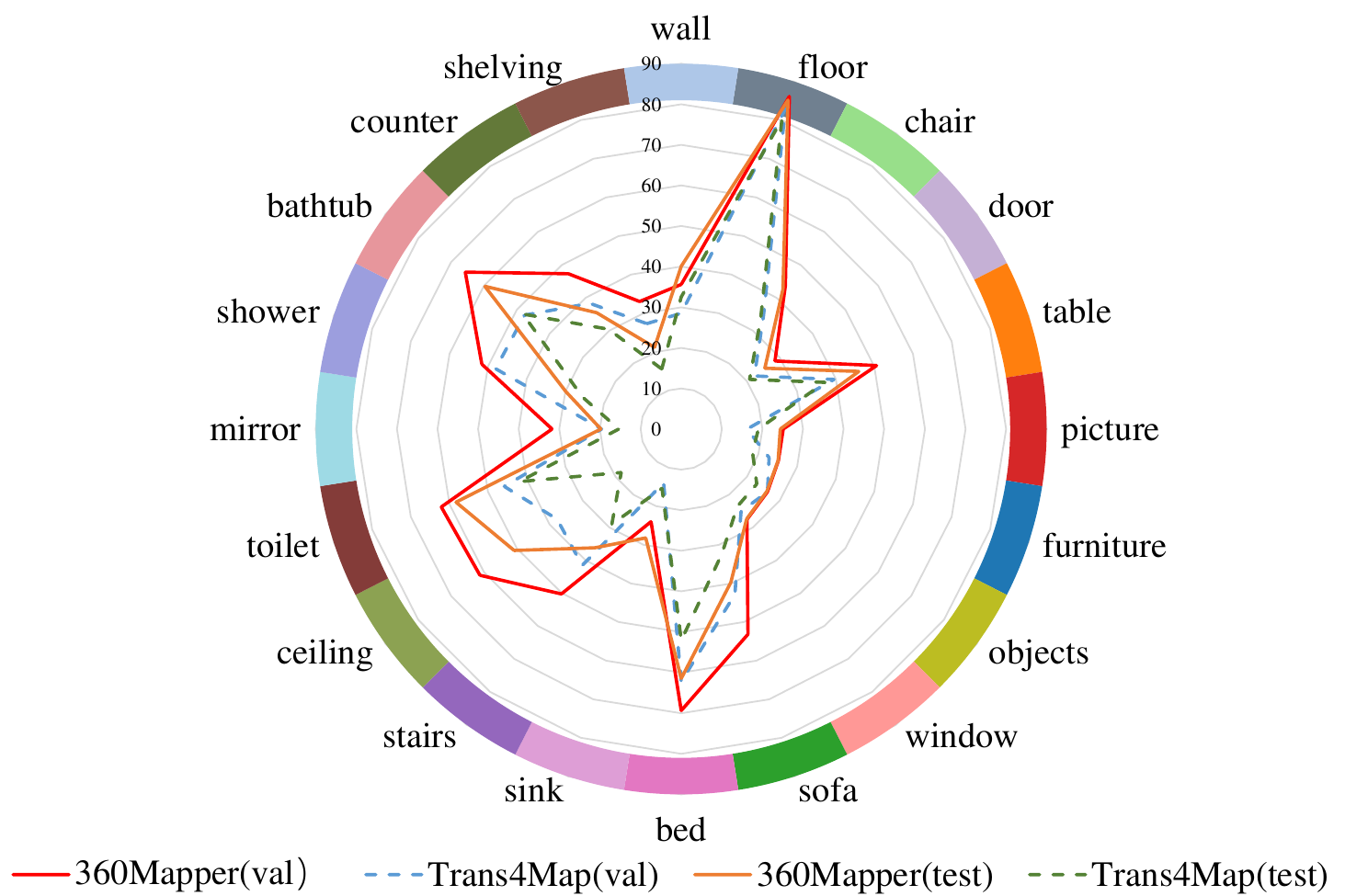}
    }
    \vskip-1ex
    \caption{\textbf{Distribution of per-class semantic mapping results} (per-class IoU in $\%$) on the 360BEV-Stanford and the 360BEV-Matterport datasets. Compared to the baseline model Trans4Map~\cite{chen2022trans4map}, our 360Mapper models achieve overall better 360BEV results.}
    \label{fig:360bev_s2d3d_per_class}
    \vskip-1ex
\end{figure*}

\noindent\textbf{Per-class Results.}
To study the per-class performance on both 360BEV datasets, we present the comparison results in Fig.~\ref{fig:360bev_s2d3d_per_class}. For comparison, both the baseline Trans4Map and our 360Mapper model are based on the same backbone, \ie, MiT-B2. On the 360BEV-Stanford dataset (Fig.~\ref{subfig-1:class_iou_s2d3d}), our 360Mapper model has significant gains on most of categories, such as \textit{board}~(${>}14\%$), \textit{wall}~(${>16\%}$), \textit{door}~(${>}28\%$), etc. On the 360BEV-Matterport dataset (Fig.~\ref{subfig-2:class_iou_mp3d}), it is readily apparent that our model can better recognize the \textit{chairs} and \textit{tables}, yielding ${>}6\%$ IoU gains compared to Trans4Map~\cite{chen2022trans4map}.
On the test set of the 360BEV-Matterport dataset, our 360Mapper obtains IoU gains with ${>}12\%$ and ${>}15\%$ on the \textit{sink} and \textit{toilet} classes, as compared to Trans4Map. Overall, the consistent improvements on both datasets show the superiority of our 360Mapper on panoramic semantic mapping. 

\begin{table}[!t]\begin{center}
\caption{\textbf{Analysis of offset mechanisms in 360Attention and backbone variants} on 360BEV-Matterport dataset.}
\vskip -1ex
\label{tab:analysis}
\setlength{\tabcolsep}{1mm}
\renewcommand{\arraystretch}{1.2}
\resizebox{\columnwidth}{!}{
    \begin{tabular}{ l l | c | c | l}
    \toprule[1pt]
    \textbf{Methods} & \textbf{Backbone} & \textbf{\#Param} & \textbf{FLOPs} & \textbf{mIoU} \\ \midrule\midrule
    
    \circled{1} Ours (360Attention offset) & MiT-B0 & 04.60M  & 248.57G & 36.98     \\
    \circled{2} Ours (360Attention offset) & MiT-B2 & 26.30M & 283.94G & 44.32 \\ 
    \circled{3} Ours (360Attention offset) & MiT-B4  & 62.91M & 341.34G &  \textbf{45.53}    \\  \midrule
    \circled{4} Ours (Multi-scale offset) & MiT-B2  & 26.43M  &284.17G &43.65~\obf{-0.67}   \\
    \circled{5} Ours (Fixed-range offset) & MiT-B2  & 26.30M & 283.44G &  43.28~\obf{-1.04}\\
    \circled{6} Ours (Separate offset) & MiT-B2 & 26.19M & 279.18G &  42.82~\obf{-1.50}\\\midrule
    \circled{7} Ours (360Attention offset) & MSCA-B  & 27.69M &274.59G & \textbf{46.31}~\gbf{+1.99} \\ 

    \bottomrule
    \end{tabular}
}
\end{center}
\vskip -4ex
\end{table}
\noindent\textbf{Analysis of 360Attention.}
To better understand 360Attention, we further conduct an analysis of the offset mechanisms in 360Attention and the backbone selection, in Table~\ref{tab:analysis}.
First, in \circled{1}\circled{2}\circled{3}, we select three model scales, \ie, MiT-B0, MiT-B2, and MiT-B4, to verify the effect of model capacity in 360Attention. The three models obtain good performance, showing that 360Attention has positive effects in different model scales. 
Besides, different offset schemes are compared among \circled{2}\circled{4}\circled{5}\circled{6}, which are deformable, multi-scale, fixed-range, and separate offset.
All of them have the same MiT-B2 backbone.
Here, \circled{2} shows the superiority of deformable offset which has a better performance ($44.32\%$). However, these comparable results prove that our 360Attention design is robust to offset mechanisms.
Further, to analyze the effect of backbone selection, we choose transformer-based MiT-B2~\cite{xie2021segformer} and CNN-based MSCA-B~\cite{guo2022segnext} as in \circled{2}\circled{7}. A stronger backbone~\cite{guo2022segnext} shows a further improvement of mIoU (${+}1.99\%$), which shows the flexibility of our approach regarding the backbone variants.

\begin{figure}[t]
    \footnotesize
    \setlength\tabcolsep{1pt}
    {
    \newcolumntype{P}[1]{>{\centering\arraybackslash}p{#1}}
    \begin{tabular}{@{}*{10}{P{0.089\columnwidth}}@{}}
    {\cellcolor[rgb]{0.68,0.78,0.91}}\textcolor{white}{wall} 
    &{\cellcolor[rgb]{0.44,0.50,0.56}}\textcolor{white}{floor}
    &{\cellcolor[rgb]{0.60,0.87,0.54}}\textcolor{black}{chair}
    &{\cellcolor[rgb]{0.77,0.69,0.84}}\textcolor{white}{door}
    &{\cellcolor[rgb]{1.00,0.50,0.05}}\textcolor{white}{table} 
    &{\cellcolor[rgb]{0.84,0.15,0.16}}\textcolor{white}{pictu.} 
    &{\cellcolor[rgb]{0.12,0.47,0.71}}\textcolor{white}{furni.}
    &{\cellcolor[rgb]{0.74,0.74,0.13}}\textcolor{black}{objec.}
    &{\cellcolor[rgb]{1.00,0.60,0.59}}\textcolor{black}{windo.}
    &{\cellcolor[rgb]{0.17,0.63,0.17}}\textcolor{white}{sofa} \\
    {\cellcolor[rgb]{0.89,0.47,0.76}}\textcolor{black}{bed}
    & {\cellcolor[rgb]{0.87,0.62,0.84}}\textcolor{black}{sink}
    &{\cellcolor[rgb]{0.58,0.40,0.74}}\textcolor{white}{stairs} 
    &{\cellcolor[rgb]{0.55,0.64,0.32}}\textcolor{white}{ceil.} 
    &{\cellcolor[rgb]{0.52,0.24,0.22}}\textcolor{white}{toilet} 
    &{\cellcolor[rgb]{0.62,0.85,0.90}}\textcolor{black}{mirror} 
    &{\cellcolor[rgb]{0.61,0.62,0.87}}\textcolor{black}{show.}
    &{\cellcolor[rgb]{0.91,0.59,0.61}}\textcolor{black}{batht.}
    &{\cellcolor[rgb]{0.39,0.47,0.22}}\textcolor{white}{count.} 
    &{\cellcolor[rgb]{0.55,0.34,0.29}}\textcolor{white}{shelv.} \\
    \end{tabular}
    }
    \centering
    \begin{tabular}{c c c c}
        \vspace{1pt}
        \raisebox{-0.5\height}{\includegraphics[width=0.18\textwidth]{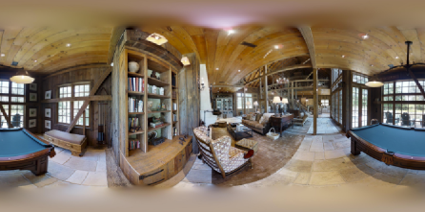}} &
        \raisebox{-0.5\height}{\includegraphics[width=0.09\textwidth]{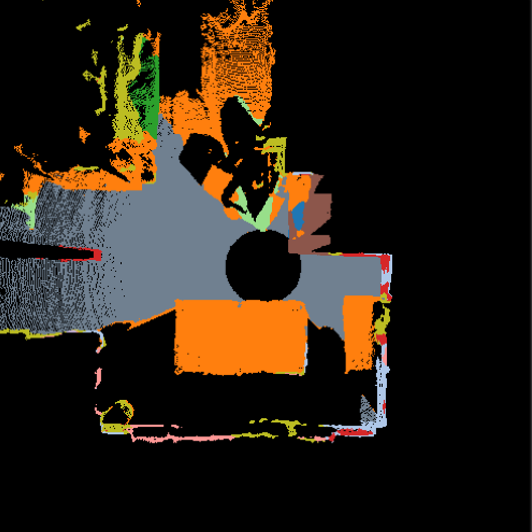}} &
        \raisebox{-0.5\height}{\includegraphics[width=0.09\textwidth]{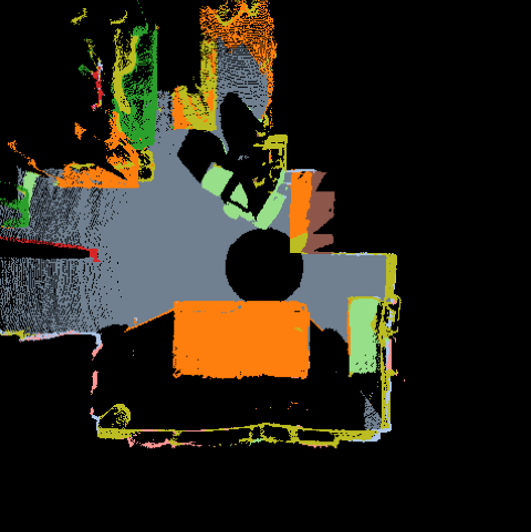}} &
        \raisebox{-0.5\height}{\includegraphics[width=0.09\textwidth]{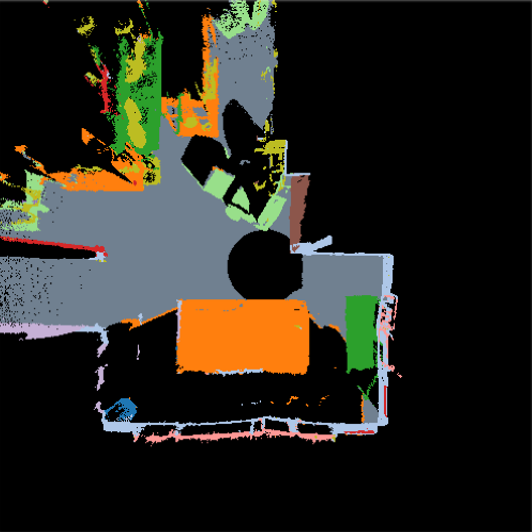}}\\

        \vspace{1pt}

        \raisebox{-0.5\height}{\includegraphics[width=0.18\textwidth]{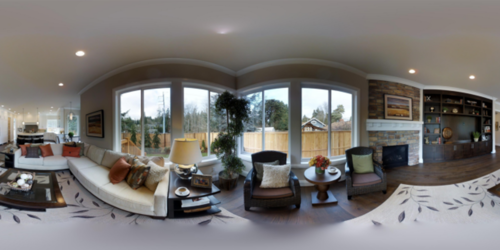}} &
        \raisebox{-0.5\height}{\includegraphics[width=0.09\textwidth]{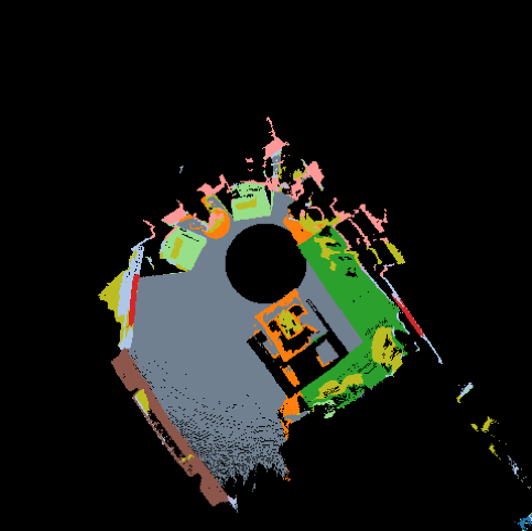}} &
        \raisebox{-0.5\height}{\includegraphics[width=0.09\textwidth]{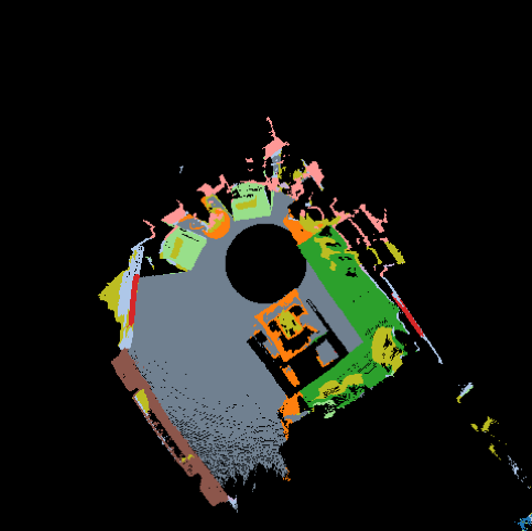}} &
        \raisebox{-0.5\height}{\includegraphics[width=0.09\textwidth]{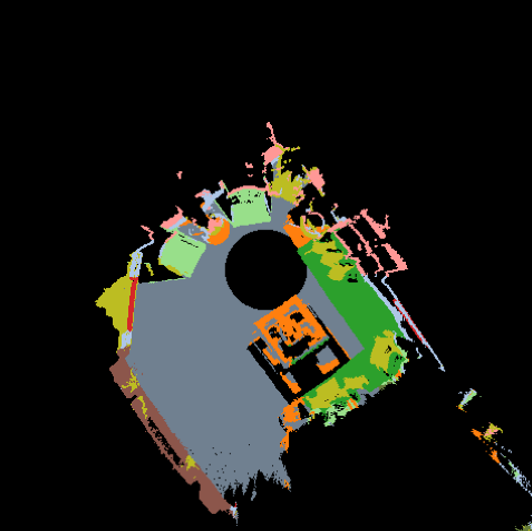}}\\
        \vspace{1pt}

        \raisebox{-0.5\height}{\includegraphics[width=0.18\textwidth]{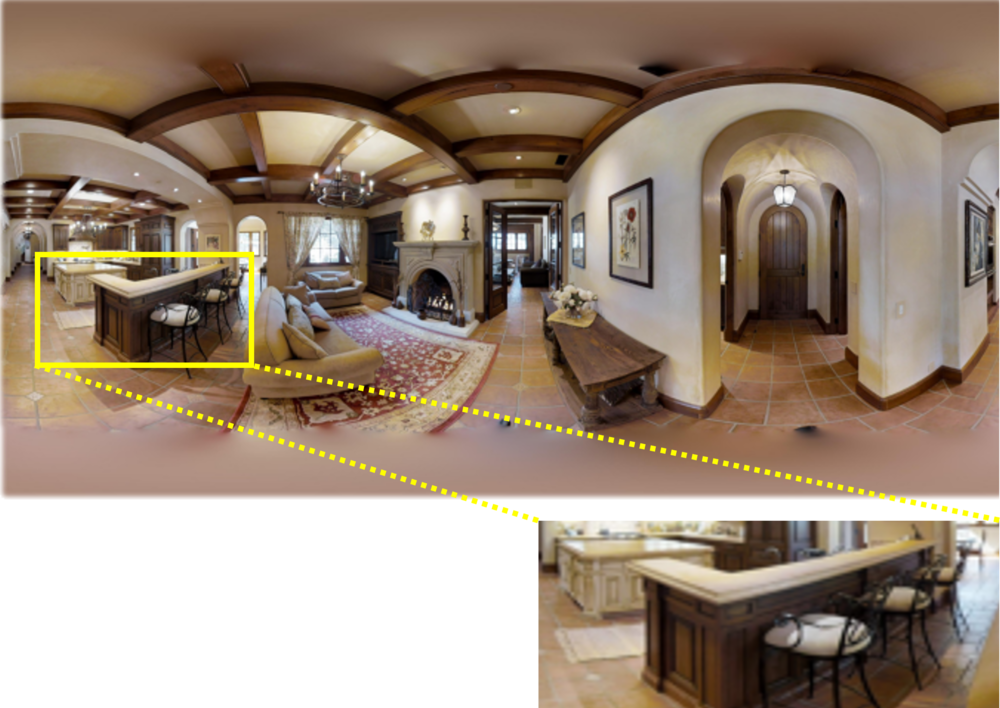}} &
        \raisebox{-0.5\height}{\includegraphics[width=0.09\textwidth]{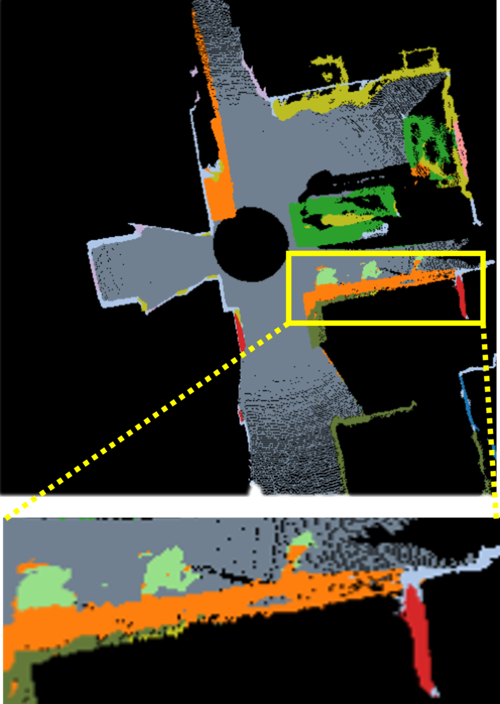}} &
        \raisebox{-0.5\height}{\includegraphics[width=0.09\textwidth]{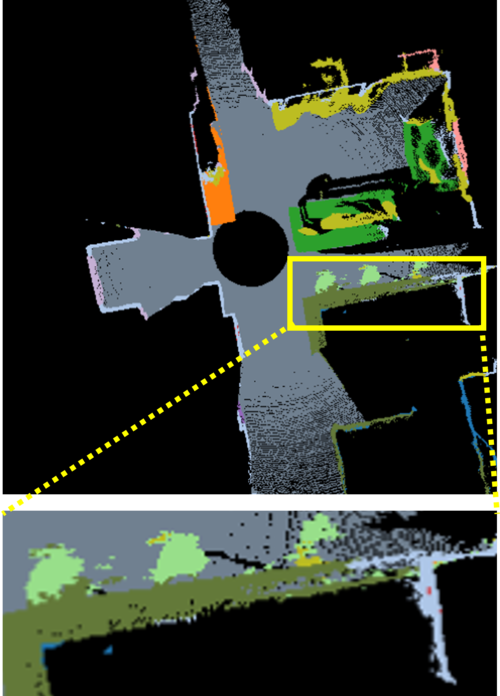}} &
        \raisebox{-0.5\height}{\includegraphics[width=0.09\textwidth]{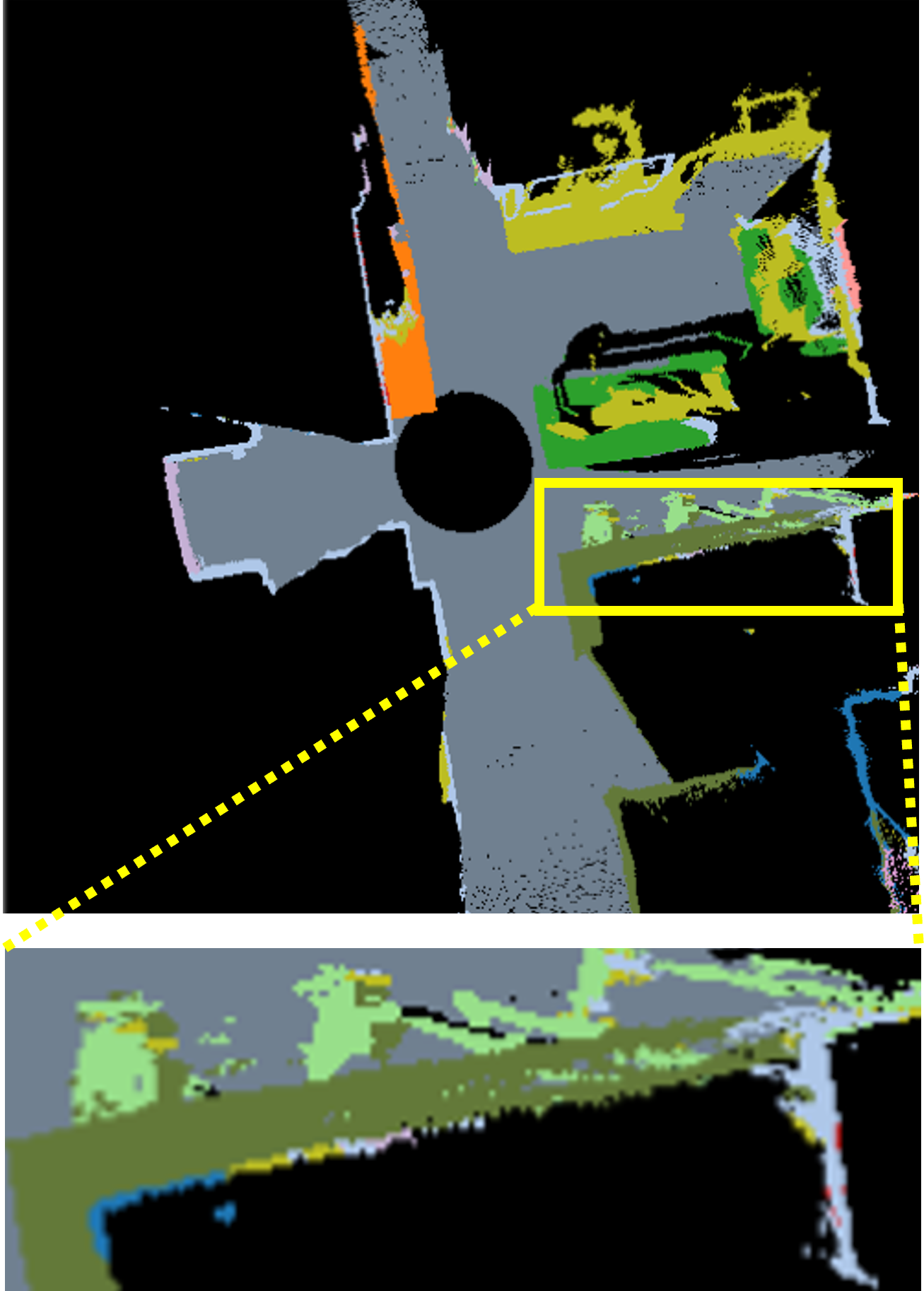}}\\
        \noalign{\vskip 2mm}   
        Input&Baseline&360Mapper&Ground Truth
    \end{tabular}
    \caption{\textbf{Qualitative analysis} on the 360BEV-Matterport dataset. Black regions are \texttt{void}. Zoom in for a better view.}
    \label{fig:vis}
    \vskip -3ex
\end{figure}

\subsection{Qualitative Analysis}
To analyze the predicted semantic maps, we visualize the results from the validation set of the 360BEV-Matterport dataset. In Fig.~\ref{fig:vis}, from left to right are input images, results of baseline~\cite{chen2022trans4map}, results of our 360Mapper, and ground truth. Thanks to the IRP projection and 360Attention, the segmentation results of 360Mapper are much better.
In the first scene in Fig.~\ref{fig:vis}, 360Mapper is able to successfully classify \textit{chairs}, while the baseline model fails, predicting several \textit{tables} and misclassifying the distant ground as another \textit{table}. In the second scene, the segmentation of the \textit{tables} derived by the baseline is incomplete. Furthermore, in the last zoomed-in scene, 360Mapper provides accurate semantic maps, such as in \textit{counter}, \textit{chair}, and \textit{wall} categories, whereas the baseline Trans4Map~\cite{chen2022trans4map} misclassifies them as \textit{tables} and \textit{doors}. Based on the qualitative analysis, our 360Mapper can effectively handle object deformations and image distortions, yielding better BEV semantic maps.

\section{Conclusion}
\label{sec:conclusion}
In this paper, we introduce 360BEV, a novel task to conduct panoramic semantic mapping in indoor environments, \ie, from a single panoramic image to a holistic BEV semantic map.
To enable this, we present 360BEV-Matterport and 360BEV-Stanford, extending off-the-shelf datasets for the presented 360BEV task. 
We revisit existing transformation paradigms and propose 360Mapper, a novel end-to-end architecture specifically designed for panoramic semantic mapping.
As a consequence, 360Mapper outperforms state-of-the-art counterparts by clear margins.

\noindent \footnotesize{\\ \textbf{Acknowledgement.}
This work was supported
in part by the ``KIT Future Fields'' project, 
in part by the Ministry of Science, Research and the Arts of Baden-Wurttemberg (MWK) through the Cooperative Graduate School Accessibility through AI-based Assistive Technology (KATE) under Grant BW6-03, in part by the Federal Ministry of Education and Research (BMBF) through a fellowship within the IFI program of the German Academic Exchange Service (DAAD), and in part by Hangzhou SurImage Technology Company Ltd. We thank HoreKA@KIT, HAICORE@KIT, and bwHPC supercomputer partitions.
}

\clearpage
{\small
\bibliographystyle{ieee_fullname}
\bibliography{main}

\begin{thebibliography}{10}\itemsep=-1pt

\bibitem{albanis2021pano3d}
Georgios Albanis, Nikolaos Zioulis, Petros Drakoulis, Vasileios Gkitsas,
  Vladimiros Sterzentsenko, Federico Alvarez, Dimitrios Zarpalas, and Petros
  Daras.
\newblock {Pano3D:} {A} holistic benchmark and a solid baseline for
  360{\textdegree} depth estimation.
\newblock In {\em CVPRW}, 2021.

\bibitem{anderson2018vision}
Peter Anderson, Qi Wu, Damien Teney, Jake Bruce, Mark Johnson, Niko
  S{\"u}nderhauf, Ian Reid, Stephen Gould, and Anton Van Den~Hengel.
\newblock Vision-and-language navigation: Interpreting visually-grounded
  navigation instructions in real environments.
\newblock In {\em CVPR}, 2018.

\bibitem{stanford2d3d}
Iro Armeni, Sasha Sax, Amir~R. Zamir, and Silvio Savarese.
\newblock Joint {2D-3D-semantic} data for indoor scene understanding.
\newblock {\em arXiv preprint arXiv:1702.01105}, 2017.

\bibitem{SMNet}
Vincent Cartillier, Zhile Ren, Neha Jain, Stefan Lee, Irfan Essa, and Dhruv
  Batra.
\newblock Semantic {MapNet}: {Building} allocentric semantic maps and
  representations from egocentric views.
\newblock In {\em AAAI}, 2021.

\bibitem{Matterport3D}
Angel Chang, Angela Dai, Thomas Funkhouser, Maciej Halber, Matthias Niessner,
  Manolis Savva, Shuran Song, Andy Zeng, and Yinda Zhang.
\newblock {Matterport3D:} {Learning} from {RGB-D} data in indoor environments.
\newblock In {\em 3DV}, 2017.

\bibitem{chen2022trans4map}
Chang Chen, Jiaming Zhang, Kailun Yang, Kunyu Peng, and Rainer Stiefelhagen.
\newblock {Trans4Map:} {Revisiting} holistic bird's-eye-view mapping from
  egocentric images to allocentric semantics with vision transformers.
\newblock In {\em WACV}, 2023.

\bibitem{gauge_equivariant}
Taco Cohen, Maurice Weiler, Berkay Kicanaoglu, and Max Welling.
\newblock Gauge equivariant convolutional networks and the icosahedral {CNN}.
\newblock In {\em ICML}, 2019.

\bibitem{tangent}
Marc Eder, Mykhailo Shvets, John Lim, and Jan-Michael Frahm.
\newblock Tangent images for mitigating spherical distortion.
\newblock In {\em CVPR}, 2020.

\bibitem{esteves2020spin_weighted_spherical}
Carlos Esteves, Ameesh Makadia, and Kostas Daniilidis.
\newblock Spin-weighted spherical {CNNs}.
\newblock In {\em NeurIPS}, 2020.

\bibitem{feng2020deep}
Di Feng, Christian Haase-Sch{\"u}tz, Lars Rosenbaum, Heinz Hertlein, Claudius
  Glaeser, Fabian Timm, Werner Wiesbeck, and Klaus Dietmayer.
\newblock Deep multi-modal object detection and semantic segmentation for
  autonomous driving: Datasets, methods, and challenges.
\newblock {\em T-ITS}, 2020.

\bibitem{grinvald2019volumetric}
Margarita Grinvald, Fadri Furrer, Tonci Novkovic, Jen~Jen Chung, Cesar Cadena,
  Roland Siegwart, and Juan Nieto.
\newblock Volumetric instance-aware semantic mapping and {3D} object discovery.
\newblock {\em RA-L}, 2019.

\bibitem{guo2022segnext}
Meng-Hao Guo, Cheng-Ze Lu, Qibin Hou, Zhengning Liu, Ming-Ming Cheng, and
  Shi-Min Hu.
\newblock {SegNeXt:} {Rethinking} convolutional attention design for semantic
  segmentation.
\newblock In {\em NeurIPS}, 2022.

\bibitem{janai2020cv4auto}
Joel Janai, Fatma G{\"u}ney, Aseem Behl, and Andreas Geiger.
\newblock Computer vision for autonomous vehicles: Problems, datasets and state
  of the art.
\newblock {\em Foundations and Trends{\textregistered} in Computer Graphics and
  Vision}, 2020.

\bibitem{spherical_unstructured_grids}
Chiyu~Max Jiang, Jingwei Huang, Karthik Kashinath, Prabhat, Philip Marcus, and
  Matthias Nie{\ss}ner.
\newblock Spherical {CNNs} on unstructured grids.
\newblock In {\em ICLR}, 2019.

\bibitem{lee2020pillarflow}
Kuan-Hui Lee, Matthew Kliemann, Adrien Gaidon, Jie Li, Chao Fang, Sudeep
  Pillai, and Wolfram Burgard.
\newblock {PillarFlow:} {End-to-end} birds-eye-view flow estimation for
  autonomous driving.
\newblock In {\em IROS}, 2020.

\bibitem{InteriorNet18}
Wenbin Li, Sajad Saeedi, John McCormac, Ronald Clark, Dimos Tzoumanikas, Qing
  Ye, Yuzhong Huang, Rui Tang, and Stefan Leutenegger.
\newblock {InteriorNet:} {Mega-scale} multi-sensor photo-realistic indoor
  scenes dataset.
\newblock In {\em BMVC}, 2018.

\bibitem{bevformer}
Zhiqi Li, Wenhai Wang, Hongyang Li, Enze Xie, Chonghao Sima, Tong Lu, Yu Qiao,
  and Jifeng Dai.
\newblock {BEVFormer:} {Learning} bird’s-eye-view representation from
  multi-camera images via spatiotemporal transformers.
\newblock In {\em ECCV}, 2022.

\bibitem{liu2022petr}
Yingfei Liu, Tiancai Wang, Xiangyu Zhang, and Jian Sun.
\newblock {PETR:} {Position} embedding transformation for multi-view {3D}
  object detection.
\newblock In {\em ECCV}, 2022.

\bibitem{luo2021self}
Chenxu Luo, Xiaodong Yang, and Alan Yuille.
\newblock Self-supervised pillar motion learning for autonomous driving.
\newblock In {\em CVPR}, 2021.

\bibitem{mattyus2015enhancing}
Gellert Mattyus, Shenlong Wang, Sanja Fidler, and Raquel Urtasun.
\newblock Enhancing road maps by parsing aerial images around the world.
\newblock In {\em ICCV}, 2015.

\bibitem{maturana2018real}
Daniel Maturana, Po-Wei Chou, Masashi Uenoyama, and Sebastian Scherer.
\newblock Real-time semantic mapping for autonomous off-road navigation.
\newblock In {\em FSR}, 2018.

\bibitem{mivcuvslik2009semantic}
Branislav Mi{\v{c}}u{\v{s}}{\'l}{\'\i}k and Jana Ko{\v{s}}eck{\'a}.
\newblock Semantic segmentation of street scenes by superpixel co-occurrence
  and {3D} geometry.
\newblock In {\em ICCVW}, 2009.

\bibitem{pan2020cross}
Bowen Pan, Jiankai Sun, Ho~Yin~Tiga Leung, Alex Andonian, and Bolei Zhou.
\newblock Cross-view semantic segmentation for sensing surroundings.
\newblock {\em RA-L}, 2020.

\bibitem{peng2023bevsegformer}
Lang Peng, Zhirong Chen, Zhangjie Fu, Pengpeng Liang, and Erkang Cheng.
\newblock {BEVSegFormer:} {Bird's} eye view semantic segmentation from
  arbitrary camera rigs.
\newblock In {\em WACV}, 2023.

\bibitem{ran2021rs}
Teng Ran, Liang Yuan, Jianbo Zhang, Dingxin Tang, and Li He.
\newblock {RS-SLAM:} {A} robust semantic {SLAM} in dynamic environments based
  on {RGB-D} sensor.
\newblock {\em IEEE Sensors Journal}, 2021.

\bibitem{savva2019habitat}
Manolis Savva, Jitendra Malik, Devi Parikh, Dhruv Batra, Abhishek Kadian,
  Oleksandr Maksymets, Yili Zhao, Erik Wijmans, Bhavana Jain, Julian Straub,
  Jia Liu, and Vladlen Koltun.
\newblock Habitat: A platform for embodied {AI} research.
\newblock In {\em CVPR}, 2019.

\bibitem{sengupta2012automatic}
Sunando Sengupta, Paul Sturgess, Lubor Ladicky, and Philip H.~S. Torr.
\newblock Automatic dense visual semantic mapping from street-level imagery.
\newblock In {\em IROS}, 2012.

\bibitem{singh2018self}
Suriya Singh, Anil Batra, Guan Pang, Lorenzo Torresani, Saikat Basu, Manohar
  Paluri, and C.~V. Jawahar.
\newblock Self-supervised feature learning for semantic segmentation of
  overhead imagery.
\newblock In {\em BMVC}, 2018.

\bibitem{hohonet}
Cheng Sun, Min Sun, and Hwann-Tzong Chen.
\newblock {HoHoNet:} 360 indoor holistic understanding with latent horizontal
  features.
\newblock In {\em CVPR}, 2021.

\bibitem{distortion_aware}
Keisuke Tateno, Nassir Navab, and Federico Tombari.
\newblock Distortion-aware convolutional filters for dense prediction in
  panoramic images.
\newblock In {\em ECCV}, 2018.

\bibitem{wang2021hrnet}
Jingdong Wang, Ke Sun, Tianheng Cheng, Borui Jiang, Chaorui Deng, Yang Zhao,
  Dong Liu, Yadong Mu, Mingkui Tan, Xinggang Wang, Wenyu Liu, and Bin Xiao.
\newblock Deep high-resolution representation learning for visual recognition.
\newblock {\em TPAMI}, 2021.

\bibitem{wijmans2019embodied}
Erik Wijmans, Samyak Datta, Oleksandr Maksymets, Abhishek Das, Georgia
  Gkioxari, Stefan Lee, Irfan Essa, Devi Parikh, and Dhruv Batra.
\newblock Embodied question answering in photorealistic environments with point
  cloud perception.
\newblock In {\em CVPR}, 2019.

\bibitem{xia2018gibson}
Fei Xia, Amir~R Zamir, Zhiyang He, Alexander Sax, Jitendra Malik, and Silvio
  Savarese.
\newblock Gibson env: Real-world perception for embodied agents.
\newblock In {\em CVPR}, 2018.

\bibitem{xie2021segformer}
Enze Xie, Wenhai Wang, Zhiding Yu, Anima Anandkumar, Jose~M. Alvarez, and Ping
  Luo.
\newblock {SegFormer:} {Simple} and efficient design for semantic segmentation
  with transformers.
\newblock In {\em NeurIPS}, 2021.

\bibitem{yang2022bevformer}
Chenyu Yang, Yuntao Chen, Hao Tian, Chenxin Tao, Xizhou Zhu, Zhaoxiang Zhang,
  Gao Huang, Hongyang Li, Yu Qiao, Lewei Lu, Jie Zhou, and Jifeng Dai.
\newblock {BEVFormer v2:} {Adapting} modern image backbones to bird's-eye-view
  recognition via perspective supervision.
\newblock In {\em CVPR}, 2023.

\bibitem{yuan2020ocr}
Yuhui Yuan, Xilin Chen, and Jingdong Wang.
\newblock Object-contextual representations for semantic segmentation.
\newblock In {\em ECCV}, 2020.

\bibitem{orientation}
Chao Zhang, Stephan Liwicki, William Smith, and Roberto Cipolla.
\newblock Orientation-aware semantic segmentation on icosahedron spheres.
\newblock In {\em ICCV}, 2019.

\bibitem{zhang2021trans4trans}
Jiaming Zhang, Kailun Yang, Angela Constantinescu, Kunyu Peng, Karin
  M{\"u}ller, and Rainer Stiefelhagen.
\newblock {Trans4Trans:} {Efficient} transformer for transparent object
  segmentation to help visually impaired people navigate in the real world.
\newblock In {\em ICCVW}, 2021.

\bibitem{trans4pass}
Jiaming Zhang, Kailun Yang, Chaoxiang Ma, Simon Rei{\ss}, Kunyu Peng, and
  Rainer Stiefelhagen.
\newblock Bending reality: Distortion-aware transformers for adapting to
  panoramic semantic segmentation.
\newblock In {\em CVPR}, 2022.

\bibitem{trans4passplus}
Jiaming Zhang, Kailun Yang, Hao Shi, Simon Rei{\ss}, Kunyu Peng, Chaoxiang Ma,
  Haodong Fu, Kaiwei Wang, and Rainer Stiefelhagen.
\newblock Behind every domain there is a shift: Adapting distortion-aware
  vision transformers for panoramic semantic segmentation.
\newblock {\em arXiv preprint arXiv:2207.11860}, 2022.

\bibitem{zhao2017pspnet}
Hengshuang Zhao, Jianping Shi, Xiaojuan Qi, Xiaogang Wang, and Jiaya Jia.
\newblock Pyramid scene parsing network.
\newblock In {\em CVPR}, 2017.

\bibitem{structured3d}
Jia Zheng, Junfei Zhang, Jing Li, Rui Tang, Shenghua Gao, and Zihan Zhou.
\newblock {Structured3D:} {A} large photo-realistic dataset for structured {3D}
  modeling.
\newblock In {\em ECCV}, 2020.

\bibitem{zheng2023complementary}
Zishuo Zheng, Chunyu Lin, Lang Nie, Kang Liao, Zhijie Shen, and Yao Zhao.
\newblock Complementary bi-directional feature compression for indoor
  360{\textdegree} semantic segmentation with self-distillation.
\newblock In {\em WACV}, 2023.

\bibitem{zhou2022cross}
Brady Zhou and Philipp Kr{\"a}henb{\"u}hl.
\newblock Cross-view transformers for real-time map-view semantic segmentation.
\newblock In {\em CVPR}, 2022.

\bibitem{zhu2020deformabledetr}
Xizhou Zhu, Weijie Su, Lewei Lu, Bin Li, Xiaogang Wang, and Jifeng Dai.
\newblock Deformable {DETR}: {Deformable} transformers for end-to-end object
  detection.
\newblock In {\em ICLR}, 2021.

\bibitem{zioulis20193D60}
Nikolaos Zioulis, Antonis Karakottas, Dimitrios Zarpalas, Federico Alvarez, and
  Petros Daras.
\newblock Spherical view synthesis for self-supervised 360{\textdegree} depth
  estimation.
\newblock In {\em 3DV}, 2019.

\end{thebibliography}
}

\clearpage
\appendix
\normalsize 

\section{Data Generation}

To perform the data generation, we use an open-source tool\footnote{The \href{https://github.com/atlantis-ar/matterport_utils}{matterport utils} tool.} to convert the 3D mesh semantic labels in Matterport3D~\cite{Matterport3D} into $194,400$ pinhole images with semantic labels. Then, every $18$ semantic label pairs are concatenated via a corresponding rotation-translation matrix, yielding $10,800$ panoramic semantic ground truth, which is referred to as 360FV-Matterport by us. These panoramic semantic images are originally annotated with $40$ object categories.
Because many of them are only a small percentage (${\ll}0.1\%$), we merges some uncommon classes and maintains the $20$ most common object categories: \texttt{wall}, \texttt{floor}, \texttt{chair}, \texttt{door}, \texttt{table}, \texttt{picture}, \texttt{furniture}, \texttt{objects}, \texttt{window}, \texttt{sofa}, \texttt{bed}, \texttt{sink}, \texttt{stairs}, \texttt{ceiling}, \texttt{toilet}, \texttt{mirror}, \texttt{shower}, \texttt{bathtub}, \texttt{counter}, and \texttt{shelving}. 
For another front-view semantic segmentation dataset, Stanford2D3D~\cite{stanford2d3d}, we keep the original object classes: \texttt{beam},  \texttt{board},  \texttt{bookcase}, \texttt{ceiling},  \texttt{chair},  \texttt{clutter}, \texttt{column},  \texttt{door},  \texttt{floor},  \texttt{sofa},  \texttt{table}, \texttt{wall},  \texttt{window}.

For the presented 360BEV-Stanford dataset, we follow the data split method of Fold-1 of the Stanford2D3D~\cite{stanford2d3d} dataset. On the BEV dataset, we use the \textit{area1}, \textit{area2}, \textit{area3}, \textit{area4} and \textit{area6} as the training data for the proposed 360BEV task, and we use the \textit{area5a} and \textit{area5b} as the validation set to evaluate the panoramic semantic mapping performance of models. The results of training and evaluation with the Fold-1 data split is similar the average scores which are calculated by using three-fold cross-validation. Besides, the validation set from Fold-1 is sufficient to evaluate the model performance on panoramic semantic mapping.

For 360BEV-Matterport, we use a different data split compared to Wijmans~\etal~\cite{wijmans2019embodied}. Instead of using synthetic simulators, all samples on our dataset are converted from the real images and labels of Matterport3D~\cite{Matterport3D} dataset, where there are $86$ unique floors on our dataset, including $61$ for training, $7$ for validations, and $18$ for testing.

\section{More Quantitative Analysis}
\subsection{Results on Stanford2D3D}
In Table~\ref{tab:s2d3d_front_every_class}, we present the per-class IoU results of front-view semantic segmentation on the Stanford2D3D dataset. The average (Avg.) scores are calculated with three folds~\cite{stanford2d3d} of cross validation, where Fold-2 is the most challenging split on the Stanford2D3D dataset. Compared to previous state-of-the-art Trans4PASS~\cite{trans4pass}, our proposed 360Mapper achieves $47.97\%$ mIoU in Fold-2 split. 
Besides, our 360Mapper model has overall better performance ($54.34\%$ in mIoU) in the average result calculated by three folds evaluation, surpassing the previous Trans4PASS model with ${+}2.24\%$ in mIoU. Furthermore, our model achieves the highest scores in $11$ of $13$ categories, including \textit{board},  \textit{bookcase}, \textit{ceiling},  \textit{chair},  \textit{clutter}, \textit{door},  \textit{floor},  \textit{sofa},  \textit{table}, \textit{wall}, and  \textit{window}. Improvements in these categories demonstrate the effectiveness of our 360Mapper model in combating distortions of 360{\textdegree} front-view images by incorporating distortion-aware 360Attention.

\begin{table*}[!t]
\renewcommand\arraystretch{1.2}
  \scriptsize
  \setlength{\tabcolsep}{7pt}

\begin{center}
\caption{\textbf{Per-class results (360FV)} on the Stanford2D3D dataset. The models are based on the MiT-B2~\cite{xie2021segformer} backbone.}
\vskip -2ex
\label{tab:s2d3d_front_every_class}
\setlength{\tabcolsep}{1mm}
\resizebox{0.99\textwidth}{!}{
    \begin{tabular}{ l | c | c| c c c c c c c c c c c c c}
    \toprule
    \textbf{Method} & \textbf{Split}  & \textbf{\rotatebox{90}{mIoU}} &  \rotatebox{90}{beam} &  \rotatebox{90}{board} &  \rotatebox{90}{bookcase} & \rotatebox{90}{ceiling} &  \rotatebox{90}{chair} &  \rotatebox{90}{clutter} & \rotatebox{90}{column} &  \rotatebox{90}{door}&  \rotatebox{90}{floor} &  \rotatebox{90}{sofa} &  \rotatebox{90}{table} & \rotatebox{90}{wall} &  \rotatebox{90}{window} \\
    \midrule\midrule
    Trans4PASS~\cite{trans4pass} & Fold-1 & {53.30} & {00.40} & {69.50} & {62.20} & {82.80} & {58.50} & {34.30} & {21.90} & {44.90} & {91.20} & {40.80} & {57.70} & {74.80} & {54.20} \\
    Trans4PASS~\cite{trans4pass} & Fold-2 & {45.70} & {12.50} & {46.90} & {32.60} & {82.30} & {64.70} & {37.50} & {20.10} & {42.70} & {86.60} & {17.70} & {45.20} & {70.30} & {35.10} \\
    Trans4PASS~\cite{trans4pass} & Fold-3 & {57.20} & {21.40} & {65.40} & {58.30} & {80.20} & {55.80} & {41.90} & {28.60} & {76.30} & {88.60} & {45.40} & {58.80} & {59.30} & {63.60} \\\hline
    \rowcolor{gray!10} Trans4PASS~\cite{trans4pass} & Avg. & 52.10 & \textbf{11.40} & 60.60 & 51.10 & 81.80 & 59.70 & 37.90 & \textbf{23.50} & {54.60} & 88.80 & {34.60} & 53.90 & 68.10 & 51.00 \\\midrule
    
    360Mapper & Fold-1  & 56.46  &00.57 &74.61 &65.03 &83.96 &62.41 &40.27 &18.72 &42.22 &93.31 &53.86 &65.90 &76.18 &58.84 \\
    360Mapper & Fold-2 &47.97 &09.32 &41.89 &40.45 &83.01 &62.27 &34.92 &25.74 &57.74 &88.02 &24.48 &42.95 &72.19 &41.22 \\
    360Mapper & Fold-3 &58.60  &08.05 &74.32 &61.05 &81.05 &63.29 &44.44 &4.64 &76.56 &90.91 &57.28 &62.52 &64.96 &72.77  \\\hline
    \rowcolor{gray!15} 360Mapper & Avg. &\textbf{54.34}&	05.98&	\textbf{63.61}&	\textbf{55.51}&	\textbf{82.67}&	\textbf{62.66}&	\textbf{39.88}&	16.37&	\textbf{58.84}&	\textbf{90.75}&	\textbf{45.21}&	\textbf{57.12}&	\textbf{71.11}&	\textbf{57.61}\\
    \bottomrule
    \end{tabular}
}
\end{center}
\vskip -3ex
\end{table*}

\subsection{Results on 360FV-Matterport}
\begin{table}[!t]\begin{center}
\caption{\textbf{Panoramic semantic segmentation (360FV)} on the \texttt{test} set of 360FV-Matterport dataset.}
\vskip-2ex
\label{tab:mp3d_front_test}
\setlength{\tabcolsep}{5mm}
\renewcommand\arraystretch{1.2}
\resizebox{\columnwidth}{!}{
    \begin{tabular}{ l | c | c | c}
    \toprule[1pt]
    \textbf{Method} & \textbf{Backbone} & \textbf{Input} & \textbf{mIoU(\%)} \\ 
    \midrule \midrule
    HoHoNet~\cite{hohonet} & ResNet-101 & RGB & 40.22  \\ 
    HoHoNet~\cite{hohonet} & ResNet-101 & RGB-D & 41.23 \\
    Trans4PASS~\cite{trans4pass} & MiT-B2 & RGB & 39.70\\ 
    Trans4PASS+~\cite{trans4passplus} & MiT-B2 & RGB & 40.41\\ 
    SegFormer~\cite{xie2021segformer} & MiT-B2 & RGB & 42.49 \\ 
    \rowcolor{gray!15} Ours & MiT-B2  & RGB &  \textbf{43.16}\\
    \bottomrule[1pt]
    \end{tabular}
}
\end{center}
\vskip -3ex
\end{table}
As shown in Table~\ref{tab:mp3d_front_test}, we present the front-view semantic segmentation results on the \texttt{test} set of 360FV-Matterport dataset. We compare our approaches with SegFormer~\cite{xie2021segformer}, Trans4PASS~\cite{trans4pass}, Trans4PASS+~\cite{trans4passplus}, HoHoNet~\cite{hohonet} with RGB and RGB-D, where HoHoNet uses ResNet-101 as backbone and the others use MiT-B2 as backbone. Compared with the well-established existing work SegFormer, our approach obtains a higher mIoU score with $43.16\%$, having a performance improvement of ${+}0.67\%$ mIoU on the test set. The test set is much more challenging than the validation set of 360FV-Matterport dataset, the results in Table~\ref{tab:mp3d_front_test} show the superiority of the proposed approach on extracting the underlying cues for the proposed task.

\begin{table*}[!t]
\renewcommand\arraystretch{1.3}
\begin{center}
\caption{\textbf{Per-class results (360FV)} on the 360FV-Matterport dataset.}
\vskip -2ex
\label{tab:mp3d_front_every_class}
\setlength{\tabcolsep}{2pt}
\resizebox{0.99\textwidth}{!}{
    \begin{tabular}{ l  |c  |c | c| c c c c c c c c c c c c c c c c c c c c}
    \toprule
    \textbf{Method} &\textbf{Backbone} & \textbf{Data}   &\textbf{\rotatebox{90}{mIoU}} &  \rotatebox{90}{wall} &  \rotatebox{90}{floor} &  \rotatebox{90}{chair} & \rotatebox{90}{door} &  \rotatebox{90}{table} &  \rotatebox{90}{picture} & \rotatebox{90}{furniture} &  \rotatebox{90}{objects}&  \rotatebox{90}{window} &  \rotatebox{90}{sofa} &  \rotatebox{90}{bed} & \rotatebox{90}{sink} &  \rotatebox{90}{stairs} &
    \rotatebox{90}{ceiling} &
    \rotatebox{90}{toilet} & \rotatebox{90}{mirror} &  \rotatebox{90}{shower} &
    \rotatebox{90}{bathtub} & \rotatebox{90}{counter} &  \rotatebox{90}{shelving} 
    \\
    \midrule\midrule

    Trans4PASS+ \cite{trans4passplus} &\texttt{MiT-B2} & \texttt{val}   &42.60  &63.37 &79.11 &39.13 &40.31 &32.76 &35.99 &30.96 &31.52 &37.52 &44.01 &63.17 &20.60 &41.76
    &77.55 &40.71 &24.27 &23.73 &58.34 &34.31 &32.90
    \\
    
    \rowcolor{gray!15} 360Mapper &\texttt{MiT-B2} & \texttt{val}   &\textbf{46.35}  &64.12 &83.14 &45.75 &44.98 &37.96 &41.08 &32.26 &35.07 &40.61 &48.69 &69.80 &25.12 &47.80
    &80.15 &45.96 &28.70 &22.31 &60.05 &38.64 &34.82
    \\\midrule

    Trans4PASS+ \cite{trans4passplus} &\texttt{MiT-B2} & \texttt{test}   &40.41  &64.32 &80.12 &41.24 &41.70 &30.86 &36.93 & 35.16 &28.27 &32.65 &33.28 &55.98 &22.93 &37.19 
    &78.36 &48.96 &17.73 &26.51 &49.65 &28.64 &22.82
    \\
   \rowcolor{gray!15} 360Mapper &\texttt{MiT-B2} & \texttt{test}  &\textbf{43.16}  &66.95 &82.24 &45.12 &47.34 &32.72 &44.35 &33.34 &29.57 &34.59  &32.08 &62.06 &28.24 &38.03
    &81.26  &45.47 &23.61 &29.01 &55.44 &28.58 &23.24
    \\
    \bottomrule
    \end{tabular}
}
\end{center}
\vskip -3ex
\end{table*}
Apart from that, per-class IoU scores on 360FV-Matterport in Table~\ref{tab:mp3d_front_every_class}. The performance of 360Mapper on both test and validation sets are demonstrated. 360Mapper delivers $46.35\%$ and $43.16\%$ mIoU performance on validation and test sets of 360FV-Matterport dataset respectively. For per-class IoUs, our model has better performance of challenging class, \eg, \textit{sink} with $25.12\%$ and $28.24\%$ on validation and test sets, surpassing Trans4PASS+~\cite{trans4passplus} with large margins. 
It notes that the small objects, \eg, \textit{furniture}, \textit{mirror}, \textit{toilet} on the test set, are still challenging for both methods. Apart from these, our models have better semantic segmentation results on $17$ of $20$ classes on the 360FV-Matterport dataset.

\subsection{Results on 360BEV-Stanford}
Per-class IoU scores on 360BEV-Stanford are shown in Table~\ref{tab:s2d3d_BEV_every_class}. 
On the 360BEV task, 360Mapper can achieve $45.78\%$ score of mIoU, outperforming the previous Trans4Map~\cite{chen2022trans4map} method with ${+}9.7\%$. Specifically, our 360Mapper achieves per-class IoU with $93.33\%$, $42.52\%$, $59.14\%$, $5.06\%$, $62.66\%$, $39.75\%$, $5.48\%$, $38.74\%$, $97.76\%$, $48.92\%$, $76.76\%$, $45.86\%$ and $24.89\%$ for \textit{void}, \textit{board}, \textit{bookcase}, \textit{ceiling}, \textit{chair}, \textit{clutter}, \textit{column}, \textit{door}, \textit{floor}, \textit{sofa}, \textit{table}, \textit{wall} and \textit{window}, respectively. Especially, the challenging objects that appear thin lines in bird's-eye views, such as \textit{doors} and \textit{walls}, can be more stably recognized by our method, which improves both IoUs with $10.23\%{\rightarrow}38.74\%$ and $29.56\%{\rightarrow}45.86\%$. The \textit{beam} class is not successfully recognized by both methods, because this BEV mechanism directly ignores objects on the ceiling. Different from the front-view semantic segmentation task, the \textit{void} class is included on the 360BEV task, because this class can be used to indicate the invisible area on the BEV semantic maps, which is important for the downstream task, such as path planing.
\begin{table*}[!t]

\renewcommand\arraystretch{1.2}
\begin{center}
\caption{\textbf{Per-class results (360BEV)} on the 360BEV-Stanford2D3D dataset.}
\vskip -2ex
\label{tab:s2d3d_BEV_every_class}
\setlength{\tabcolsep}{5pt}
\resizebox{0.99\textwidth}{!}{
    \begin{tabular}{ l | c | c| c c c c c c c c c c c c c c}
    \toprule[1pt]
    \textbf{Method} & \textbf{Backbone} & \textbf{\rotatebox{90}{mIoU}} & \rotatebox{90}{void}  &\rotatebox{90}{beam} &  \rotatebox{90}{board} &  \rotatebox{90}{bookcase} & \rotatebox{90}{ceiling} &  \rotatebox{90}{chair} &  \rotatebox{90}{clutter} & \rotatebox{90}{column} &  \rotatebox{90}{door}&  \rotatebox{90}{floor} &  \rotatebox{90}{sofa} &  \rotatebox{90}{table} & \rotatebox{90}{wall} &  \rotatebox{90}{window} \\
    \midrule\midrule

    Trans4Map~\cite{chen2022trans4map} &\texttt{MiT-B2} &36.08 &64.17  &0.00 &28.10 &52.96 &0.45 &52.30 &34.71 &6.40 &10.23 &92.18 &44.29 &68.22 &29.56 &21.44 \\

    \rowcolor{gray!15} 360Mapper &\texttt{MiT-B2} &\textbf{45.78} &93.33  
    &0.00 &42.52 &59.14 &5.06 &62.66 &39.75 &5.48 &38.74 &97.76 &48.92 &76.76 &45.86 &24.89 \\
 
    \bottomrule[1pt]
    \end{tabular}
}
\end{center}
\vskip -3ex
\end{table*}

\subsection{Results on 360BEV-Matterport}
\begin{table}[!t]
\begin{center}\caption{ \textbf{Panoramic semantic mapping (360BEV)} on the \texttt{test} set of 360BEV-Matterport dataset.}
\label{tab:mp3d_topdown_test}
\vskip -2ex
\setlength{\tabcolsep}{1mm}
\resizebox{\columnwidth}{!}{
\begin{tabular}{ l c c c c l }
\toprule
\textbf{Method} & \textbf{Backbone} & \textbf{Acc} & \textbf{mRecall} & \textbf{mPrecision} & \textbf{mIoU}\\ 
\midrule \midrule
\multicolumn{6}{c}{\textit{(1)Early projection: Proj. $\rightarrow$ Enc. $\rightarrow$ Seg.}} \\\midrule
SegFormer~\cite{xie2021segformer} & MiT-B2 & 69.72 &35.28 & 40.41 & 24.04\\
SegNeXt\cite{guo2022segnext}  & MSCA-B& 69.99 & 36.25 &41.96 & 25.22\\
\midrule
\multicolumn{6}{c}{\textit{(2) Late projection: Enc. $\rightarrow$ Seg. $\rightarrow$ Proj.}} \\\midrule
HoHoNet~\cite{hohonet} & ResNet101  & 62.89 & 35.18 & 39.54  & 22.01\\
Trans4PASS~\cite{trans4pass} & MiT-B2 & 53.50 &  29.35 & 33.53  & 16.53 \\
Trans4PASS+~\cite{trans4passplus} & MiT-B2 & 57.24 & 30.639 & 34.49 & 17.72 \\
SegFormer~\cite{xie2021segformer} & MiT-B2 & 62.91 & 35.35 & 39.64  & 22.02\\
\midrule
\multicolumn{6}{c}{\textit{(3) Intermediate projection: Enc. $\rightarrow$ Proj. $\rightarrow$ Seg.}} \\\midrule
BEVFormer~\cite{bevformer}& MiT-B2 & 72.04 & 36.69  & 47.90 & 27.46\\
Trans4Map~\cite{chen2022trans4map}& MiT-B0 & 71.78 & 38.27& 43.77 & 26.52  \\
Trans4Map~\cite{chen2022trans4map}& MiT-B2  &72.94 &45.45 & 47.03 &31.08   \\
Trans4Map~\cite{chen2022trans4map}& MiT-B4 &73.60 &44.33 &49.91 &31.79\\
\rowcolor{gray!25} Ours & MiT-B0 &76.02 &43.11 &50.41 &31.35 ~\gbf{+4.83} \\ 
\rowcolor{gray!25} Ours & MiT-B2 &78.04 &54.47 &54.27  & 38.78 ~\gbf{+7.70} \\ 
\rowcolor{gray!15} Ours & MSCA-B &79.17 &55.16 &57.27 & 40.27\\ 
\bottomrule
\end{tabular}
}
\end{center}
\vskip-5ex
\end{table}
The 360BEV results on the \texttt{test} set of 360BEV-Matterport are demonstrated in Table~\ref{tab:mp3d_topdown_test}.
We further compare our approach with three backbones, \eg, MiT-B0, MiT-B2 from SegFormer~\cite{xie2021segformer} and MSCA-B from SegNeXt~\cite{guo2022segnext} 
on the test set of the 360BEV-Matterport for the panoramic semantic mapping task. Methods based on intermediate projection show the most promising results compared with those based on early projection and late projection. The result is consistent compared with the ones demonstrated on the validation set of 360BEV-Matterport dataset. 360Mapper still delivers the state-of-the art results for the proposed 360BEV task on the test set, indicating the effectiveness of the proposed architecture. Especially, our 360Mapper with MiT-B2 backbone ($38.78\%$) can surpass Trans4Map with MiT-B2 ($31.08\%$) as well as the one with MiT-B4 ($31.79\%$). Besides, the proposed method based on MSCA-B backbone achieves the best result with $40.27\%$ in mIoU. 

Per-class IoU scores on the 360BEV-Matterport dataset are presented in Table~\ref{tab:mp3d_BEV_every_class}. The performance of 360Mapper under MiT-B2 from SegFormer~\cite{xie2021segformer} and MSCA-B from SegNeXt~\cite{guo2022segnext}  are included, which achieves promising performance for the 360BEV task. Compared to Trans4Map~\cite{chen2022trans4map}, our 360Mapper with the same MiT-B2 backbone can achieve respective $44.32\%$ and $38.78\%$ in mIoU on the validation set and the test set. The \textit{void} class is also included on the 360BEV-Matterport dataset. 
Besides, if using a stronger backbone, \eg, MSCA-B~\cite{guo2022segnext}, our proposed mehods can achieve higher semantic mapping results on both of validation and test sets of 360BEV-Matterport dataset, which are $46.31\%$ and $40.27\%$ in mIoU, respectively.

\begin{table*}[!t]
\renewcommand\arraystretch{1.5}
\begin{center}
\caption{\textbf{Per-class results (360BEV)} on the 360BEV-Matterport dataset.}
\label{tab:mp3d_BEV_every_class}
\setlength{\tabcolsep}{3pt}
\resizebox{0.99\textwidth}{!}{
    \begin{tabular}{ l |c |c | c| c c c c c c c c c c c c c c c c c c c c c}
    \toprule[1pt]
    \textbf{Method} & \textbf{Backbone} & \textbf{Data}  &\textbf{\rotatebox{90}{mIoU}} & \rotatebox{90}{void} & \rotatebox{90}{wall} &  \rotatebox{90}{floor} &  \rotatebox{90}{chair} & \rotatebox{90}{door} &  \rotatebox{90}{table} &  \rotatebox{90}{picture} & \rotatebox{90}{furniture} &  \rotatebox{90}{objects}&  \rotatebox{90}{window} &  \rotatebox{90}{sofa} &  \rotatebox{90}{bed} & \rotatebox{90}{sink} &  \rotatebox{90}{stairs} &
    \rotatebox{90}{ceiling} &
    \rotatebox{90}{toilet} & \rotatebox{90}{mirror} &  \rotatebox{90}{shower} &
    \rotatebox{90}{bathtub} & \rotatebox{90}{counter} &  \rotatebox{90}{shelving} 
    \\
    \midrule\midrule
    
    Trans4Map\cite{chen2022trans4map} &\texttt{MiT-B2} & \texttt{val} &36.72 &47.87 
    &28.52 &82.96  &34.44 &22.27 &39.58 &16.28 &22.75 &26.29 &25.08 &42.81 &62.25
    &13.95 &41.51 &37.79 &45.82 &19.56 &48.05 &47.71 &38.25 &27.31
    \\
        
    \rowcolor{gray!25} 360Mapper &\texttt{MiT-B2} & \texttt{val} &44.32 &74.30 
    &31.94 &85.85 &42.01 &26.71 &46.40 &23.21 &25.00 &24.87 &27.36 &51.37 &66.59
    &20.99 &47.07 &54.97 &56.91 &29.50 &55.70 &63.16 &45.82 &31.04
    \\

    \rowcolor{gray!10} 360Mapper &\texttt{MSCA-B} & \texttt{val} &\textbf{46.31} &74.43 
    &35.62 &86.17 &43.60 &28.56 &50.61 &25.11 &25.17 &26.26 &27.56 &53.17 &69.36
    &24.02 &50.24 &61.26 &62.11 &31.77 &51.60 &65.71 &47.32 &33.06
    \\ \midrule
    
    Trans4Map\cite{chen2022trans4map} &\texttt{MiT-B2} & \texttt{test}   &31.08  &40.51 
    &32.54  &80.21  &33.23 &20.85  &37.21 &19.01  &18.46  &23.05  &23.56 &32.35 &52.08 &15.34
    &29.02 &18.27 &41.90 &15.39 &25.58 &48.19 &30.38 &15.52
    \\
    
    \rowcolor{gray!25}360Mapper &\texttt{MiT-B2} & \texttt{test}   &38.78  &60.36 &36.77 &84.34 &39.93 &24.41 &44.58 &25.23 &21.97 &25.20 &27.06 &36.59 &60.84 &28.46
    &35.60 &49.69 &57.39 &19.35 &25.84 &56.91 &37.23 &16.60
    \\
    
    \rowcolor{gray!10}360Mapper &\texttt{MSCA-B} & \texttt{test} &\textbf{40.27}  &62.82 &40.09 &85.22 &42.60 &25.48 &46.00 &24.37 &25.11 &26.08  &27.39 &39.68 &61.45 &28.18
    &36.17 &50.88 &58.31 &19.77  &29.85 &59.78 &35.39 &21.14
    \\
    
    \bottomrule[1pt]
    \end{tabular}
}
\end{center}
\vskip -3ex
\end{table*}

\section{More Qualitative Analysis}

\subsection{Analysis on Stanford2D3D}
\begin{figure*}[t]
    \footnotesize
    \setlength\tabcolsep{1pt}
    {
    \newcolumntype{P}[1]{>{\centering\arraybackslash}p{#1}}
    \begin{tabular}{@{}*{13}{P{0.1475\columnwidth}}@{}}
     {\cellcolor[rgb]{1,   0,   0.16}}\textcolor{white}{beam} 
    &{\cellcolor[rgb]{1,   0.28, 0}}\textcolor{white}{board}
    &{\cellcolor[rgb]{1,   0.73, 0}}\textcolor{black}{bookcase}
    &{\cellcolor[rgb]{0.8,  1,   0}}\textcolor{black}{ceiling}
    &{\cellcolor[rgb]{0.36, 1,   0}}\textcolor{black}{chair} 
    &{\cellcolor[rgb]{0,    1,  0.08}}\textcolor{black}{clutter} 
    &{\cellcolor[rgb]{0,    1,  0.55}}\textcolor{black}{column}
    &{\cellcolor[rgb]{0,    1,  0.99}}\textcolor{black}{door}
    &{\cellcolor[rgb]{0,   0.56, 1 }}\textcolor{white}{floor}
    &{\cellcolor[rgb]{0,   0.09, 1 }}\textcolor{white}{sofa} 
    &{\cellcolor[rgb]{0.35, 0,   1  }}\textcolor{white}{table}
    & {\cellcolor[rgb]{0.8,  0,   1 }}\textcolor{white}{wall}
    &{\cellcolor[rgb]{1,    0,   0.75}}\textcolor{white}{window}\\
    \end{tabular}
    }
    \centering
    \begin{tabular}{c c c c}
        \vspace{1pt}
        \raisebox{-0.5\height}{\includegraphics[width=0.24\textwidth]{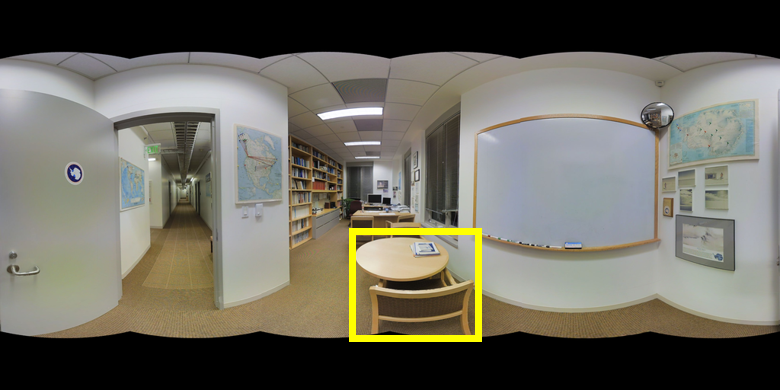}} &
        \raisebox{-0.5\height}{\includegraphics[width=0.24\textwidth]{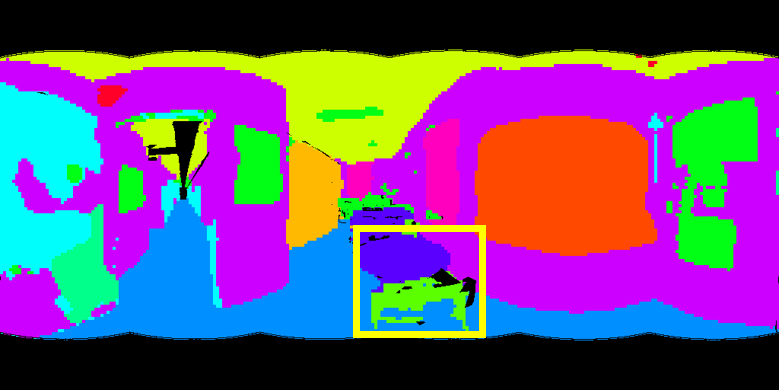}} &
        \raisebox{-0.5\height}{\includegraphics[width=0.24\textwidth]{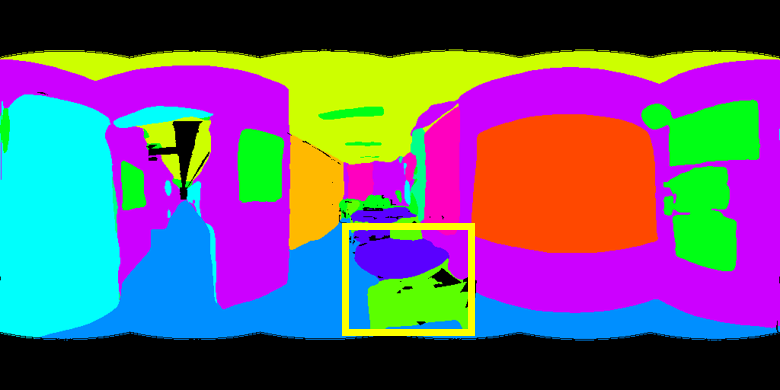}} &
        \raisebox{-0.5\height}{\includegraphics[width=0.24\textwidth]{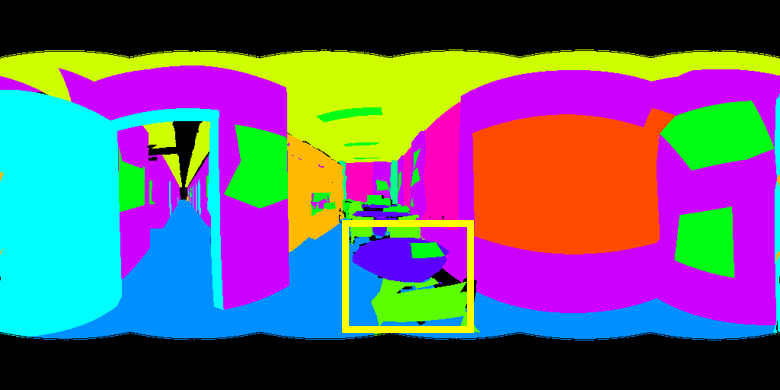}}\\

        \vspace{1pt}

        \raisebox{-0.5\height}{\includegraphics[width=0.24\textwidth]{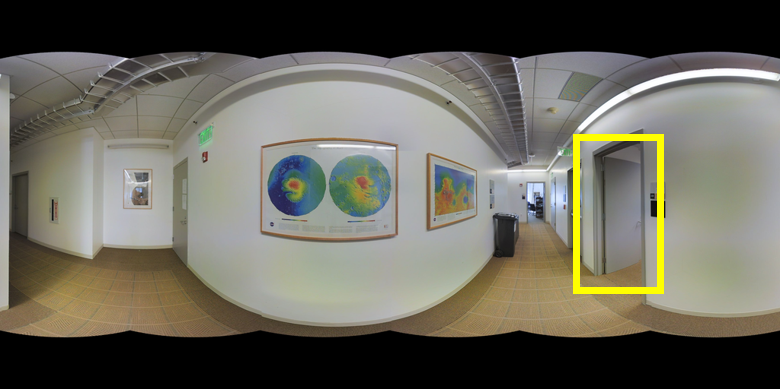}} &
        \raisebox{-0.5\height}{\includegraphics[width=0.24\textwidth]{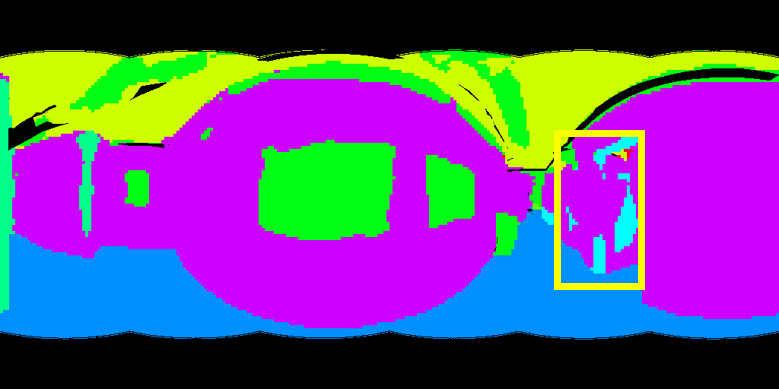}} &
        \raisebox{-0.5\height}{\includegraphics[width=0.24\textwidth]{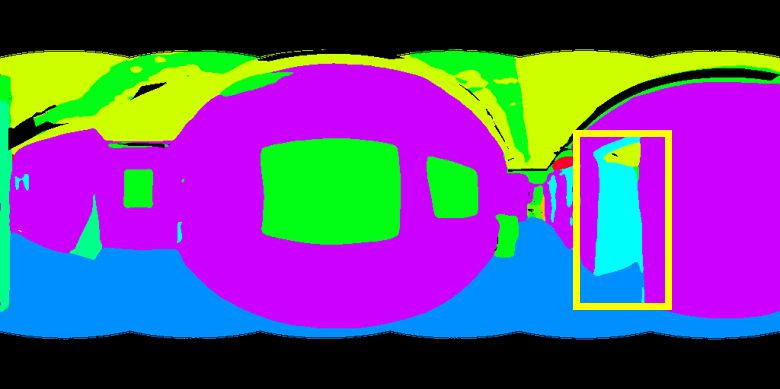}} &
        \raisebox{-0.5\height}{\includegraphics[width=0.24\textwidth]{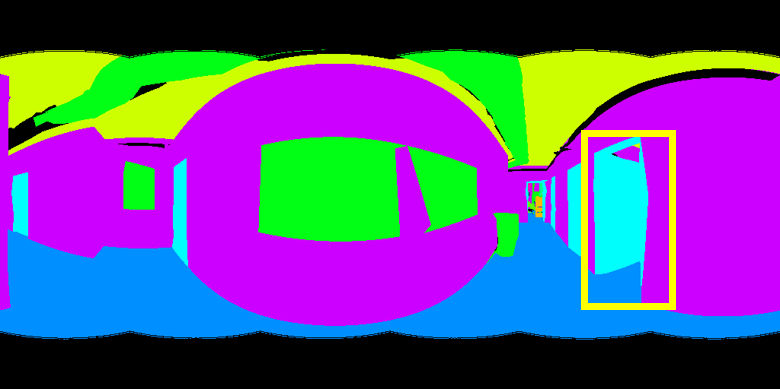}}\\
        
        \vspace{1pt}

        \raisebox{-0.5\height}{\includegraphics[width=0.24\textwidth]{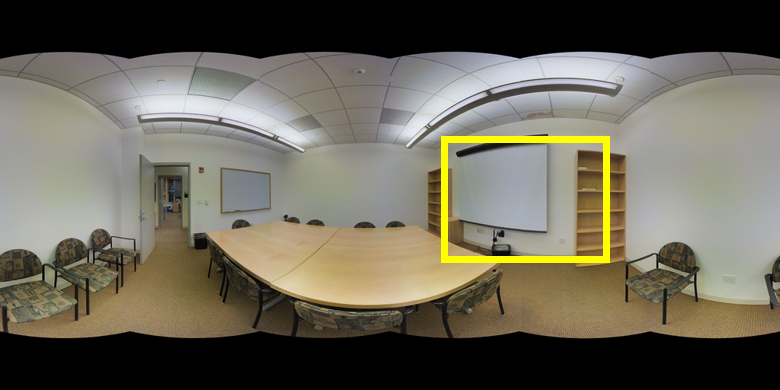}} &
        \raisebox{-0.5\height}{\includegraphics[width=0.24\textwidth]{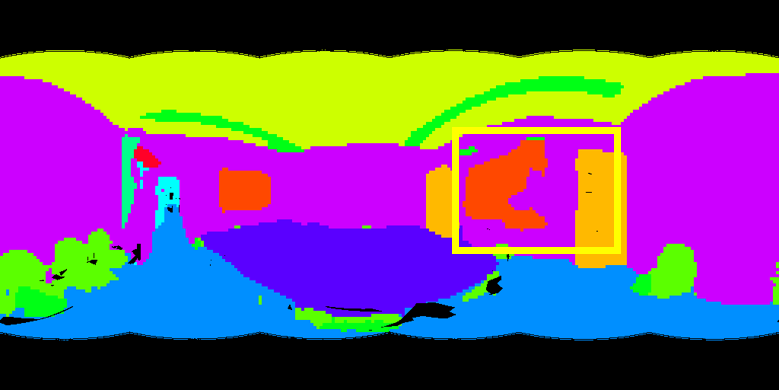}} &
        \raisebox{-0.5\height}{\includegraphics[width=0.24\textwidth]{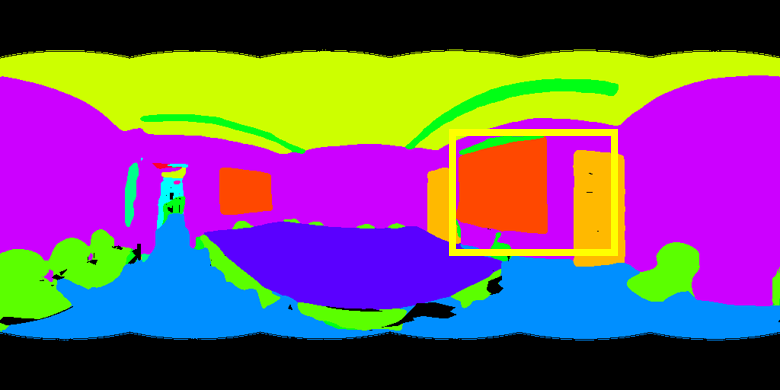}} &
        \raisebox{-0.5\height}{\includegraphics[width=0.24\textwidth]{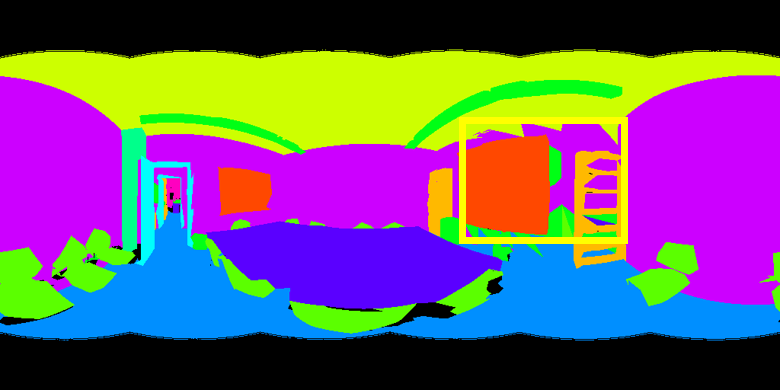}}\\
        
        \vspace{1pt}

        \raisebox{-0.5\height}{\includegraphics[width=0.24\textwidth]{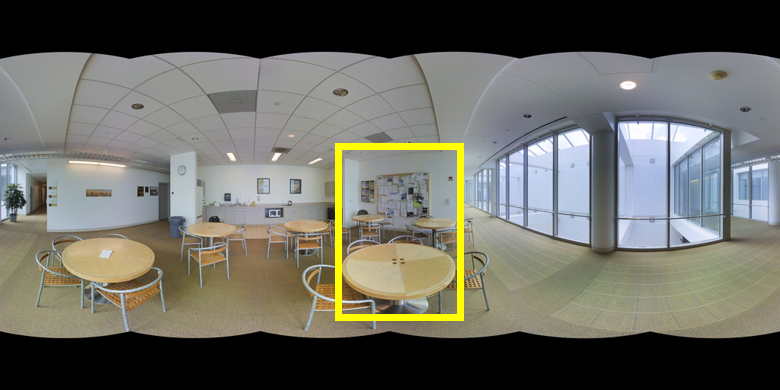}} &
        \raisebox{-0.5\height}{\includegraphics[width=0.24\textwidth]{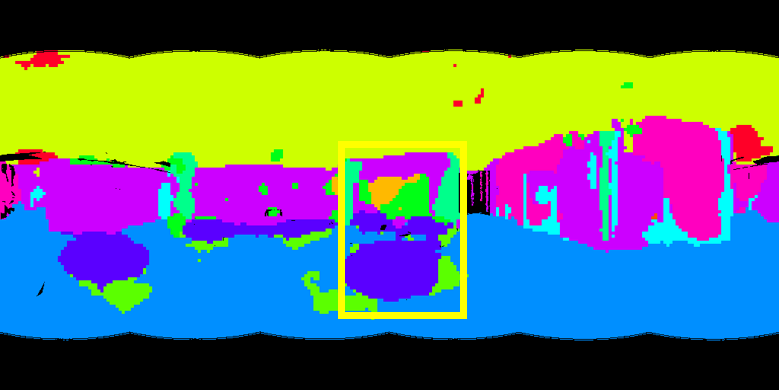}} &
        \raisebox{-0.5\height}{\includegraphics[width=0.24\textwidth]{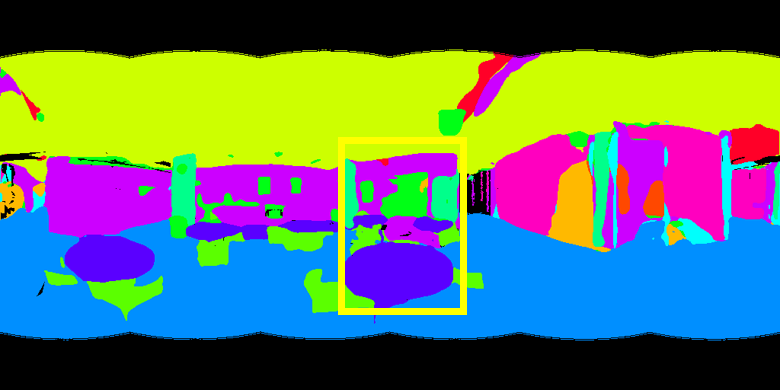}} &
        \raisebox{-0.5\height}{\includegraphics[width=0.24\textwidth]{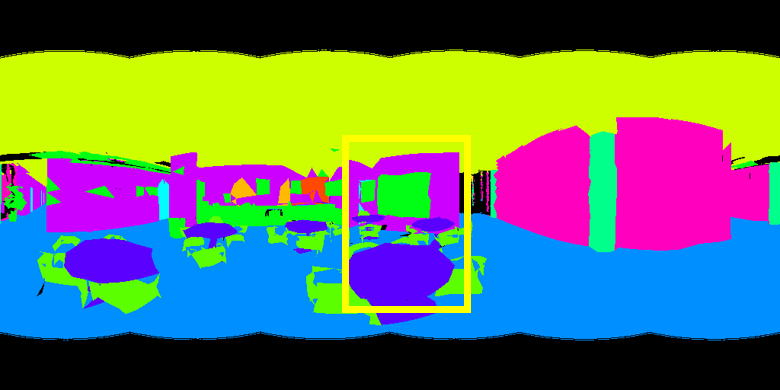}}\\
        
        \vspace{1pt}
        RGB Input&Baseline&Ours&Ground Truth
    \end{tabular}
    \vskip -2ex
    \caption{\textbf{360FV visualization and qualitative analysis} on the Stanford2D3D dataset.}
    \label{fig:s2d3d_fv}
\end{figure*}

The visualization of front-view semantic segmentation (360FV) on the Stanford2D3D dataset is shown in Fig.~\ref{fig:s2d3d_fv}, where the RGB input, the prediction of the baseline, the prediction of our model and the ground truth are depicted from left to the right. The corresponding color map is showcased at the top of Fig.~\ref{fig:s2d3d_fv}. Compared with the baseline Trans4Pass\cite{trans4pass}, the panoramic semantic segmentation results of our model have clear boundaries among different objects which is much more similar to the ground truth, \eg, the \textit{door} and the \textit{clutter} of the second sample. Our method also show promising performance on the objects with small spatial size, \eg, \textit{chairs}, compared with the baseline in the last sample, indicating that our 360Attention approach is good at grasping underlying context feature and cues through the deformable sampling locations. 

\subsection{Analysis on 360FV-Matterport}
\begin{figure*}[t]
    \footnotesize
    \setlength\tabcolsep{1pt}
    {
    \newcolumntype{P}[1]{>{\centering\arraybackslash}p{#1}}
    \begin{tabular}{@{}*{20}{P{0.093\columnwidth}}@{}}
    {\cellcolor[rgb]{0.68,0.78,0.91}}\textcolor{white}{wall} 
    &{\cellcolor[rgb]{0.44,0.50,0.56}}\textcolor{white}{floor}
    &{\cellcolor[rgb]{0.60,0.87,0.54}}\textcolor{black}{chair}
    &{\cellcolor[rgb]{0.77,0.69,0.84}}\textcolor{white}{door}
    &{\cellcolor[rgb]{1.00,0.50,0.05}}\textcolor{white}{table} 
    &{\cellcolor[rgb]{0.84,0.15,0.16}}\textcolor{white}{pictu.} 
    &{\cellcolor[rgb]{0.12,0.47,0.71}}\textcolor{white}{furni.}
    &{\cellcolor[rgb]{0.74,0.74,0.13}}\textcolor{black}{objec.}
    &{\cellcolor[rgb]{1.00,0.60,0.59}}\textcolor{black}{windo.}
    &{\cellcolor[rgb]{0.17,0.63,0.17}}\textcolor{white}{sofa} 
    &{\cellcolor[rgb]{0.89,0.47,0.76}}\textcolor{black}{bed}
    & {\cellcolor[rgb]{0.87,0.62,0.84}}\textcolor{black}{sink}
    &{\cellcolor[rgb]{0.58,0.40,0.74}}\textcolor{white}{stairs} 
    &{\cellcolor[rgb]{0.55,0.64,0.32}}\textcolor{white}{ceil.} 
    &{\cellcolor[rgb]{0.52,0.24,0.22}}\textcolor{white}{toilet} 
    &{\cellcolor[rgb]{0.62,0.85,0.90}}\textcolor{black}{mirror} 
    &{\cellcolor[rgb]{0.61,0.62,0.87}}\textcolor{black}{show.}
    &{\cellcolor[rgb]{0.91,0.59,0.61}}\textcolor{black}{batht.}
    &{\cellcolor[rgb]{0.39,0.47,0.22}}\textcolor{white}{count.} 
    &{\cellcolor[rgb]{0.55,0.34,0.29}}\textcolor{white}{shelv.} \\
    \end{tabular}
    }
    \centering
    \begin{tabular}{c c c c}
        \vspace{1pt}
        \raisebox{-0.5\height}{\includegraphics[width=0.24\textwidth]{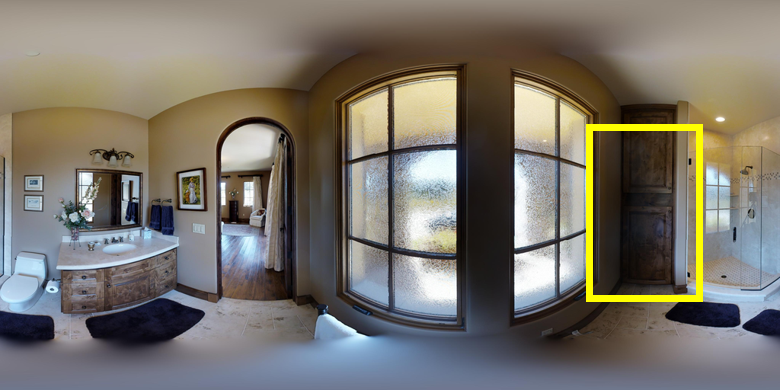}} &
        \raisebox{-0.5\height}{\includegraphics[width=0.24\textwidth]{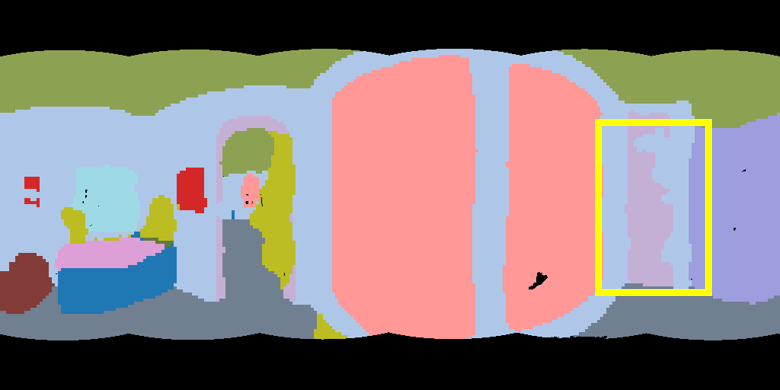}} &
        \raisebox{-0.5\height}{\includegraphics[width=0.24\textwidth]{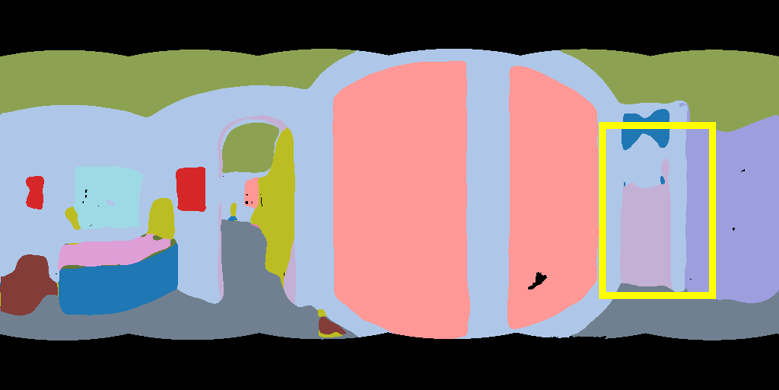}} &
        \raisebox{-0.5\height}{\includegraphics[width=0.24\textwidth]{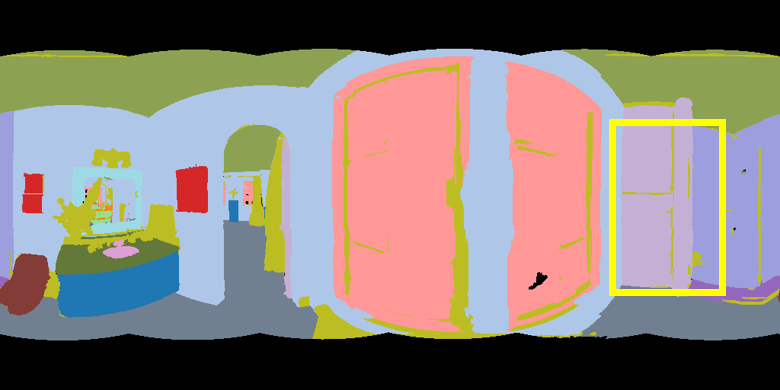}}\\

        \vspace{1pt}

        \raisebox{-0.5\height}{\includegraphics[width=0.24\textwidth]{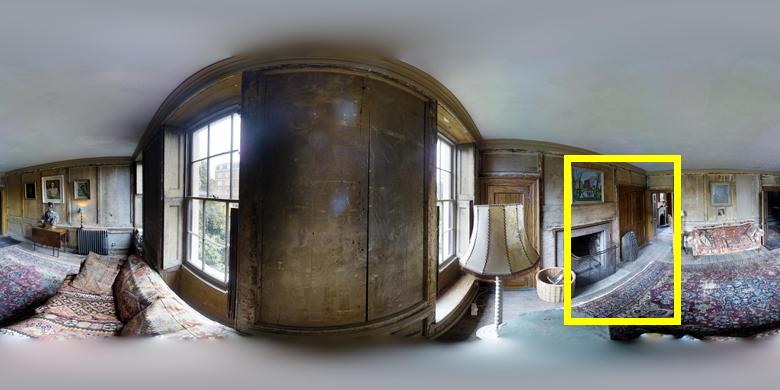}} &
        \raisebox{-0.5\height}{\includegraphics[width=0.24\textwidth]{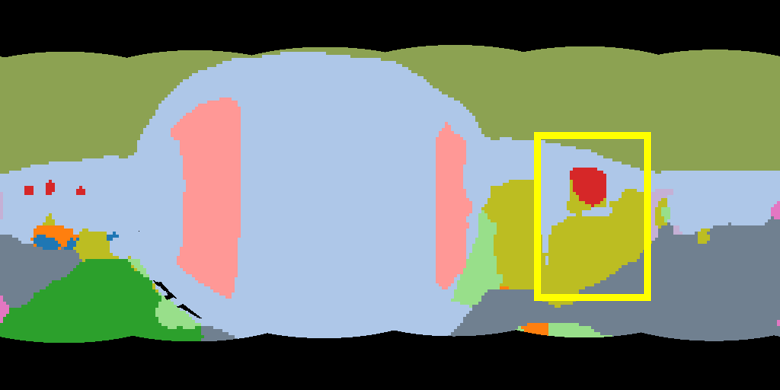}} &
        \raisebox{-0.5\height}{\includegraphics[width=0.24\textwidth]{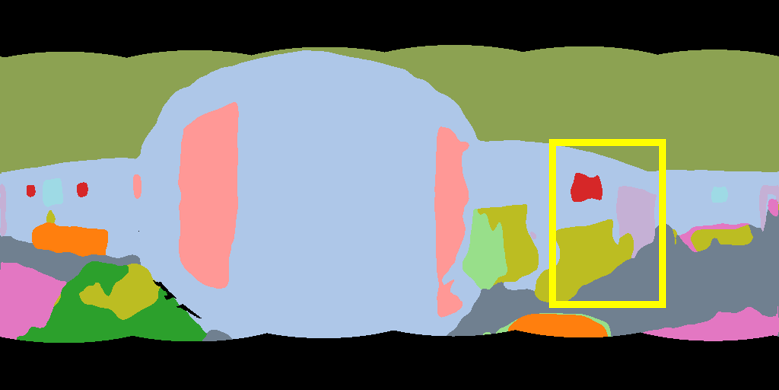}} &
        \raisebox{-0.5\height}{\includegraphics[width=0.24\textwidth]{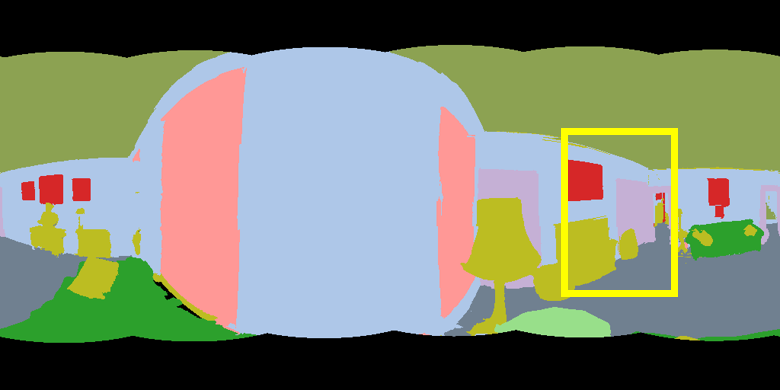}}\\
        
        \vspace{1pt}

        \raisebox{-0.5\height}{\includegraphics[width=0.24\textwidth]{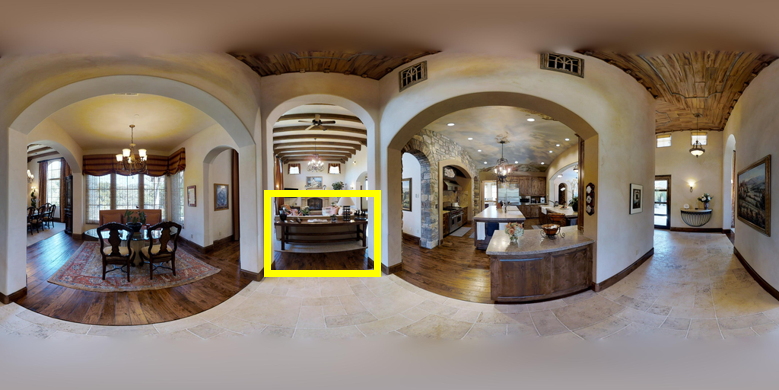}} &
        \raisebox{-0.5\height}{\includegraphics[width=0.24\textwidth]{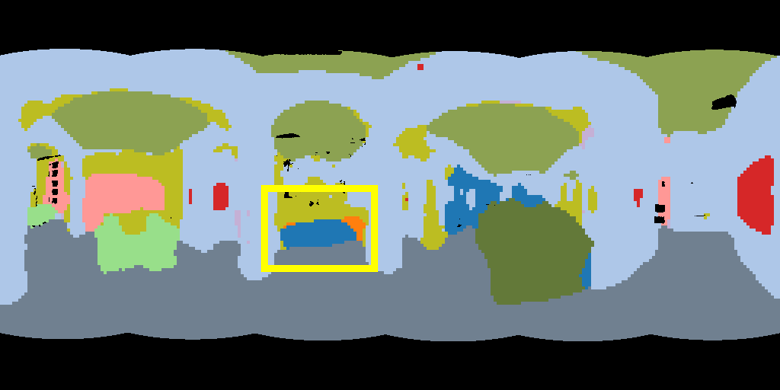}} &
        \raisebox{-0.5\height}{\includegraphics[width=0.24\textwidth]{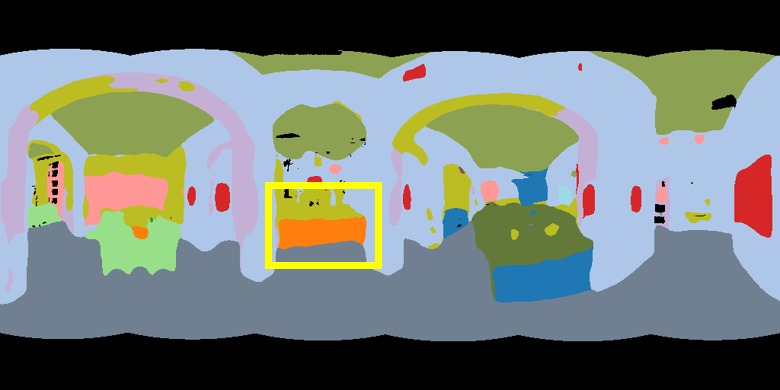}} &
        \raisebox{-0.5\height}{\includegraphics[width=0.24\textwidth]{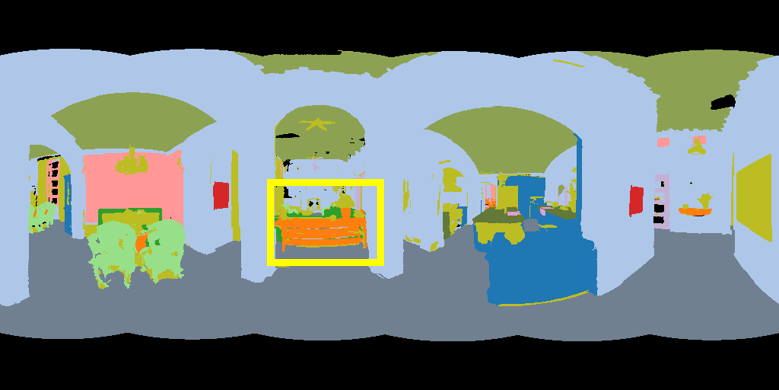}}\\
        
        \vspace{1pt}

        \raisebox{-0.5\height}{\includegraphics[width=0.24\textwidth]{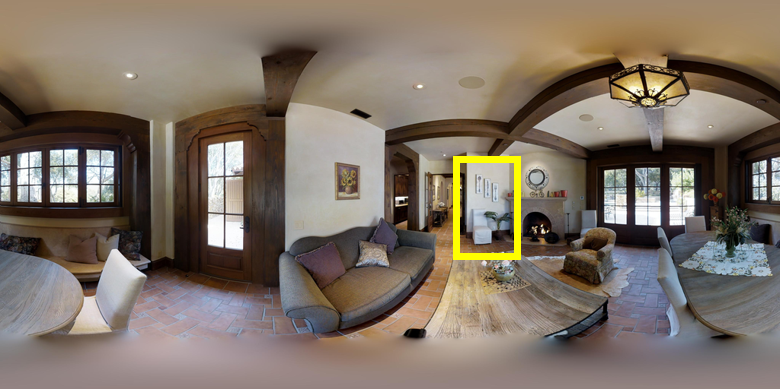}} &
        \raisebox{-0.5\height}{\includegraphics[width=0.24\textwidth]{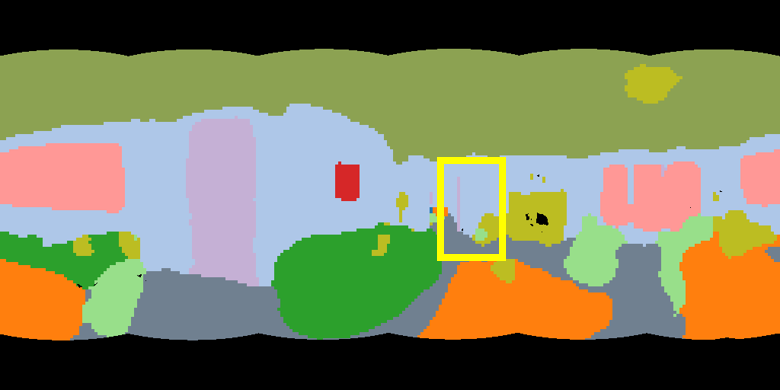}} &
        \raisebox{-0.5\height}{\includegraphics[width=0.24\textwidth]{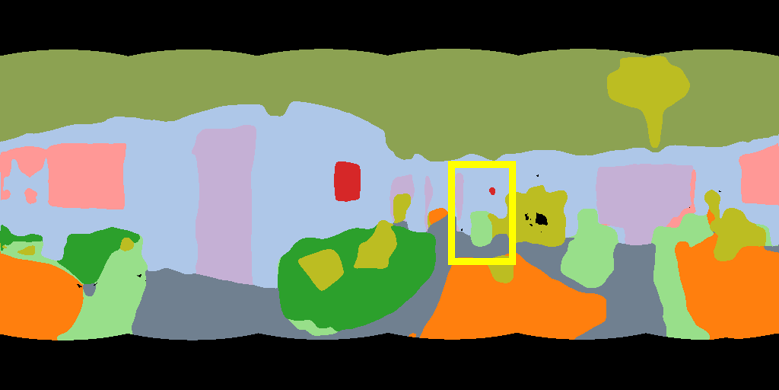}} &
        \raisebox{-0.5\height}{\includegraphics[width=0.24\textwidth]{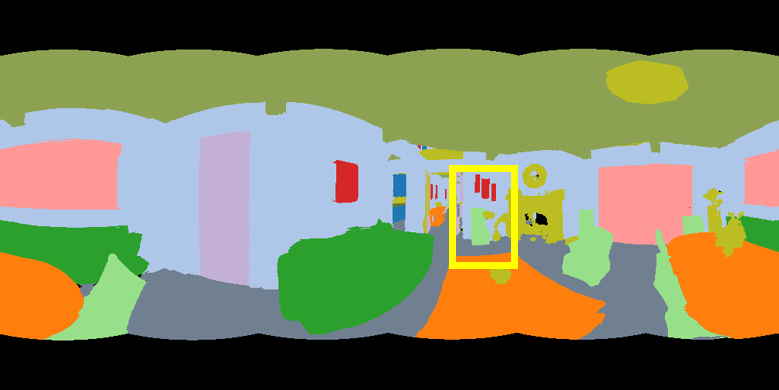}}\\
        \vspace{1pt}
        
        \raisebox{-0.5\height}{\includegraphics[width=0.24\textwidth]{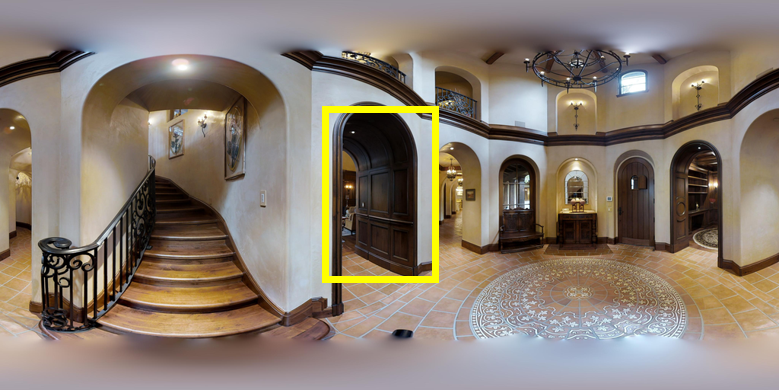}} &
        \raisebox{-0.5\height}{\includegraphics[width=0.24\textwidth]{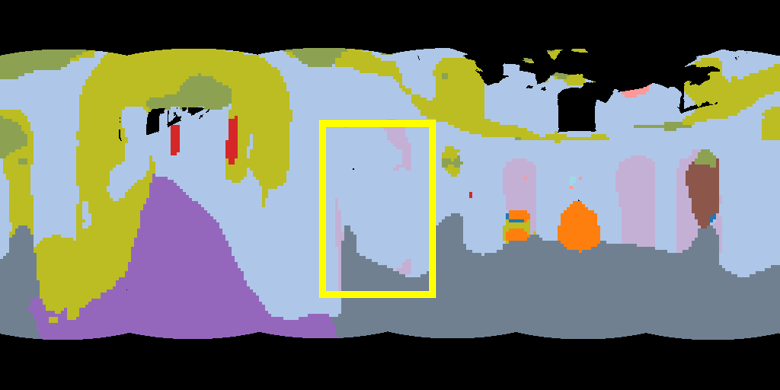}} &
        \raisebox{-0.5\height}{\includegraphics[width=0.24\textwidth]{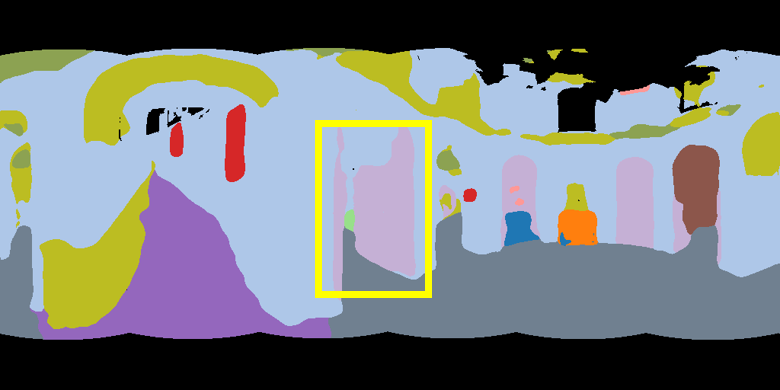}} &
        \raisebox{-0.5\height}{\includegraphics[width=0.24\textwidth]{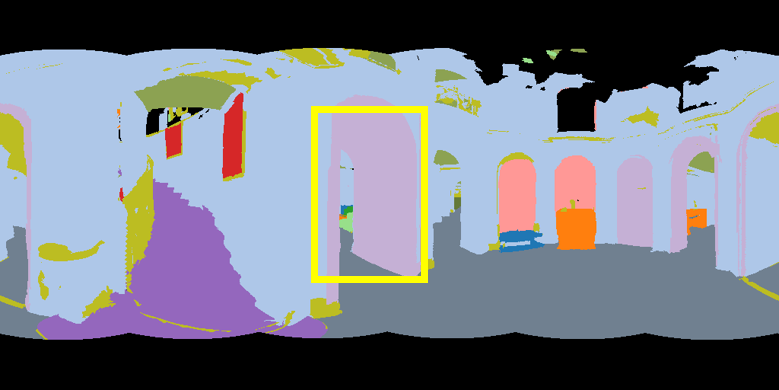}}\\
        
        \vspace{1pt}
        RGB Input&Baseline&Ours&Ground Truth
    \end{tabular}
    \vskip -2ex
    \caption{\textbf{360FV visualization and qualitative analysis} on the 360FV-Matterport dataset.} 
    \label{fig:mp3d_fv}
    \vskip -3ex
\end{figure*}

Fig.~\ref{fig:mp3d_fv} is the front-view semantic segmentation visualization of the presented 360FV-Matterport dataset, providing a detailed depiction of the spatial distribution of different semantic classes. Compared with the baseline method Trans4Pass\cite{trans4pass}, our model produces segmentation results exhibit more precise contours and clearer boundaries between different objects, which closely resemble the ground truth segmentation labels, \eg, the \textit{toilet} and the \textit{door} of the first sample. In the second row, the \textit{door} on the right side is not recognized by the baseline model. In contrast to the baseline method, our model is able to accurately distinguish the \textit{door} class from its surrounding \textit{object} and \textit{wall} classes, despite its small size and low contrast with the surrounding environment. The \textit{table} in the center of the third sample are correctly predicted by our model while it is erroneously segmented by the baseline as \textit{furniture}. This highlights the superior performance of 360Mapper in panoramic semantic segmentation under challenging conditions. In the last two rows, the small \textit{chair} by the wall and the \textit{door} are correctly recognized by our model.

\subsection{Analysis on 360BEV-Stanford}
\begin{figure*}[t]
    \footnotesize
    \setlength\tabcolsep{1pt}
    \begin{center}
    {
    \newcolumntype{P}[1]{>{\centering\arraybackslash}p{#1}}
    \begin{tabular}{@{}*{14}{P{0.135\columnwidth}}@{}}
    {\cellcolor[rgb]{0,   0,   0}}\textcolor{white}{void} 
    &{\cellcolor[rgb]{1,   0,   0.16}}\textcolor{white}{beam} 
    &{\cellcolor[rgb]{1,   0.28, 0}}\textcolor{white}{board}
    &{\cellcolor[rgb]{1,   0.73, 0}}\textcolor{black}{bookcase}
    &{\cellcolor[rgb]{0.8,  1,   0}}\textcolor{black}{ceiling}
    &{\cellcolor[rgb]{0.36, 1,   0}}\textcolor{black}{chair} 
    &{\cellcolor[rgb]{0,    1,  0.08}}\textcolor{black}{clutter} 
    &{\cellcolor[rgb]{0,    1,  0.55}}\textcolor{black}{column}
    &{\cellcolor[rgb]{0,    1,  0.99}}\textcolor{black}{door}
    &{\cellcolor[rgb]{0,   0.56, 1 }}\textcolor{white}{floor}
    &{\cellcolor[rgb]{0,   0.09, 1 }}\textcolor{white}{sofa} 
    &{\cellcolor[rgb]{0.35, 0,   1  }}\textcolor{white}{table}
    & {\cellcolor[rgb]{0.8,  0,   1 }}\textcolor{white}{wall}
    &{\cellcolor[rgb]{1,    0,   0.75}}\textcolor{white}{window}\\
    \end{tabular}
    }
    \end{center}
    \vskip -3mm
    \centering
    \begin{tabular}{c c c c}
        \vspace{1pt}
        \raisebox{-0.5\height}{\includegraphics[width=0.38\textwidth]{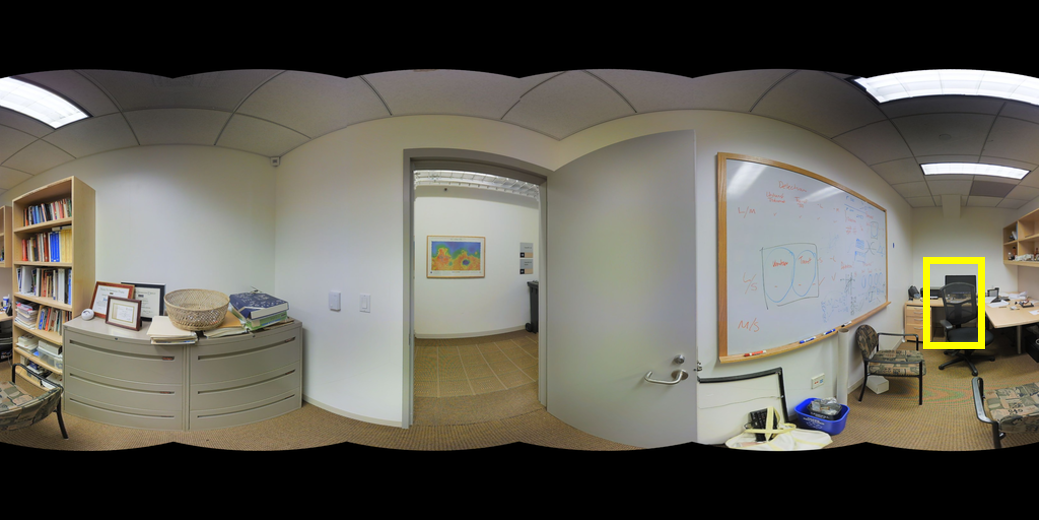}} &
        \raisebox{-0.5\height}{\includegraphics[width=0.19\textwidth]{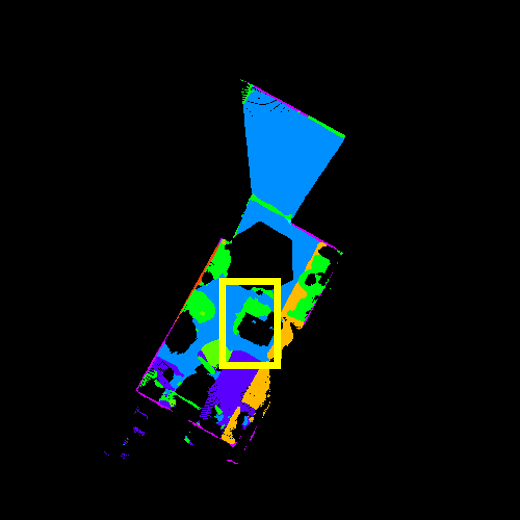}} &
        \raisebox{-0.5\height}{\includegraphics[width=0.19\textwidth]{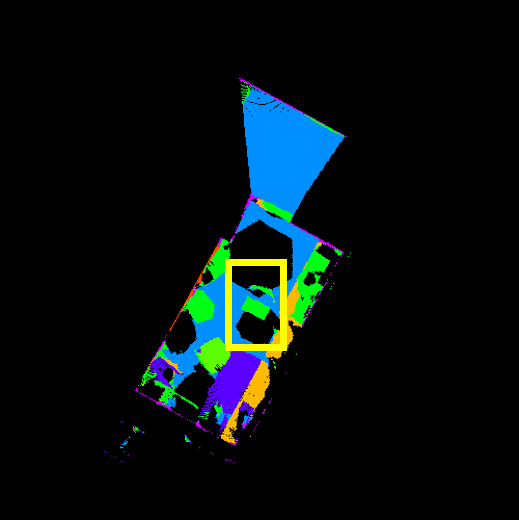}} &
        \raisebox{-0.5\height}{\includegraphics[width=0.19\textwidth]{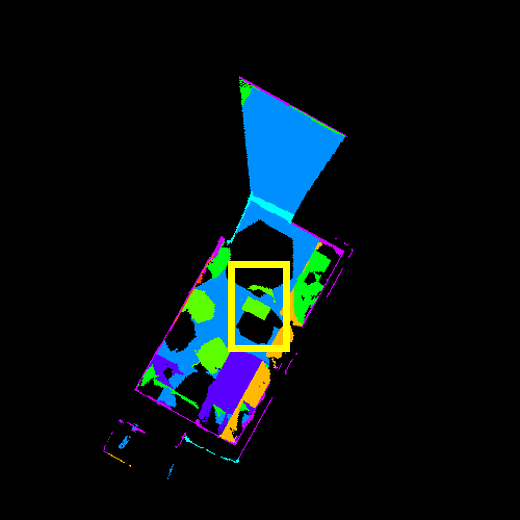}}\\

        \vspace{1pt}

        \raisebox{-0.5\height}{\includegraphics[width=0.38\textwidth]{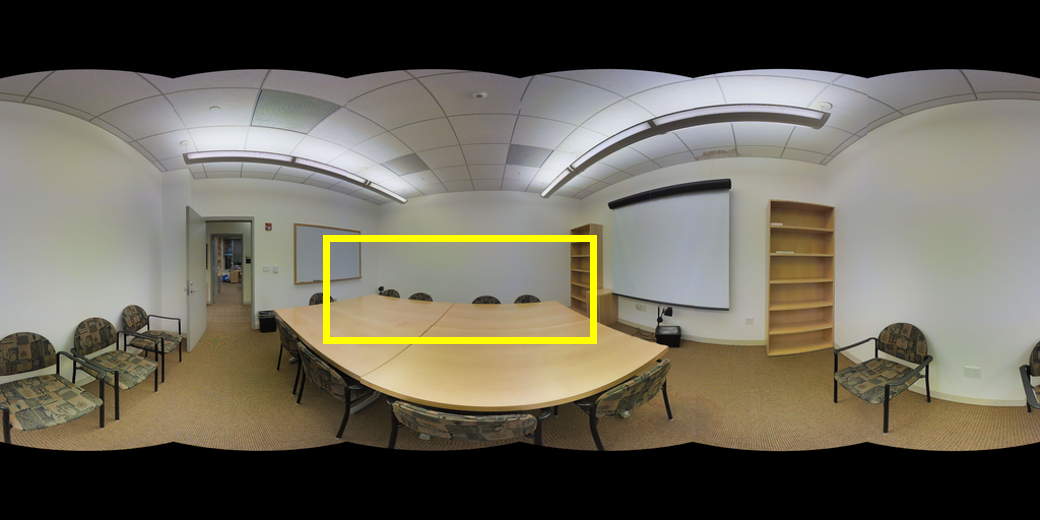}} &
        \raisebox{-0.5\height}{\includegraphics[width=0.19\textwidth]{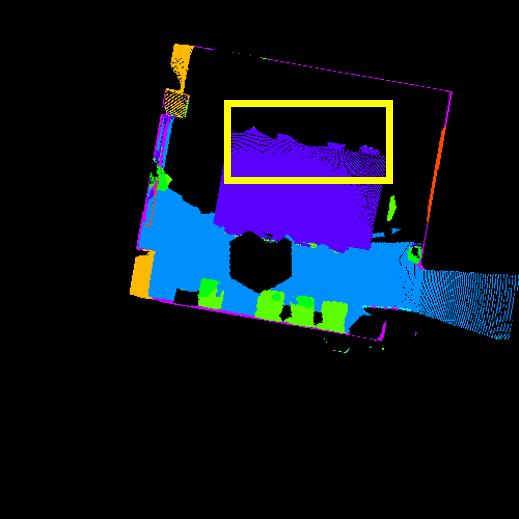}} &
        \raisebox{-0.5\height}{\includegraphics[width=0.19\textwidth]{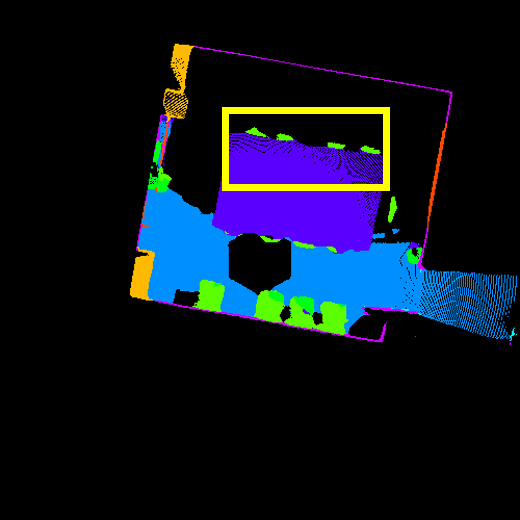}} &
        \raisebox{-0.5\height}{\includegraphics[width=0.19\textwidth]{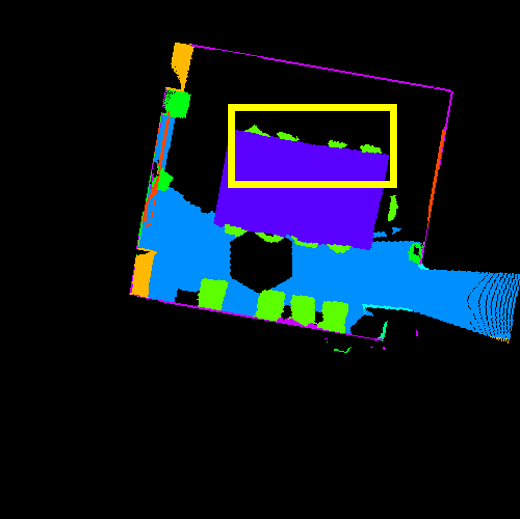}}\\
        
        \vspace{1pt}

        \raisebox{-0.5\height}{\includegraphics[width=0.38\textwidth]{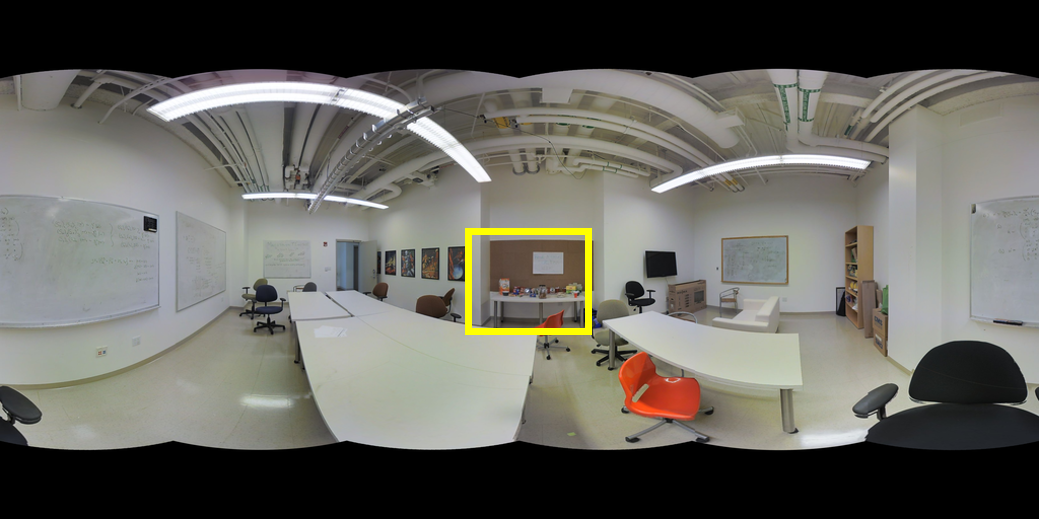}} &
        \raisebox{-0.5\height}{\includegraphics[width=0.19\textwidth]{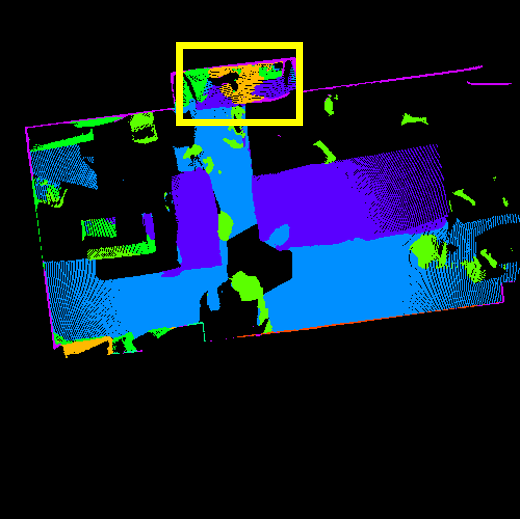}} &
        \raisebox{-0.5\height}{\includegraphics[width=0.19\textwidth]{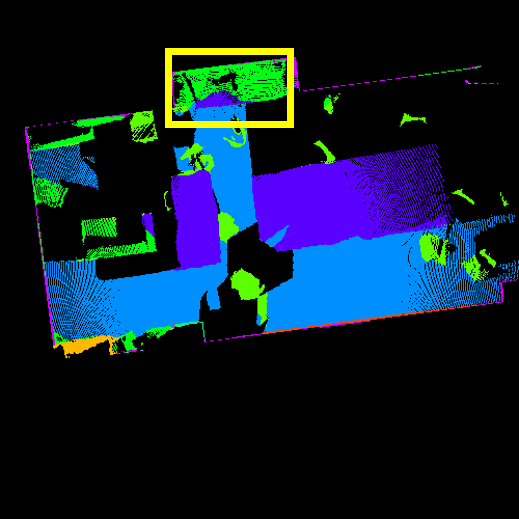}} &
        \raisebox{-0.5\height}{\includegraphics[width=0.19\textwidth]{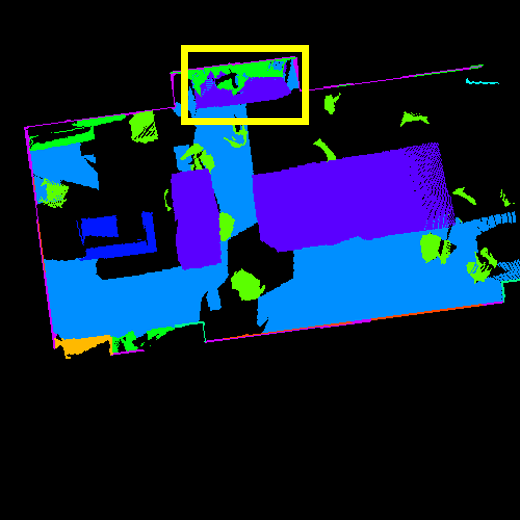}}\\
        
        \vspace{1pt}

        \raisebox{-0.5\height}{\includegraphics[width=0.38\textwidth]{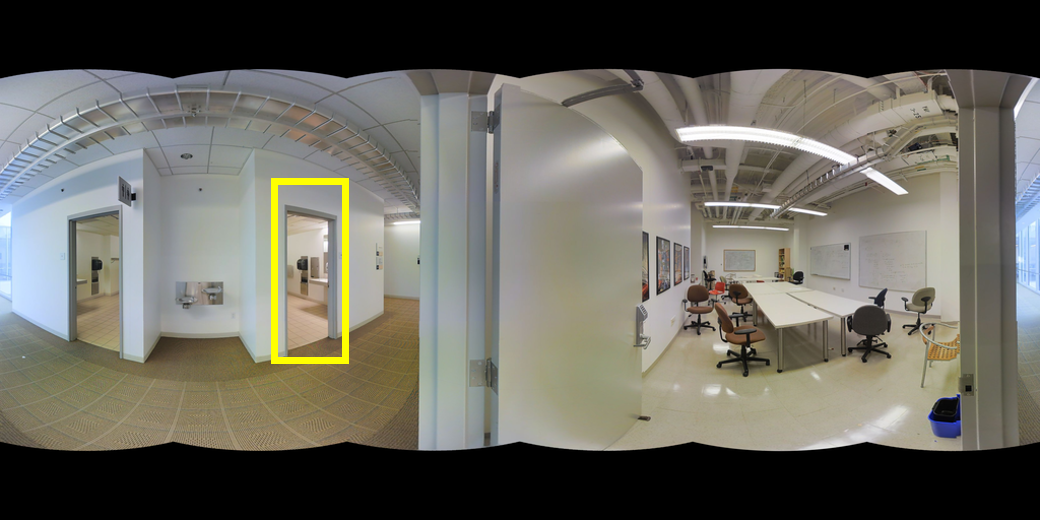}} &
        \raisebox{-0.5\height}{\includegraphics[width=0.19\textwidth]{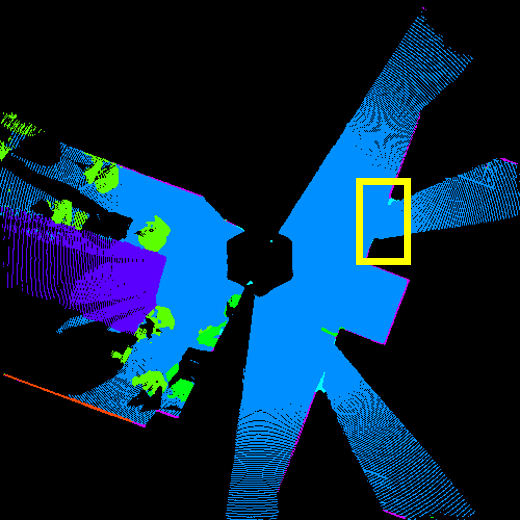}} &
        \raisebox{-0.5\height}{\includegraphics[width=0.19\textwidth]{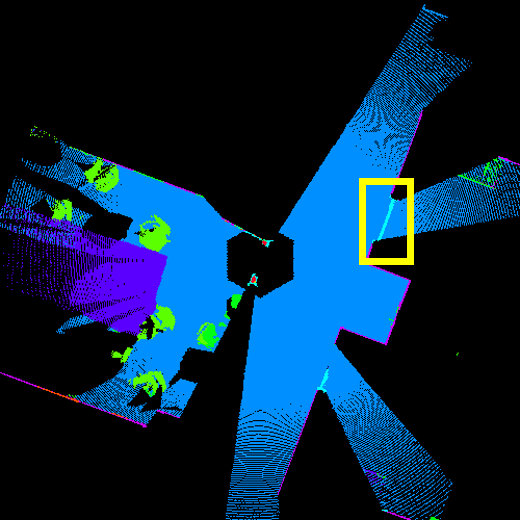}} &
        \raisebox{-0.5\height}{\includegraphics[width=0.19\textwidth]{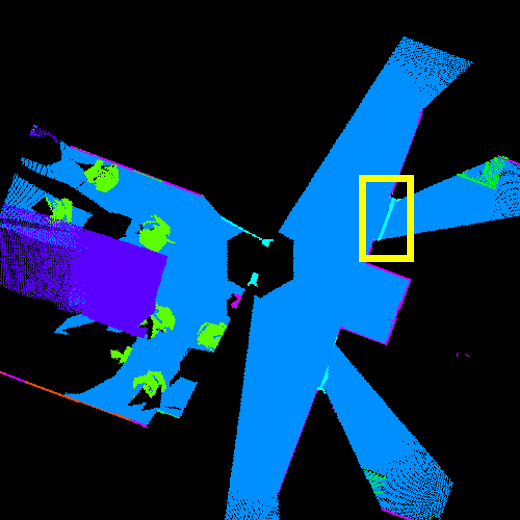}}\\
        
        \noalign{\vskip 2mm}
        RGB Input&Baseline&360Mapper&Ground Truth
    \end{tabular}
    \caption{\textbf{360BEV visualization and qualitative analysis} on the 360BEV-Stanford dataset. Black regions are the \texttt{void} class, indicating the invisible areas in BEV semantic maps. Zoom in for better view.}
    \label{fig:s2d3d_BEV}
    \vskip -3ex
\end{figure*}

We further introduce the qualitative results of 360BEV task on the 360BEV-Standford dataset in Fig.~\ref{fig:s2d3d_BEV}. The RGB input, the BEV semantic mapping results of the baseline and 360 Mapper, the BEV semantic mapping ground truth are depicted from left to right, where the color map is shown at the top of Fig.~\ref{fig:s2d3d_BEV}. The \textit{chairs} of the first and the second sample are correctly predicted by our method while they are partially or entirely missed by the baseline. Compared with the 360Mapper, the baseline shows more false prediction especially regarding some furniture, \eg, the false predicted \textit{bookcase} at the third sample, which should be predicted as \textit{chairs}. At the last row of Fig.~\ref{fig:s2d3d_BEV}, the challenging \textit{door} is not recognized by the baseline model, while our 360Mapper can provide accurate \textit{door} segmentation result, even it is a thin line in the BEV map. Our method shows overall superior performance on the proposed task compared with the baseline in terms of the semantic segmentation performance on small objects, which further illustrates the strength by using 360Attention.

\subsection{Analysis on 360BEV-Matterport}
\begin{figure*}[t]
    \footnotesize
    \setlength\tabcolsep{1pt}
    \begin{center}
    {
    \newcolumntype{P}[1]{>{\centering\arraybackslash}p{#1}}
    \begin{tabular}{@{}*{21}{P{0.087\columnwidth}}@{}}
     {\cellcolor[rgb]{0,   0,   0}}\textcolor{white}{void} 
    &{\cellcolor[rgb]{0.68,0.78,0.91}}\textcolor{white}{wall} 
    &{\cellcolor[rgb]{0.44,0.50,0.56}}\textcolor{white}{floor}
    &{\cellcolor[rgb]{0.60,0.87,0.54}}\textcolor{black}{chair}
    &{\cellcolor[rgb]{0.77,0.69,0.84}}\textcolor{white}{door}
    &{\cellcolor[rgb]{1.00,0.50,0.05}}\textcolor{white}{table} 
    &{\cellcolor[rgb]{0.84,0.15,0.16}}\textcolor{white}{pictu.} 
    &{\cellcolor[rgb]{0.12,0.47,0.71}}\textcolor{white}{furni.}
    &{\cellcolor[rgb]{0.74,0.74,0.13}}\textcolor{black}{objec.}
    &{\cellcolor[rgb]{1.00,0.60,0.59}}\textcolor{black}{windo.}
    &{\cellcolor[rgb]{0.17,0.63,0.17}}\textcolor{white}{sofa} 
    &{\cellcolor[rgb]{0.89,0.47,0.76}}\textcolor{black}{bed}
    & {\cellcolor[rgb]{0.87,0.62,0.84}}\textcolor{black}{sink}
    &{\cellcolor[rgb]{0.58,0.40,0.74}}\textcolor{white}{stairs} 
    &{\cellcolor[rgb]{0.55,0.64,0.32}}\textcolor{white}{ceil.} 
    &{\cellcolor[rgb]{0.52,0.24,0.22}}\textcolor{white}{toilet} 
    &{\cellcolor[rgb]{0.62,0.85,0.90}}\textcolor{black}{mirror} 
    &{\cellcolor[rgb]{0.61,0.62,0.87}}\textcolor{black}{show.}
    &{\cellcolor[rgb]{0.91,0.59,0.61}}\textcolor{black}{batht.}
    &{\cellcolor[rgb]{0.39,0.47,0.22}}\textcolor{white}{count.} 
    &{\cellcolor[rgb]{0.55,0.34,0.29}}\textcolor{white}{shelv.} \\
    \end{tabular}
    }
    \end{center}
    \vskip -3mm
    \centering
    \begin{tabular}{c c c c}
        \vspace{1pt}
        \raisebox{-0.5\height}{\includegraphics[width=0.38\textwidth]{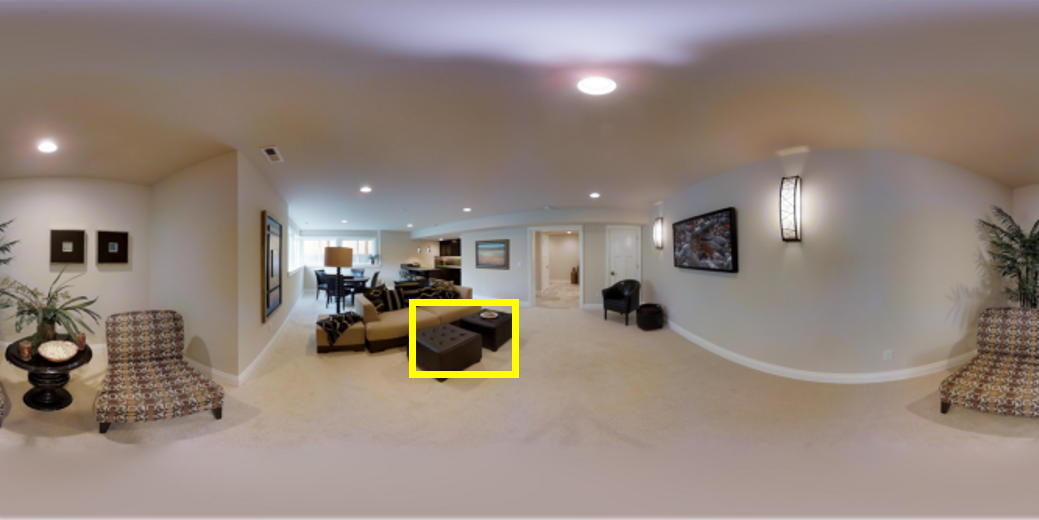}} &
        \raisebox{-0.5\height}{\includegraphics[width=0.19\textwidth]{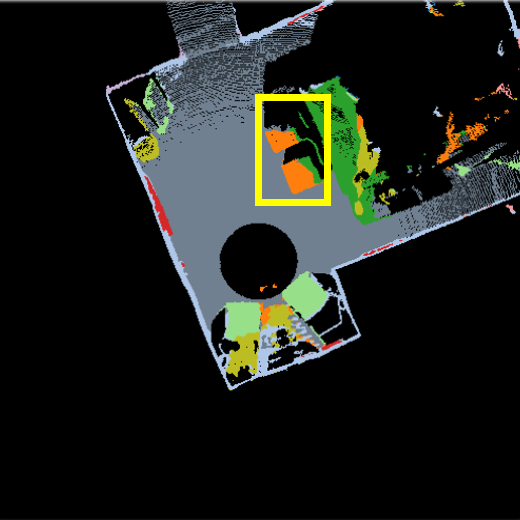}} &
        \raisebox{-0.5\height}{\includegraphics[width=0.19\textwidth]{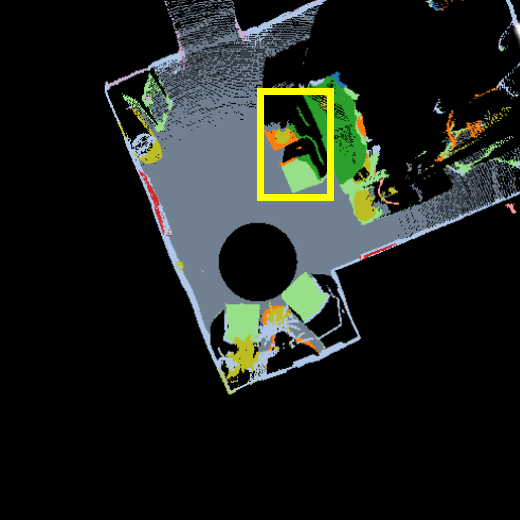}} &
        \raisebox{-0.5\height}{\includegraphics[width=0.19\textwidth]{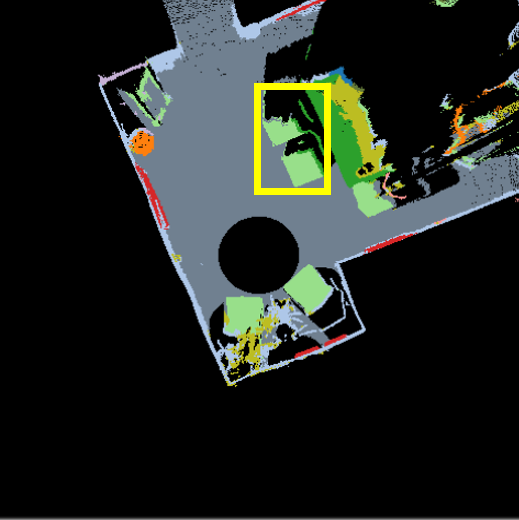}}\\
        \vspace{1pt}

        \raisebox{-0.5\height}{\includegraphics[width=0.38\textwidth]{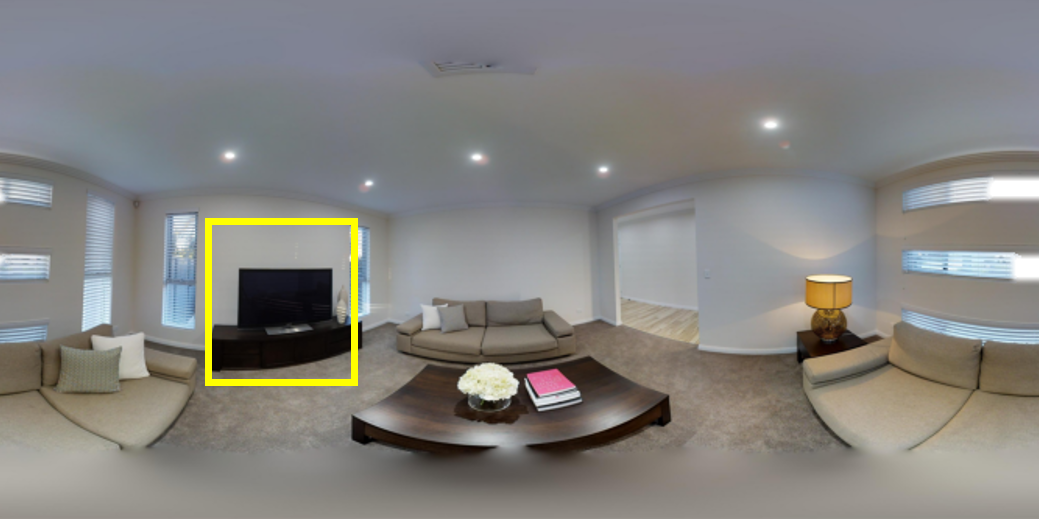}} &
        \raisebox{-0.5\height}{\includegraphics[width=0.19\textwidth]{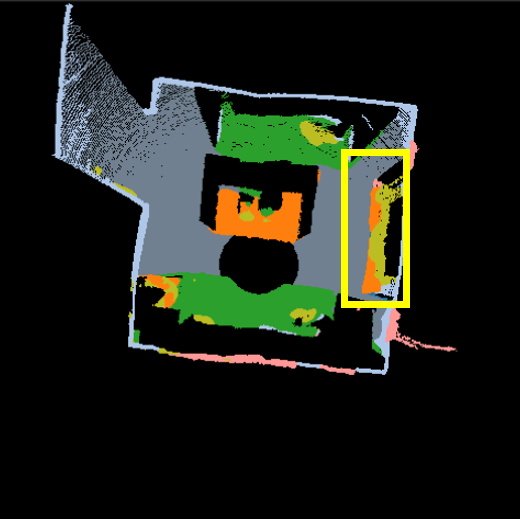}} &
        \raisebox{-0.5\height}{\includegraphics[width=0.19\textwidth]{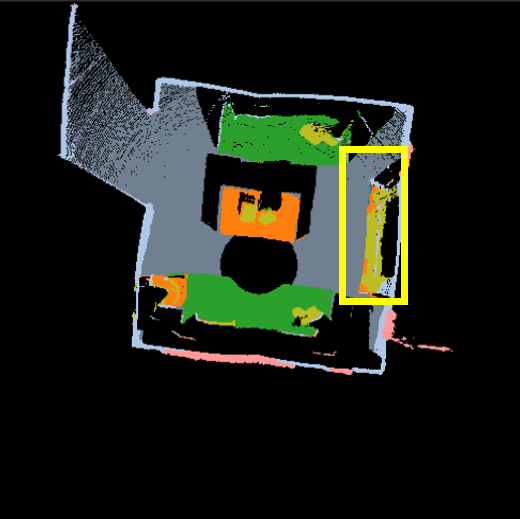}} &
        \raisebox{-0.5\height}{\includegraphics[width=0.19\textwidth]{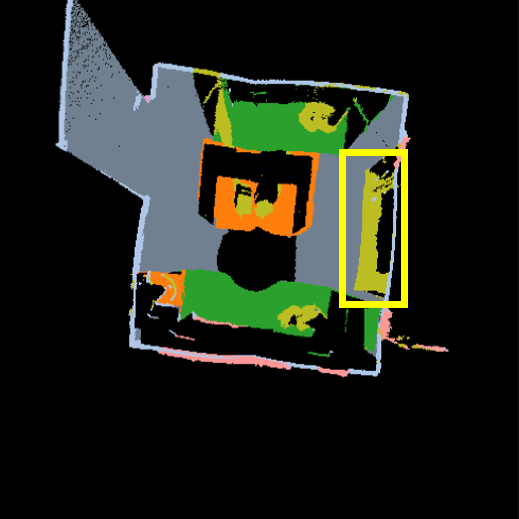}}\\
        
        \vspace{1pt}

        \raisebox{-0.5\height}{\includegraphics[width=0.38\textwidth]{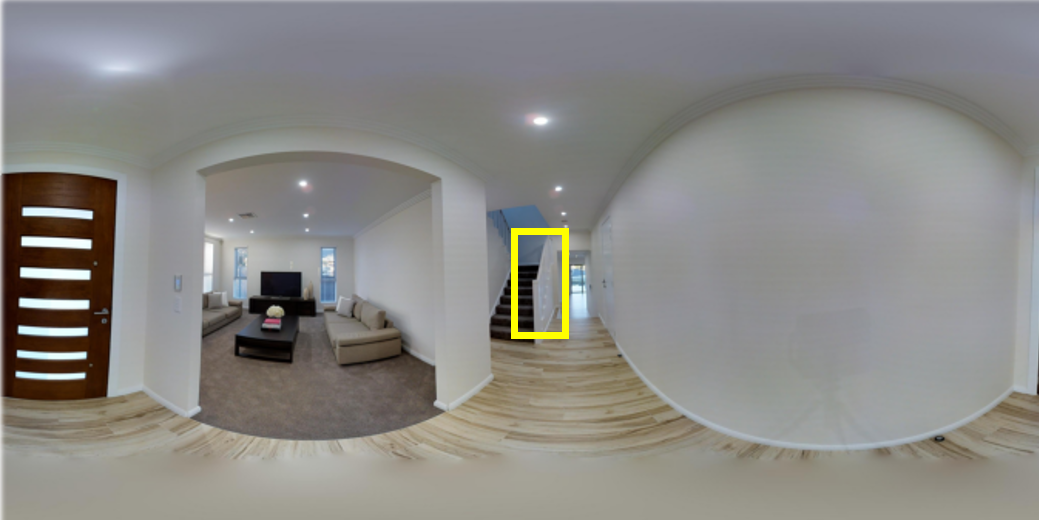}} &
        \raisebox{-0.5\height}{\includegraphics[width=0.19\textwidth]{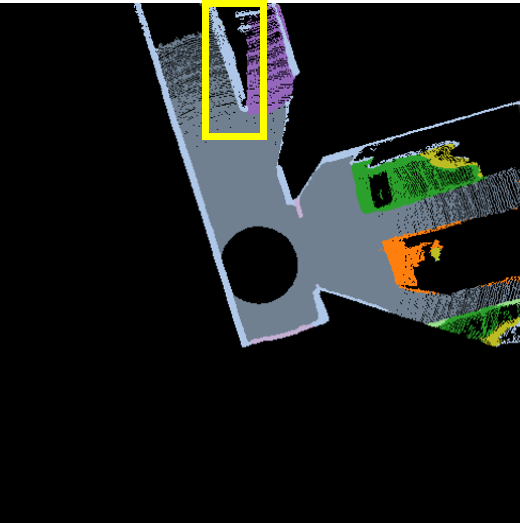}} &
        \raisebox{-0.5\height}{\includegraphics[width=0.19\textwidth]{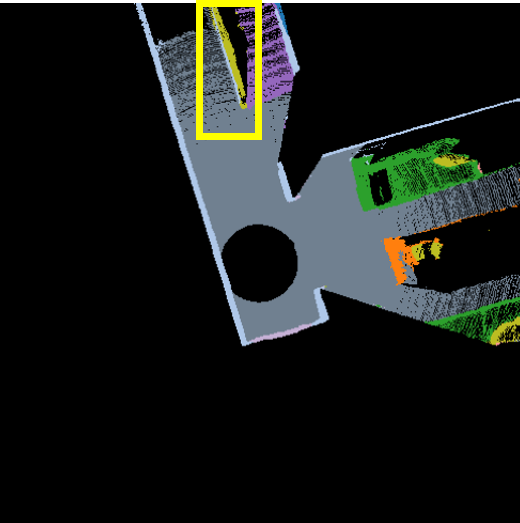}} &
        \raisebox{-0.5\height}{\includegraphics[width=0.19\textwidth]{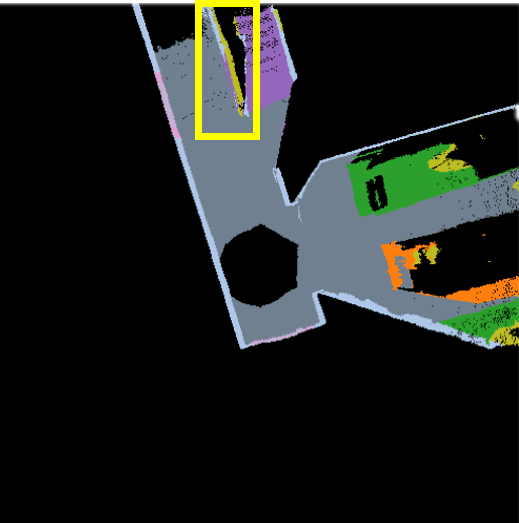}}\\
        
        \vspace{1pt}

        \raisebox{-0.5\height}{\includegraphics[width=0.38\textwidth]{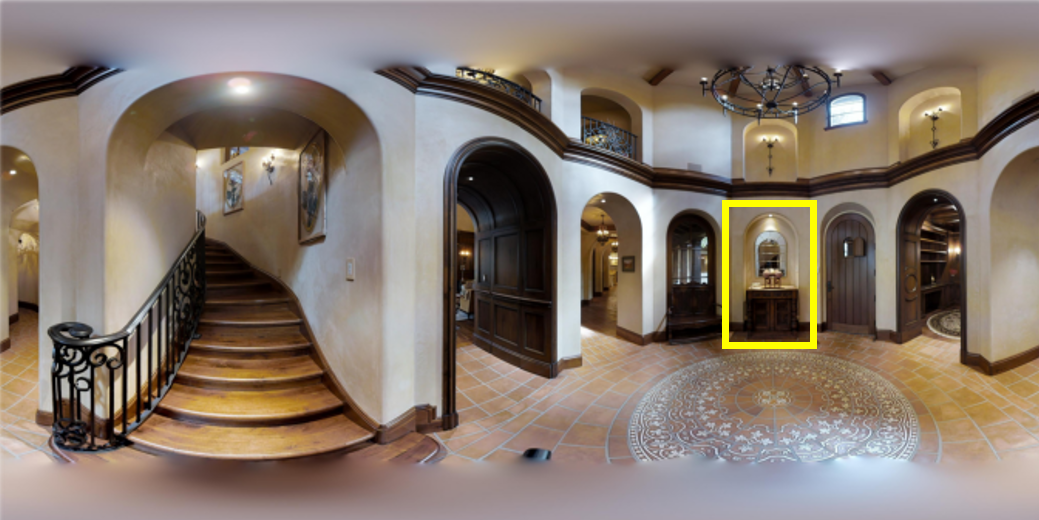}} &
        \raisebox{-0.5\height}{\includegraphics[width=0.19\textwidth]{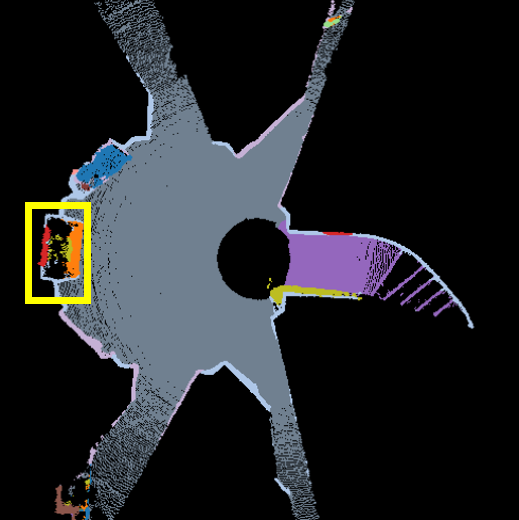}} &
        \raisebox{-0.5\height}{\includegraphics[width=0.19\textwidth]{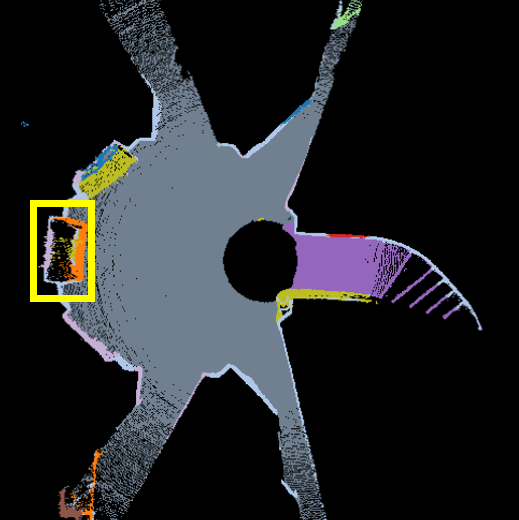}} &
        \raisebox{-0.5\height}{\includegraphics[width=0.19\textwidth]{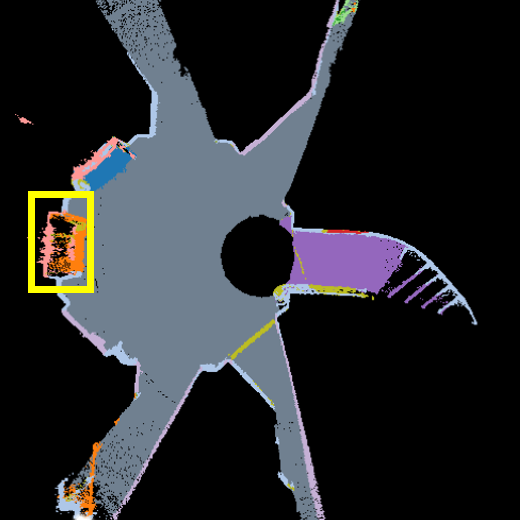}}\\
        \vspace{1pt}

        \raisebox{-0.5\height}{\includegraphics[width=0.38\textwidth]{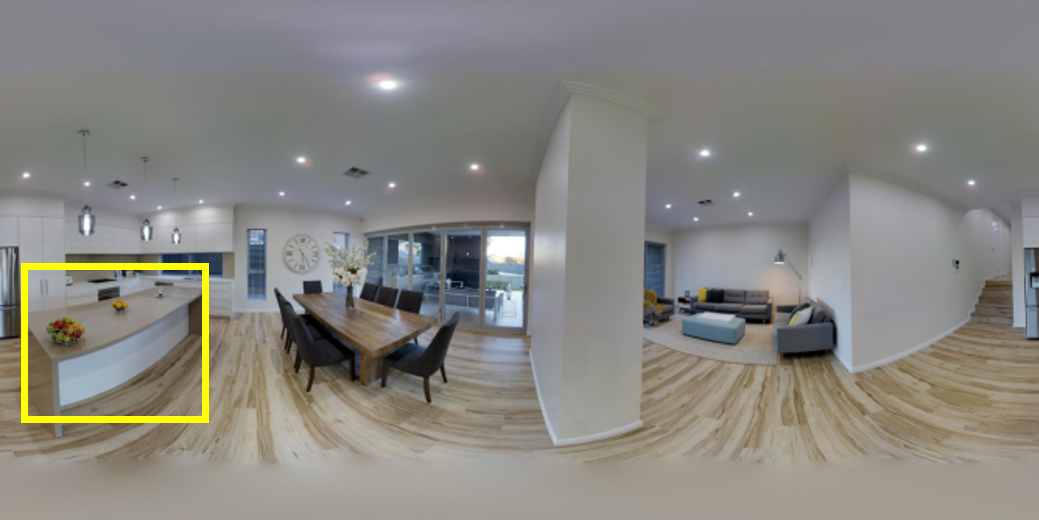}} &
        \raisebox{-0.5\height}{\includegraphics[width=0.19\textwidth]{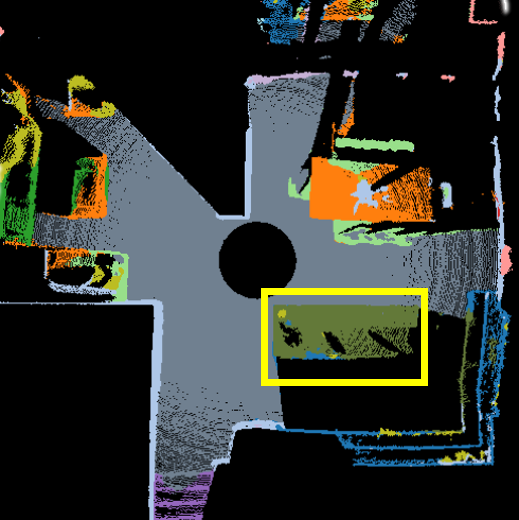}} &
        \raisebox{-0.5\height}{\includegraphics[width=0.19\textwidth]{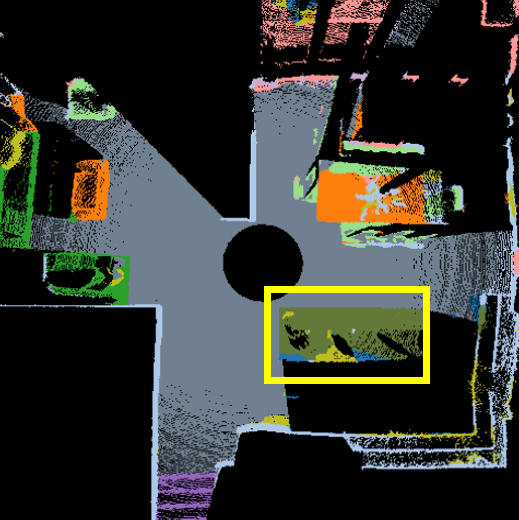}} &
        \raisebox{-0.5\height}{\includegraphics[width=0.19\textwidth]{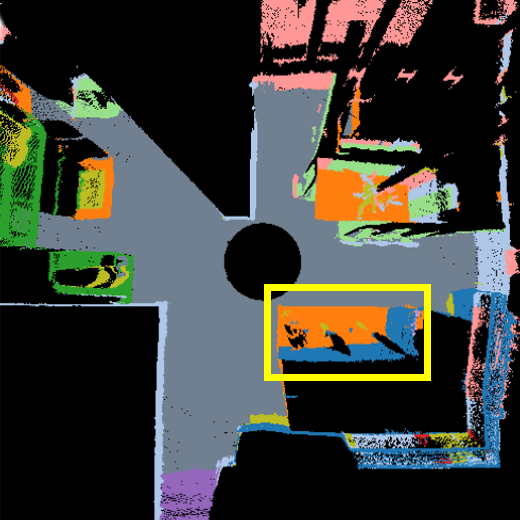}}\\
        
        \noalign{\vskip 2mm}
        RGB Input&Baseline&360Mapper&Ground Truth
    \end{tabular}
    \caption{\textbf{360BEV visualization and qualitative analysis} on the 360BEV-Matterport dataset. Black regions are the \texttt{void} class, indicating the invisible areas in BEV semantic maps. Zoom in for better view.}
    \label{fig:mp3d_360bev}
    \vskip -3ex
\end{figure*}

Fig.~\ref{fig:mp3d_360bev} presents qualitative results for the 360BEV task on the 360BEV-Matterport dataset. We observe that our 360Mapper outperforms the baseline method Trans4Map~\cite{chen2022trans4map} in terms of accurately segmenting small objects. In particular, the baseline method exhibits more false predictions, such as the misclassified \textit{chair} in the first sample and \textit{object} misidentified as \textit{table} in the second sample. Surprisingly, the different steps of \textit{stairs} in the third and the fourth sample are recognized correctly by both methods.
However, we find the fifth sample to be particularly challenging, as both the baseline and our 360Mapper recognize the object in the center of the image as a \textit{counter}, which is a \textit{table} as shown in the ground truth. This failure case shows the difficulty of accurately distinguishing between similar object categories from the context of panoramic images to the bird's-eye-view semantic maps.


\end{document}